\documentclass{article}
\usepackage[preprint]{neurips}
\usepackage[american]{babel}
\usepackage{natbib}
\usepackage[hyphens]{url}
\usepackage{graphicx}
\usepackage{caption}
\usepackage[T1]{fontenc}
\usepackage{listings}
\usepackage{booktabs}

\usepackage{mathtools}
\usepackage{amsfonts}
\usepackage{amsmath}
\usepackage{amssymb}
\usepackage{amsthm}
\usepackage{nicefrac}
\usepackage{xcolor}
\usepackage{tikz}
\usepackage{tcolorbox}
\usepackage{multirow}
\usepackage{float}
\usepackage{array}
\usepackage{tabularx}
\usepackage{subcaption}
\usepackage{pifont}
\newcommand{\yes}{\textcolor{green!55!black}{\ding{51}}}
\newcommand{\no}{\textcolor{red!75!black}{\ding{55}}}
\usepackage{fontawesome5}
\usepackage{makecell}
\usepackage{colortbl}
\usepackage[inline]{enumitem}
\usepackage{pgfplots}
\pgfplotsset{compat=1.18}
\usepackage{pgfplotstable}
\usepackage{placeins}
\usepackage{commath}
\usepackage[capitalize,noabbrev]{cleveref}
\crefname{algocf}{Algorithm}{Algorithms}
\Crefname{algocf}{Algorithm}{Algorithms}
\usepackage{longtable}
\tcbuselibrary{skins, breakable, listings}

\usepackage[ruled,vlined]{algorithm2e}
\SetKw{KwNotIn}{not\;in}
\SetKw{KwAnd}{and}
\SetAlgoNlRelativeSize{-1}
\SetNlSty{textbf}{}{}
\DontPrintSemicolon
\SetKw{KwRet}{return}
\SetKw{KwBreak}{break}
\SetKw{KwContinue}{continue}
\SetKwProg{Fn}{Function}{:}{}

\definecolor{agentA}{HTML}{2E86AB}
\definecolor{agentB}{HTML}{E07A5F}
\definecolor{stepbg}{HTML}{F7F9FC}
\definecolor{notebg}{HTML}{FFF8E7}
\definecolor{actionbg}{HTML}{EEF6EE}
\definecolor{backbg}{HTML}{FFF0F0}
\definecolor{successgreen}{HTML}{2D6A4F}
\definecolor{lightgray}{HTML}{F2F2F2}
\definecolor{darkgray}{HTML}{444444}
\definecolor{andcolor}{RGB}{255, 100, 0}
\definecolor{orcolor}{RGB}{34, 139, 34}
\definecolor{actioncolor}{RGB}{130,31,158}
\definecolor{entercolor}{RGB}{1, 50, 32}
\definecolor{exitcolor}{RGB}{1, 50, 32}
\definecolor{failedcolor}{RGB}{204, 0, 0}
\definecolor{expansioncolor}{RGB}{102, 0, 204}
\definecolor{repaircolor}{RGB}{0, 153, 153}
\definecolor{checkcolor}{RGB}{153, 153, 0}
\definecolor{updatecolor}{RGB}{0, 51, 102}
\definecolor{summarizercolor}{RGB}{102, 51, 0}
\definecolor{pastelblue}{RGB}{222, 235, 247}
\definecolor{pastelgreen}{RGB}{223, 240, 216}
\definecolor{pastelorange}{RGB}{255, 236, 214}
\definecolor{lightergray}{gray}{0.95}
\definecolor{deletecolor}{RGB}{220, 50, 50}
\definecolor{addcolor}{RGB}{0, 130, 60}
\definecolor{revisecolor}{RGB}{0, 80, 180}
\definecolor{structcolor}{RGB}{150, 80, 0}

\newtcolorbox{agentbox}[2][]{
  colback=#2!8,
  colframe=#2,
  fonttitle=\bfseries\small,
  boxrule=0.8pt,
  arc=4pt,
  left=6pt, right=6pt, top=4pt, bottom=4pt,
  #1
}

\newtcolorbox{stepbox}[1][]{
  colback=stepbg,
  colframe=darkgray!30,
  boxrule=0.4pt,
  arc=2pt,
  left=5pt, right=5pt, top=3pt, bottom=3pt,
  #1
}

\newcommand{\SA}{\textsc{StructuredAgent}}
\newcommand{\sa}{{\textsc{StructuredAgent}}}
\newcommand{\bca}{{BasicClaudeAction}}
\newcommand{\SUCCESS}{\textsc{success}}
\newcommand{\samem}{{StructuredAgentMem}}
\newcommand{\ao}{{AgentOccam}}
\newcommand{\wrep}{{WebarenaReplication}}

\newcommand{\sr}{{SteP}}

\newcommand{\eg}{\textit{e.g., }}

\newcommand{\AND}{\textcolor{andcolor}{\textsc{And}}}
\newcommand{\OR}{\textcolor{orcolor}{\textsc{Or}}}
\newcommand{\ANDOR}{{\textsc{And/Or}}}
\newcommand{\ACTION}{\textcolor{actioncolor}{\textsc{Action}}}
\newcommand{\ENTERING}{\textcolor{entercolor}{\textsc{Entering}}}
\newcommand{\EXITING}{\textcolor{exitcolor}{\textsc{Exiting}}}
\newcommand{\FAILED}{\textcolor{failedcolor}{\textsc{Failed}}}
\newcommand{\NODEEXPANSION}{{\textsc{Node Expansion}}}
\newcommand{\NODEREPAIR}{{\textsc{Node Repair}}}
\newcommand{\NODECHECK}{{\textsc{Node Completion Check}}}
\newcommand{\GLOBALTREE}{{\textsc{Global Tree Update}}}
\newcommand{\SUMMARIZER}{{\textsc{Observation Summarizer}}}

\newcolumntype{L}{>{\raggedright\arraybackslash}X}

\newcommand{\BibTeX}{\rm B\kern-.05em{\sc i\kern-.025em b}\kern-.08em\TeX}

\newcommand{\DELETE}[1]{}
\newcommand{\DELETENOTE}[1]{}
\newcommand{\ADD}[1]{}
\newcommand{\REVISE}[1]{}
\newcommand{\STRUCT}[1]{}

\theoremstyle{plain}

\theoremstyle{definition}

\theoremstyle{remark}

\title{\SA{}: Planning with AND/OR Trees for Long-Horizon Web Tasks}

\author{%
  Elita A. Lobo \\
  University of Massachusetts Amherst \\
  \texttt{elobo@umass.edu} \\
  \And
  Xu Chen \\
  Amazon \\
  \texttt{frank.chenxu95@gmail.com} \\
  \And
  Jingjing Meng \\
  Amazon\thanks{Work done while at Amazon} \\
  \texttt{jingjing.meng1@gmail.com} \\
  \And
  Nan Xi \\
  Virginia Commonwealth University \\
  \texttt{xin@vcu.edu} \\
  \And
  Yang Jiao \\
  Amazon \\
  \texttt{jaoyan@amazon.com} \\
  \And
  Chirag Agarwal \\
  University of Virginia \\
  \texttt{qze3wn@virginia.edu} \\
  \And
  Yair Zick \\
  University of Massachusetts Amherst \\
  \texttt{yzick@umass.edu} \\
  \And
  Yan Gao \\
  Amazon \\
  \texttt{yanngao@amazon.com} \\
}

\begin{document}

\maketitle
\begin{abstract}
Existing LLM-based web agents struggle on complex, long-horizon tasks due to limited in-context memory, weak planning abilities, and greedy behaviors that lead to premature termination. To address these challenges, we propose \SA{}, a hierarchical planning framework that interleaves planning and execution via dynamic $\ANDOR$ trees. The framework separates structural planning from LLM-based reasoning, enabling principled error recovery through node repair, systematic exploration of alternatives via OR nodes, and modular plans that can facilitate human intervention. On WebArena (630 tasks), \SA{} achieves a $\sim$53\% success rate vs.\ $\sim$46\% for AgentOccam, and on complex multi-constraint Amazon shopping tasks, gains reach 10\% over the strongest baseline.
\end{abstract}

    \begin{figure*}[t]
    \centering
\includegraphics[width=0.80\linewidth]{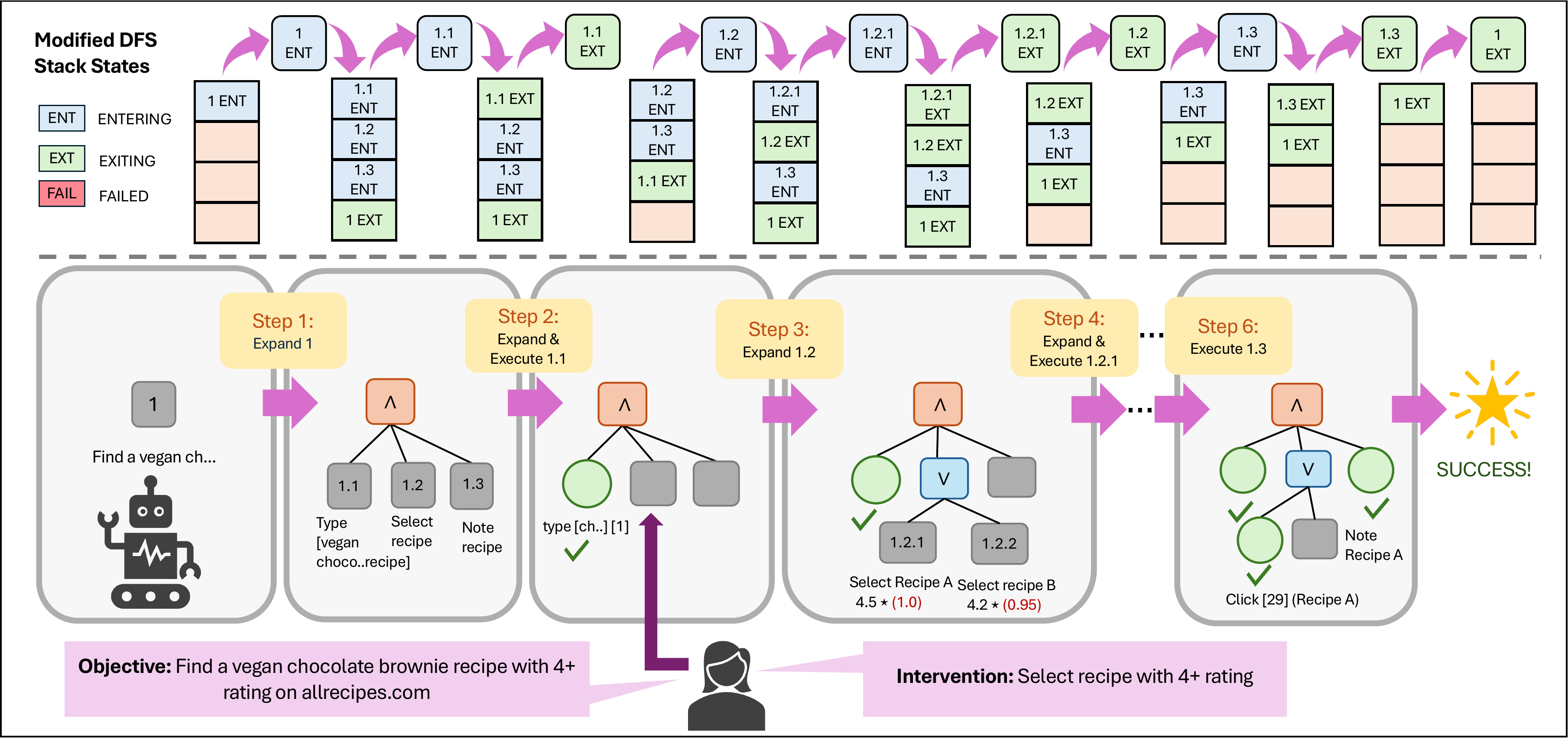}
    \setlength{\abovecaptionskip}{2pt}
    \caption{Illustration of \SA{} solving a web task via greedy DFS of a dynamically constructed $\ANDOR$ tree. The root node represents the task objective and is expanded into sub-tasks that are progressively refined and executed. Node types are color-coded to distinguish OR ($\vee$), AND ($\wedge$), and ACTION nodes. The top half tracks the corresponding DFS stack states, where each node enters the stack in one of three states: ENTERING (ENT), EXITING (EXT), or FAILED (FAIL). Node repair and pruning are supported but not triggered in this example, and human intervention is optional.}
    \label{fig:structuredagentexample}
\end{figure*}

\section{Introduction}
  
  Recent advances in large language models (LLMs) have enabled agentic systems for sequential decision-making. Such agents must perceive their environment, reason across multiple time steps, and take actions that
  optimize long-term objectives. A particularly promising application area is the development of web-browsing agents, which navigate and interact with the internet to accomplish complex, goal-driven tasks
  including information seeking, form filling, and transactional activities. Despite notable progress, building robust and interpretable web-based agents remains challenging.

    \looseness=-1 Although many of these tasks are intuitively straightforward for
  humans, LLM-powered web agents continue to struggle with complex, multi-step
  browsing~\citep{yang2025agentoccam,erdogan2025planandactimprovingplanningagents,qi2025webrl}.
  We focus on agents without specialized memory or retrieval machinery, the
  prevailing setup among publicly reported web agents. In this setting, four
  challenges account for the bulk of observed failures:
  \begin{enumerate*}[label={(\arabic*)}]
      \item {Limited in-context memory}, which prevents the agent from
      tracking alternative solution paths across dense web pages, where a single
      page can exceed 20k tokens (see~Figure~\ref{res:tokendist}). Even when
      long-context models are used, existing agents condition only on the
      current observation and a summary of the interaction history.
      \item {Difficulty constructing and executing complex plans} that
      must remain coherent over extended horizons.
      \item {Lack of robust error handling}, leaving agents unable to
      recover from failures or explore alternative solutions~\citep{koh2025tree}.
      \item {Greedy decision-making}, which leads to premature task
      termination before all user constraints are
      satisfied~\citep{schmied2025llmsgreedyagentseffects}
      (see~Appendix~\ref{app:example}).
  \end{enumerate*}
  % \looseness=-1 Although many of these tasks are intuitively straightforward for
  % humans, LLM-powered web agents continue to struggle with complex, multi-step
  % browsing~\citep{yang2025agentoccam,erdogan2025planandactimprovingplanningagents,qi2025webrl}.
  % We focus on agents that are not equipped with specialized memory or retrieval
  % machinery, acting instead from the current observation and a bounded action
  % history, as is the case for most publicly reported web agents. In this setting,
  % four challenges account for the bulk of observed failures:
  % %
  % \begin{enumerate*}[label={(\arabic*)}]
  %     \item \textit{Limited in-context memory}, which restricts the agent's
  %     ability to track alternative solution paths across dense web pages, where a single web-page may exceed 20k tokens (see~Figure~\ref{res:tokendist}).
  %     Long-context models relax this constraint but do not remove it, since agents
  %     typically condition only on the current page and a truncated history,
  %     \item \textit{Inability to construct and execute complex plans} over extended
  %     horizons;
  %     \item \textit{Lack of robust error-handling mechanisms} to recover from
  %     failures and explore alternative solutions~\citep{koh2025tree}; and
  %     \item \textit{Greedy decision-making} that leads to premature task
  %     termination without satisfying all user
  %     constraints~\citep{schmied2025llmsgreedyagentseffects}
  %     (see~Appendix~\ref{app:example}).
  % \end{enumerate*}

\looseness=-1 In cognitive science~\citep{newell1972human}, human problem solving is framed as heuristic search coupled with \textit{means-ends analysis}, where complex tasks are recursively decomposed into
  manageable subproblems. This perspective motivates hierarchical planning in AI, where agents improve efficiency by breaking down complex goals into nested sub-goals and actions at multiple levels of abstraction.
  Such methods are particularly well-suited to open-ended, web-based tasks characterized by uncertainty and expansive state and action spaces.
  
  Classical hierarchical planners, however, rest on assumptions that web environments violate. A* and AO* require admissible heuristics or fully specified domain models (e.g., PDDL), which are unavailable when
  state and action spaces are unbounded and only partially observable. Monte Carlo tree search (MCTS) is prohibitively expensive, requiring multiple rollouts before selecting an action at every step. Hierarchical
  task network (HTN) planners depend on pre-specified decomposition schemas, which complex web tasks rarely provide. The closest applicable works~\citep{koh2025tree,erdogan2025planandactimprovingplanningagents}
  either operate at the action level without hierarchical structure or construct linear plans without backtracking.
  
  \looseness=-1 Recent research has explored explicit planning mechanisms for web agents. \citet{koh2025tree} use tree-based planning with A* search to explore alternative solution paths upon failure. However,
  their method operates at the level of individual actions, which limits exploitation of compositional task structure, and the absence of admissible heuristics can lead to inefficient path switching.
  \citet{yang2025agentoccam} propose incremental tree construction with prune-and-branch strategies, but rely solely on the LLM for plan construction and execution, often resulting in simplistic plans and limited
  error recovery. \citet{erdogan2025planandactimprovingplanningagents} introduce dynamic plan revisions, but permit only sequential plans determined prior to execution, without error back-propagation.
  
  \looseness=-1 To address these challenges, we propose \SA{}, a hierarchical planning framework that interleaves planning and execution via ordered $\ANDOR$ trees. The key architectural choice is a separation of concerns: the framework constructs and maintains the planning tree, while the LLM is invoked only for local operations such as node expansion or repair. Our main contributions are:
  
  \begin{itemize}[leftmargin=0.35cm, itemsep=0pt, topsep=2pt]
      \item We introduce a hierarchical planning framework enabling inference-time planning via ordered \ANDOR{} trees. The framework supports compositional reasoning, dynamic plan revision, and error
  backpropagation, while producing modular plans that can facilitate error localization and human intervention.
  %     \item For web-based information-seeking tasks, we develop a structured memory module that tracks candidate entities and their satisfied constraints, improving constraint satisfaction during planning and
  % execution.
      \item We evaluate on WebArena (630 tasks), WebVoyager (129 tasks), and a custom 60-task shopping benchmark, achieving 7--10\% improvement over the strongest baseline on complex tasks.
  \end{itemize}

\section{Preliminaries}\label{sec:preliminaries}
Web-browsing tasks are partially observable and inherently stochastic, as web page content and structure can change dynamically over time, either unpredictably or in response to the agent's actions, and the agent lacks access to the full underlying browser state. The agent is typically limited to the current HTML page, prior actions, and extracted metadata, requiring it to make decisions under uncertainty while maintaining internal beliefs about the environment.

 We model the agent-browser interaction as a  Partially Observable Markov Decision Process (POMDP) $\langle\mathcal{O},\mathcal{S},\mathcal{A},R,T,p_0,\gamma\rangle$~\cite{kaelbling1998planning}.
  An \emph{observation} $o \in \mathcal{O}$ comprises the task description,
  agent-constructed plans, interaction metadata and the HTML DOM, $s \in
  \mathcal{S}$ is the true underlying \emph{state}, and an \emph{action} $a \in
  \mathcal{A}$ is either an atomic browser operation (click, scroll, type, go back)
  or a token sequence describing a sub-goal or strategy. The transition kernel
  $T(s'\mid s,a)$ is induced by the website and unknown to the agent, $p_0$ is the
  initial state distribution, and $\gamma$ discounts future reward. We represent the agent's history-dependent \emph{policy} $\pi$ using an LLM. Specifically, at timestep $t$, the observation history $h_t = \{o_1, \dots, o_t\}$ is encoded as a token sequence $x_t = [x_1, \dots, x_m]$, and the action $a_t = [y_1, \dots, y_n]$ is generated autoregressively as $\Pr(a_t \mid h_t) = \prod_{i=1}^{n} \pi(y_i \mid x_1, \dots, x_m, y_1, \dots, y_{i-1})$. We provide the full formalization of the problem in Appendix~\ref{app:pomdp}.

% \looseness=-1 Next, we formalize the structure of the $\ANDOR$ planning tree and describe its construction and traversal. We then detail the four tree operations and the observation and nodes aggregation module used to monitor task progress and agent state. Finally, we present the planning algorithm that integrates these components for efficient and adaptive execution.

\section{\SA{}}
The central idea of \SA{} is to interleave planning and execution via a dynamically constructed AND/OR tree. As new information becomes available, the agent must expand, revise, and prune the planning tree to
  eliminate unpromising paths and focus search on promising ones.
  
  Rather than delegating full tree management to the LLM, as in AgentOccam~\citep{yang2025agentoccam}, \SA{} manages all tree structure directly, invoking the LLM only for targeted, well-defined operations such as node
  expansion, repair, completion checking, and pruning. This separation improves reliability and facilitates interpretability.
  
  \subsection{AND/OR Planning Tree Structure}
  
  The planning tree is composed of three node types:
  (1) \AND{} nodes represent conjunctive sub-goals. All children must succeed for the parent to succeed. Children may be ordered (executed sequentially) or unordered.
  (2) \OR{} nodes represent alternative strategies. The parent succeeds as soon as any child succeeds.
  (3) \ACTION{} nodes are leaves corresponding to concrete browser operations such as typing a query, clicking a button, scrolling, navigating back, or recording a note. We detail the tree operations in the following subsections.
  
  \looseness=-1 This hierarchical structure allows the agent to decompose tasks into sub-goals and reason explicitly about alternative solution paths.
  For example, when solving a task such as \texttt{``find a coffee filter under \$50''}, the agent might construct an \AND{} node with two children representing searching for ``coffee filter'' and applying a price filter.
  Each sub-goal can further decompose or branch into \OR{} nodes that encode different navigation or filtering strategies. This flexibility allows the agent to represent a wide variety of decompositions while capturing the sequential dependencies that arise in web-based workflows.

In Figure~\ref{fig:intervention} we highlight a benefit of our framework's interpretable hierarchical 
planning structure, namely that it naturally supports human-in-the-loop oversight, whereby an operator can 
inspect the \ANDOR{} tree, detect incorrect sub-goal decompositions, and inject 
corrective sub-goals at any layer of the plan before execution proceeds further.

\subsection{$\ANDOR$ Tree Construction and Traversal}
Given a web-based task, the goal is to interleave planning and execution so that the agent can continually adapt its strategy as new information becomes available. To achieve this, we use an LLM as a high-level controller that issues localized tree-manipulation directives. These directives determine the type of a node, expand the node, repair it by adding or removing children, or verify whether its objective has been satisfied. The surrounding framework manages tree traversal and executes the corresponding operations.

We use a modified greedy iterative depth-first search (DFS) to expand nodes in $\SA$, starting at the root node that represents the task objective. When the agent encounters an $\AND$ node, it processes its children sequentially according to the node’s prescribed ordering. When it encounters an $\OR$ node, the agent selects a single child to explore by querying an LLM for an estimate of its likelihood of success based on summaries of prior actions and observations. 
% A learned value function may also guide this selection following prior work \citep{koh-etal-2024-visualwebarena}. 
If the agent reaches an $\ACTION$ node, it executes the action in the environment and updates the tree with newly observed information. This update may require pruning or revising certain nodes.

Unlike standard DFS, which operates on a fixed graph, our traversal incrementally constructs and maintains an evolving $\ANDOR$ tree whose structure changes as the agent gathers new information. In web-based environments, the search space is large, stochastic, and difficult to model, so sub-goals generated during planning may be imprecise or infeasible, leading to node failures during execution. To handle such failures, \SA{} supports systematic node revision and, after repeated unsuccessful attempts, node pruning. Pruning a child of an \AND{} node invalidates all subsequent siblings in its prescribed execution order, and these structural changes must propagate upward to preserve global plan coherence, potentially requiring nodes to be revisited multiple times during traversal.

\looseness=-1 These requirements motivate two modifications to standard DFS. First, we introduce three execution states for nodes in the DFS stack: \ENTERING{}, \EXITING{}, and \FAILED{}. A node enters
  \ENTERING{} when it has not yet been processed or when some of its children remain unprocessed. It transitions to \EXITING{} once execution completes for \ACTION{} nodes, once all children have been processed
  for \AND{} nodes, or once the selected child has been processed for \OR{} nodes. In the \EXITING{} state, the framework evaluates whether the node's objective has been satisfied. A node enters \FAILED{} if execution fails
  for \ACTION{} nodes or if any child fails or is pruned for \AND{}/\OR{} nodes, indicating that revision or pruning is required. These states support upward propagation of structural changes and allow the
  framework to process a node multiple times as the tree evolves.
  
  \looseness=-1 Second, classical iterative DFS uses only a binary visited/unvisited distinction, which is insufficient when dynamic pruning and revision require processing a node multiple times. To handle this,
  we assign each node one of six status values (summarized in~Table~\ref{tab:node_statuses} in the appendix) that capture structural, execution, and failure conditions, enabling the framework to manage the additional
  complexity introduced by dynamic tree evolution.
\begin{figure*}[h!]
    \centering
    \includegraphics[width=0.8\linewidth]{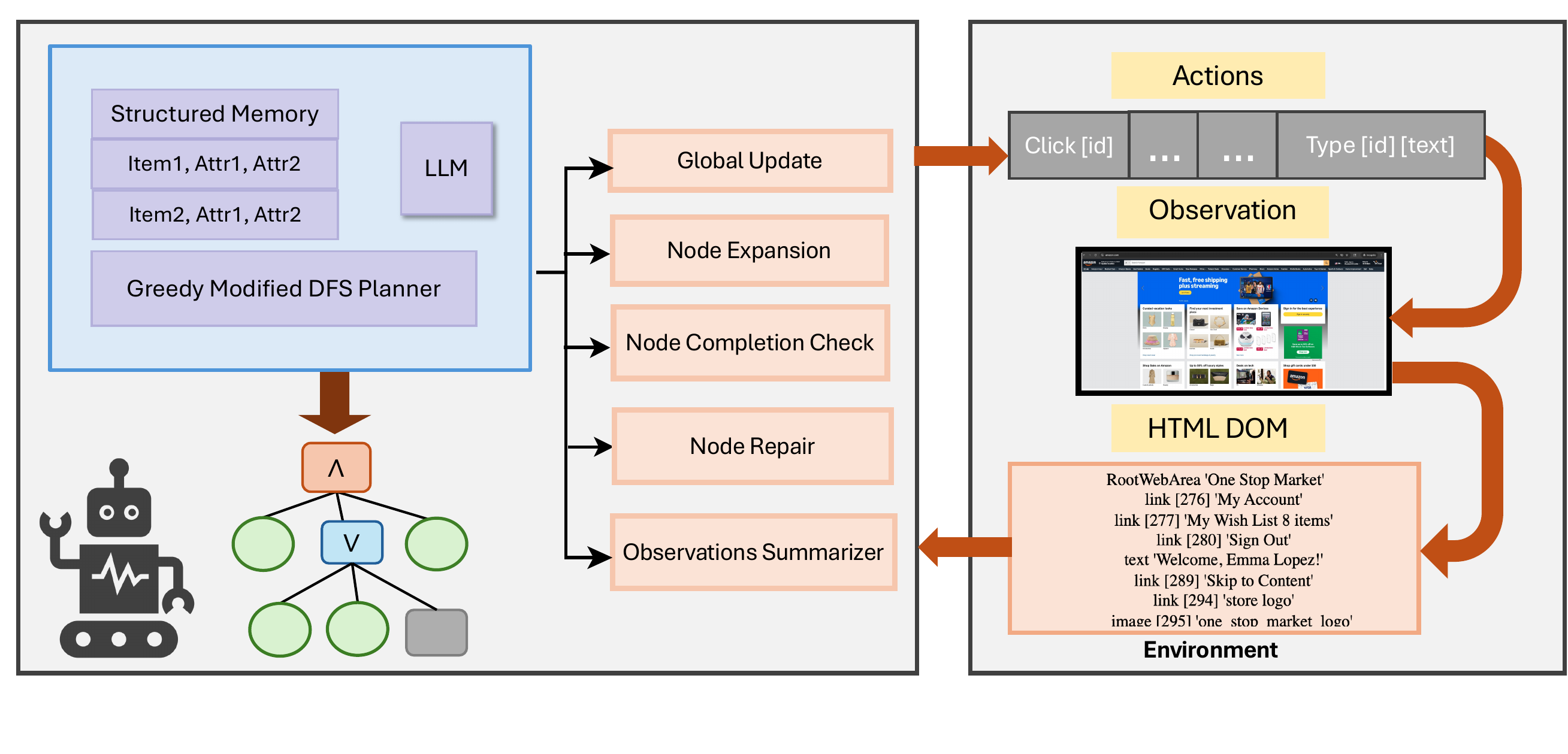}
    \vspace{-0.15in}
    \caption{Illustration of the \SA{} Framework.}
    \label{fig:framework}
\end{figure*}

We present the high-level tree construction and execution procedure used by $\SA$ in~\cref{alg:main}. We provide an illustration of node state transitions in~Figure~\ref{fig:dfs_states} (appendix). Full details of the planning algorithm are provided in Section~\ref{sec:planning_detail}. We now describe the fundamental tree operations and the agent modules required by the $\SA$ framework.

\begin{figure*}
    \centering
\includegraphics[width=1.0\linewidth]{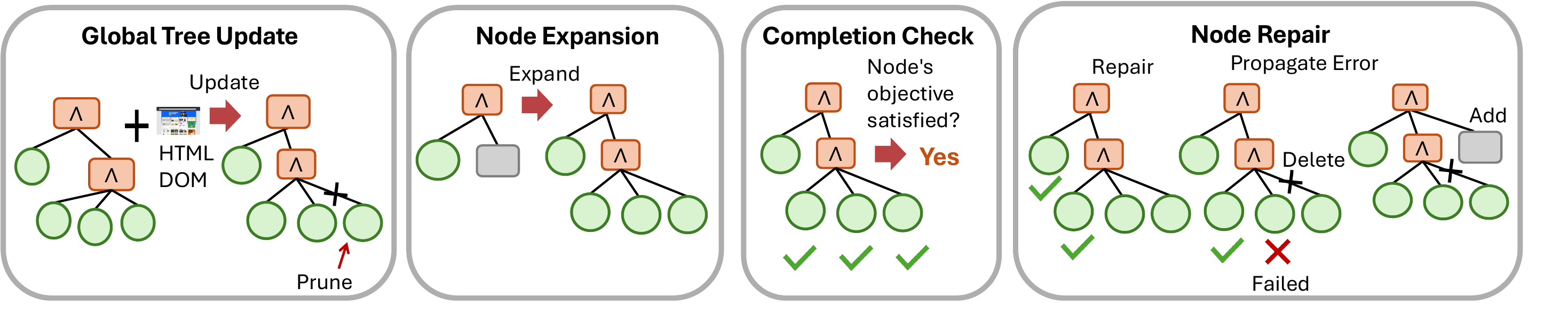}
    \caption{Overview of the primary operations used to construct and maintain the \ANDOR{} tree structure.}
    \label{fig:treeoperations}
\end{figure*}

 \subsection{Tree Operations and Agent Modules}\label{sec:treeoperations}
  
  We now describe the operations that construct and maintain the $\ANDOR$ tree under the guidance of an LLM acting as a high-level controller (illustrated in Figure~\ref{fig:treeoperations}, appendix). Since storing the full observation history in-context is infeasible,
  the agent maintains concise internal representations of browser interactions as proxies for the full state (detailed in~Table~\ref{tab:internal_representations}).
  
  Building on these representations, we introduce five specialized operators that prompt the backbone LLM to perform context-specific inference, with the framework verifying and executing the model's predictions.
  The \NODEEXPANSION{} operator determines node types and generates children: sub-goals for \AND{} nodes, executable actions for \ACTION{} nodes, and scored candidate strategies for \OR{} nodes. \NODEREPAIR{}
  revises failed nodes by proposing structural modifications or pruning when recovery is no longer viable. \GLOBALTREE{} maintains tractability by pruning irrelevant branches and refining node descriptions as new
  context becomes available. \NODECHECK{} evaluates whether \AND{} nodes have met their objectives by analyzing child outcomes and task summaries. Finally, \SUMMARIZER{} produces internal representations of task
  progress and observations, enabling context-aware reasoning. We provide the detailed prompts, inputs, outputs, and further specifications in~Appendices~\ref{app:prompts} and~\ref{app:treeoperations}.

\begin{algorithm}[t]
\scriptsize
\caption{TREE\_CONSTRUCTION(root)}
\label{alg:main}
\SetAlgoLined
stack $\gets [(root,\mathrm{ENTERING})]$\; counter $\gets 0$\;
\While{stack $\neq \emptyset$ \textbf{and} counter $<$ BUDGET}{
  (node, state) $\gets$ stack.pop()\;
  \lIf{$(node.status{=}\mathrm{SUCCESS} \wedge node.rev\_cnt{\ge}\mathrm{MAX}) \vee (node.status{=}\mathrm{DELETED})$}{\textbf{continue}}
  \uIf{$node.status=\mathrm{PRUNED}$}{BACKTRACK\_FAILURE(node)\; \textbf{continue}\;}
  \uIf{$state=\mathrm{ENTERING}$}{
    \lIf{node.type $=$ UNKNOWN}{node $\gets$ \NODEEXPANSION{}(node)}
    node.status $\gets$ VISITED\;
    \uIf{node.type $=$ ACTION}{
      result $\gets$ EXECUTE(node)\;
      \SUMMARIZER{}(node, result)\;
      \GLOBALTREE{}(tree)\;
      stack.push((node, EXITING))\;
    }
    \uElseIf{node.type $=$ AND}{
      stack.push((node, EXITING))\;
      \lForEach{child in reverse}{stack.push((child, ENTERING))}
    }
    \Else{
      stack.push((node, EXITING))\;
      stack.push((bestChild, ENTERING))\;
    }
  }
  \uElseIf{$state=\mathrm{EXITING}$}{
    PROCESS\_NODE\_EXITING(node, stack)\;
  }
  \ElseIf{$state=\mathrm{FAILED}$}{
    \uIf{node.type $=$ ACTION}{
      node.status $\gets$ PRUNED\;
      BACKTRACK\_FAILURE(node)\;
    }
    \uElseIf{node.type $=$ AND}{
      \lIf{rev\_cnt $<$ MAX}{node $\gets$ \NODEREPAIR{}(node)\; stack.push((node, ENT))}
      \lElse{node.status $\gets$ PRUNED\; BACKTRACK\_FAILURE(node)}
    }
    \ElseIf{node.type $=$ OR}{
      \uIf{$\exists$ unprocessed child}{
        stack.push((node, EXITING))\;
        stack.push((nextChild, ENTERING))\;
      }
      \lElseIf{rev\_cnt $<$ MAX}{node $\gets$ \NODEREPAIR{}(node)\; stack.push((node, ENT))}
      \lElse{node.status $\gets$ PRUNED\; BACKTRACK\_FAILURE(node)}
    }
  }
}
\end{algorithm}

  \looseness=-1 Equipped with the $\ANDOR$ tree and core operators, we now describe the modified greedy, iterative DFS procedure that \SA{} uses to build, update, and execute its plan. The agent maintains a DFS
  stack and processes each node in one of three states: \ENTERING{}, \EXITING{}, or \FAILED{}.
  
  \looseness=-1 When the agent encounters a node in the \ENTERING{} state, it checks whether the node has been expanded. If not, it invokes \NODEEXPANSION{} to determine the node type and generate its children: an
  ordered list of sub-goals for \AND{} nodes, ranked alternative strategies for \OR{} nodes, or a browser-level action for \ACTION{} nodes. The node is then re-pushed in \EXITING{} and its children are pushed in
  \ENTERING{}. For \AND{} nodes, all children are pushed in order and for \OR{} nodes, only the highest-ranked unprocessed child is pushed. For \ACTION{} nodes, the agent immediately executes the browser operation,
  updates its internal context via \SUMMARIZER{}, and invokes \GLOBALTREE{} to prune irrelevant branches and refine node descriptions.
  
  \looseness=-1 When the agent processes a node in the \EXITING{} state, it evaluates whether the node has satisfied its objective. An \ACTION{} node is marked successful or failed based on the browser interaction
  outcome. An \AND{} node succeeds only if all valid children succeed, confirmed via \NODECHECK{}. An \OR{} node succeeds as soon as any child succeeds. Once marked successful, a node is not revisited.
  
  \looseness=-1 When the agent encounters a node in the \FAILED{} state, it initiates recovery. If the failed node is an \ACTION{} node, it is pruned. If the failed node is a child of an \AND{} node, remaining unexecuted
  siblings are also removed since the conjunctive requirement can no longer be met. If an \AND{} node fails, the agent checks its revision budget: if budget remains, it calls \NODEREPAIR{} to add or modify
  children. However, if the repair is unsuccessful or the budget is exhausted, the node and all descendants are pruned and failure propagates upward. If an \OR{} node fails, the agent selects the next unprocessed child if one
  exists; otherwise it attempts repair within the revision limit or prunes the node, propagating failure upward.
  
  \looseness=-1 Through iterative expansion, execution, evaluation, and targeted repair, this procedure enables \SA{} to construct and adapt its hierarchical plan while incorporating new observations and
  responding to failures. See~Appendix~\ref{app:algorithm} for full algorithmic details.
  
  % \looseness=-1\textbf{Example.} We illustrate the DFS on the task \texttt{``Find a vegan chocolate brownie recipe with 4+ rating on allrecipes.com''} (see~\cref{fig:framework}). The root is expanded as an \AND{} node with three ordered children: search ($1.1$), select recipe ($1.2$), and note recipe ($1.3$). Node~$1.2$ is an \OR{} node with two ranked alternatives. The agent executes $1.1$ (search), then explores only the top-ranked child of $1.2$ (Recipe~A), which succeeds, so $1.2.2$ is never explored. Finally $1.3$ succeeds and the task completes. No failures occur in this trace; had $1.2$ exhausted both alternatives, failure would propagate upward and trigger \NODEREPAIR{} on the root. A full step-by-step trace appears in~Appendix~\ref{app:example_trace}.

 \looseness=-1\textbf{Example.} Consider the task \texttt{``Find a vegan
  chocolate brownie recipe with 4+ rating on allrecipes.com''}
  (see~\cref{fig:framework}). The agent begins with the root on the stack in the
  \ENTERING{} state. \NODEEXPANSION{} identifies it as an \AND{} node with three
  ordered children: search~($1.1$), select recipe~($1.2$), and note
  recipe~($1.3$). The root is then re-pushed in the \EXITING{} state and its
  children are pushed in the prescribed order. Node~$1.1$ then executes and
  succeeds. Node~$1.2$ expands into an \OR{} node with two ranked alternatives,
  but only the top-ranked child~$1.2.1$ (Recipe~A) is pushed. Once it succeeds,
  \NODECHECK{} confirms that the \OR{} objective is satisfied, so node~$1.2.2$ is
  never explored. \SUMMARIZER{} updates the internal context after each action,
  and \GLOBALTREE{} makes no modifications in this trace. After node~$1.3$
  succeeds, the root is popped in the \EXITING{} state, \NODECHECK{} confirms
  that all three \AND{} children succeeded, and the agent terminates with
  \SUCCESS{}. No node enters the \FAILED{} state here. Had node~$1.2$ exhausted
  both alternatives, the sibling node~$1.3$ would have been pruned and the failure would
  have propagated to the root, invoking \NODEREPAIR{} if the revision budget
  permitted. We provide the full step-by-step trace in
  Appendix~\ref{app:example_trace}.

\section{Related Work}\label{sec:relatedwork}

  \paragraph{LLM Inference-Based Web Agents:}
  Several
  works~\citep{putta2025agent,zhang2024webpilotversatileautonomousmultiagent,he-etal-2024-webvoyager,zhou2024languageagenttreesearch}
  leveraged closed-source LLMs or multimodal models for complex web tasks via
  inference-time strategies. \citet{wang2025agent} extracted reusable patterns
  from prior trajectories to guide navigation, while others employed Monte Carlo
  Tree Search~\citep{zhang2024webpilotversatileautonomousmultiagent} or value
  networks~\citep{koh-etal-2024-visualwebarena} for exploration and
  backtracking. Reflexion~\citep{shinn2023reflexionlanguageagentsverbal} added
  self-reflection, often at the cost of frequent resets. Hierarchical planning
  has also been explored in this setting.
  \Citet{erdogan2025planandactimprovingplanningagents} proposed an optimization
  layer for higher-level planning, and \citet{sodhi2024step} proposed dynamic
  multi-level control that relies on domain-specific prompts.
  \Citet{yang2025agentoccam} introduced a lightweight tree with text filtering
  and incremental revision, but their agent struggles on complex tasks due to
  limited error recovery. In contrast, our $\ANDOR$ planning tree supports
  robust error handling and explicit dependency tracking, and generates modular
  trajectories (see Table~\ref{tab:example1}).

 \paragraph{Fine-Tuned LLMs as Web Agents:}
  A complementary line of work fine-tunes open-weight models on expert
  trajectories via imitation or reinforcement
  learning~\citep{Lai2024AutoWebGLM,hong2024cogagentvisuallanguagemodel,zhang-etal-2024-android,murty2025nnetnavunsupervisedlearningbrowser,pareja2024unveilingsecretrecipeguide,prasad-etal-2024-adapt,Bai2024DigiRL,zhou2024archer,wang2024offlinereinforcementlearningllm}.
  These approaches face challenges such as brittleness under distribution shift
  and the cost of collecting expert demonstrations for imitation learning. We
  expand on this discussion in Appendix~\ref{app:relatedworks}.

  \section{Experiments}
  
  % \looseness=-1\textbf{Setup.} We evaluate on \textit{WebArena}~\citep{zhou2024webarenarealisticwebenvironment} (630 tasks across Map, Shopping, Reddit, and GitLab) and \textit{WebVoyager}~\citep{he-etal-2024-webvoyager} (129 tasks on live websites), augmented with 60 custom Amazon shopping tasks requiring multi-constraint product retrieval (10--30 steps). All experiments use Claude~3.5~Sonnet as the backbone unless noted otherwise. We compare against \ao{}~\citep{yang2025agentoccam}, which uses LLM-driven prune-and-branch planning, and \bca{}, a flat Claude agent conditioning on full action history. For WebArena we additionally include \sr{}~\citep{sodhi2024step} and the reference WebArena agent (\wrep{}). Task completion is evaluated with three LLM judges (GPT-4.1, Kimi-k2, Gemini-2.5-Flash) for WebVoyager and Amazon tasks and the official code-based evaluators for WebArena, with additional human annotation on a subset of Amazon tasks. Full details appear in~Appendix~\ref{app:setup}.

\looseness=-1\textbf{Setup.} We evaluate \SA{} on three benchmarks. The first
  is an Amazon dataset of $100$ tasks, which combines the $40$ Amazon tasks
  from WebVoyager with $60$ custom tasks that each impose multiple constraints.
  The second, WebVoyager-Other, consists of the remaining $89$ WebVoyager tasks.
  The third is WebArena, where we evaluate on $630$ tasks from the original
  benchmark and on WebArena-Verified, an audited release that reviews every
  task, reference answer, and evaluator to remove brittle string matching and
  ambiguous success criteria. All experiments use Claude~3.5~Sonnet as the
  backbone unless noted otherwise. We compare against
  \ao{}~\citep{yang2025agentoccam}, which uses LLM-driven prune-and-branch
  planning, and \bca{}, a flat agent that conditions on the full action history.
  On WebArena, we additionally include \sr{}~\citep{sodhi2024step} and the
  reference WebArena agent (\wrep{}). We measure task completion (success rate)
  using three LLM judges (GPT-4.1, Kimi-k2, Gemini-2.5-Flash) for WebVoyager and
  Amazon tasks and the official code-based evaluators for WebArena. We further
  collect human annotations on a subset of Amazon tasks. We provide full details
  of the setup in Appendix~\ref{app:setup}.

  \begin{table*}[t]
  \centering
  \small
  \resizebox{\linewidth}{!}{%
  \begin{tabular}{lcccccc}
  \toprule
  \textbf{Method} & \textbf{Map $\uparrow$} & \textbf{Shopping $\uparrow$} & \textbf{Reddit $\uparrow$} & \textbf{GitLab $\uparrow$} & \textbf{Overall $\uparrow$} \\
   & \textbf{(n=128)} & \textbf{(n=192)} & \textbf{(n=114)} & \textbf{(n=196)} & \textbf{(n=630)} \\
  \midrule
  \sa{}$^\dagger$       & 0.45 [0.37, 0.53] & \textbf{0.53 [0.46, 0.60]} & \textbf{0.73 [0.64, 0.80]} & 0.46 [0.39, 0.53] & \textbf{0.53 [0.49, 0.56]} \\
  \ao{} (Claude)        & \textbf{0.46 [0.38, 0.55]} & 0.38 [0.31, 0.45] & 0.60 [0.51, 0.69] & \textbf{0.47 [0.40, 0.54]} & 0.46 [0.43, 0.50] \\
  \ao{} (GPT-4o)        & 0.45 [0.36, 0.53] & 0.35 [0.29, 0.42] & 0.62 [0.54, 0.71] & 0.33 [0.26, 0.40] & 0.41 [0.38, 0.45] \\
  \sr{} (GPT-4o)        & 0.33 [0.26, 0.41] & 0.27 [0.21, 0.33] & 0.56 [0.47, 0.65] & 0.22 [0.17, 0.28] & 0.32 [0.29, 0.36] \\
  \bca{}                & 0.37 [0.29, 0.45] & 0.30 [0.24, 0.36] & 0.30 [0.22, 0.39] & 0.28 [0.22, 0.34] & 0.31 [0.28, 0.34] \\
  \wrep{}               & 0.31 [0.24, 0.39] & 0.31 [0.25, 0.38] & 0.20 [0.13, 0.27] & 0.29 [0.23, 0.35] & 0.28 [0.25, 0.32] \\
  \bottomrule
  \end{tabular}}
  \caption{Success rates on WebArena across four task categories, as mean [95\% bootstrap CI]. \textbf{Bold} is best per column. \newline $^\dagger$~$p < 0.05$ vs.\ all methods (one-sided permutation test).}
  \label{res:webarena_results}
  \end{table*}

 \begin{table*}[t]
  \centering
  \begin{minipage}[t]{0.57\textwidth}
  \centering
  \footnotesize
  \setlength{\tabcolsep}{3pt}
  \renewcommand{\arraystretch}{1}
  \resizebox{\linewidth}{!}{%
  \begin{tabular}{ll|ccc|c|ccc|c}
  \toprule
  & & \multicolumn{4}{c|}{\textbf{Amazon}} & \multicolumn{4}{c}{\textbf{WebVoyager Other}} \\
  \cmidrule(lr){3-6} \cmidrule(lr){7-10}
  \textbf{Backbone} & \textbf{Method} & kimi & Gemini & GPT & Avg & kimi & Gemini & GPT & Avg \\
  \midrule
  \multirow{4}{*}{\textbf{C3.5}}
  & \sa{}           & \textbf{62.0} & \textbf{47.0} & 55.0 & \textbf{54.7} & 80.9 & \textbf{73.0} & 80.9 & 78.3 \\
  & \ao{}           & 52.0 & 35.0 & 47.0 & 44.7 & \textbf{84.3} & 71.9 & \textbf{83.2} & \textbf{79.8} \\
  & \bca{}          & 48.0 & 35.0 & 45.0 & 42.7 & 74.2 & 67.4 & 70.8 & 70.8 \\
  & \sr{}           & 33.0 & 29.0 & 33.0 & 31.6 & 74.2 & 60.7 & 69.7 & 68.2 \\
  \midrule
  \multirow{3}{*}{\textbf{C3.7}}
  & \sa{}           & \textbf{56.7} & \textbf{36.7} & \textbf{46.7} & \textbf{46.7} & \multicolumn{4}{c}{\multirow{3}{*}{\textit{not evaluated}}} \\
  & \ao{}           & 45.0 & 30.0 & 41.7 & 38.9 & \multicolumn{4}{c}{} \\
  & \bca{}          & 28.3 & 21.7 & 20.0 & 23.3 & \multicolumn{4}{c}{} \\
  \bottomrule
  \end{tabular}%
  }
  \captionof{table}{Success rates (\%) on Amazon shopping tasks and WebVoyager Other under three LLM judges (Kimi-K2, Gemini-2.5-Flash, GPT-4.1), with Claude~3.5 (C3.5) and Claude~3.7 (C3.7) backbones. \SA{} leads on Amazon by 10 points (C3.5) and 7.8 points (C3.7); on WebVoyager Other, \SA{} and \ao{} fall within 1.5 points (78--80\%).}
  \label{res:combined}
  \label{res:claude37_amazon}
  \end{minipage}
  \hfill
  \begin{minipage}[t]{0.33\textwidth}
  \centering
  \footnotesize
  \setlength{\tabcolsep}{3pt}
  \resizebox{\ifdim\width>\linewidth \linewidth\else\width\fi}{!}{%
  \begin{tabular}{lccc}
  \toprule
  \textbf{Site (n)} & \textbf{AO} & \textbf{SA} & \textbf{$p$} \\
  \midrule
  Shopping (56) & 35.7 & \textbf{39.3} & .422 \\
  Shop.\ Admin (53) & 11.3 & \textbf{15.1} & .386 \\
  GitLab (56) & 30.4 & \textbf{51.8} & .016$^*$ \\
  Reddit (42) & 66.7 & \textbf{76.2} & .235 \\
  Cross-site (48) & 0.0 & 0.0 & 1.00 \\
  \midrule
  \textbf{Overall (255)} & 27.8 & \textbf{35.7} & \textbf{.037}$^*$ \\
  \bottomrule
  \end{tabular}}
  \captionof{table}{WebArena-Verified Hard. \SA{} leads \ao{} by 7.8 points overall on 255 tasks ($p = 0.037$).}
  \label{tab:wa_hard}
  \end{minipage}
  \end{table*}

  \begin{figure*}[t]
  \centering
  \includegraphics[width=\textwidth]{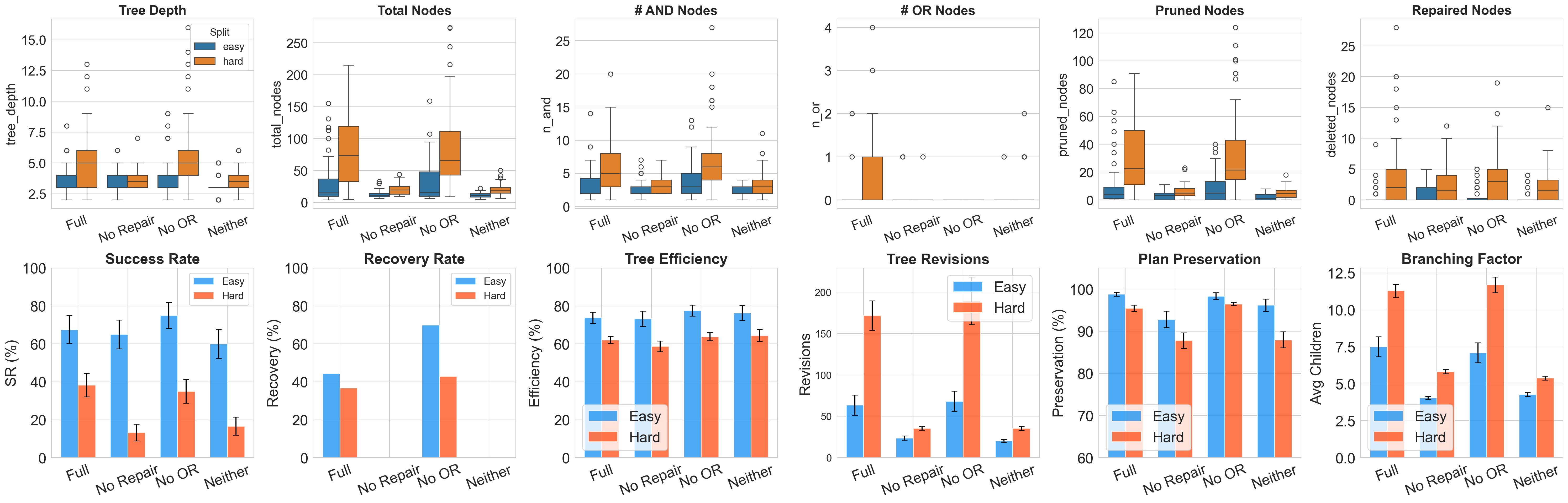}
  \caption{Ablation analysis on Amazon tasks. Top: tree structure metrics (depth, nodes, pruning). Bottom: success, repair trigger, recovery rate, and revisions. The full agent (Repair+OR) builds deeper trees and achieves non-zero recovery on hard tasks.}
  \label{fig:ablation_stats}
  \end{figure*}
  
  % \looseness=-1\textbf{Summary of key findings.} \SA{} achieves the highest overall performance on both benchmarks, with gains of +14\% on Amazon Easy, +7--9\% on Amazon Hard, and +6\% on WebArena overall compared
  % to the strongest baseline (\ao{}). The largest improvements occur on complex, multi-step tasks requiring error recovery and constraint tracking (Shopping +14.8\%, Reddit +12.5\%), validating that hierarchical
  % planning provides the greatest benefit where it is most needed.
  
  \subsection{Results}
  
  % \noindent\looseness=-1\textbf{RQ1) How does \SA{} compare with LLM-based agents on short-horizon and complex long-horizon tasks?}
  % \Cref{res:combined} reports success rates on Amazon and Weboyager Other tasks using Claude 3.5 as the backbone. \SA{} achieves the highest average performance on Amazon tasks, outperforming \ao{} by 10\%. On
  % Weboyager Other, all methods perform within 1.5\% of each other (78--80\%), confirming that these short-horizon tasks do not discriminate between planning approaches; the value of hierarchical planning emerges
  % on harder tasks.
  
  % On the human-annotated subset of Amazon tasks, we observe higher success rates under human evaluation than under LLM-as-a-Judge. These tasks require the agent to retrieve information about multiple items while satisfying several
  % simultaneous constraints, and LLM judges penalize partial satisfaction more harshly than human annotators. Nevertheless, both protocols agree on the relative ordering of agents, and the three LLM judges agree with one another, indicating the observed trends are robust to the choice of evaluator.
  
  % \looseness=-1\samem{} helps most on complex Amazon tasks that require tracking multiple candidate solutions, but slightly degrades performance on simpler tasks, where the structured tabular format generates overly strict intermediate constraints that act as noise. It therefore works best for complex information-seeking tasks requiring the agent to simultaneously track multiple items
  % and satisfy diverse constraints.

 \noindent\looseness=-1\textbf{RQ1) How does \SA{} compare with LLM-based agents on short-horizon and complex long-horizon tasks?}
  \Cref{res:combined} reports success rates on Amazon and WebVoyager-Other tasks using Claude~3.5~Sonnet as the backbone. \SA{} achieves the highest average performance on Amazon tasks, outperforming \ao{} by $10$
   points. On WebVoyager-Other, \SA{} and \ao{} fall within $1.5$ points of each other ($78$--$80\%$), suggesting that these short-horizon tasks do not discriminate between planning approaches. The value of
  hierarchical planning emerges only on the harder Amazon tasks, which require tracking multiple candidate items under several simultaneous constraints.
  
  % On the human-annotated subset of Amazon tasks, human evaluation and LLM-as-a-Judge agree on the relative ordering of agents, and the three LLM judges agree with one another, indicating that the observed trends
  % are robust to the choice of evaluator.

\begin{figure*}
  \centering
  \begin{subfigure}[b]{0.45\linewidth}
      \centering
      \includegraphics[width=\textwidth]{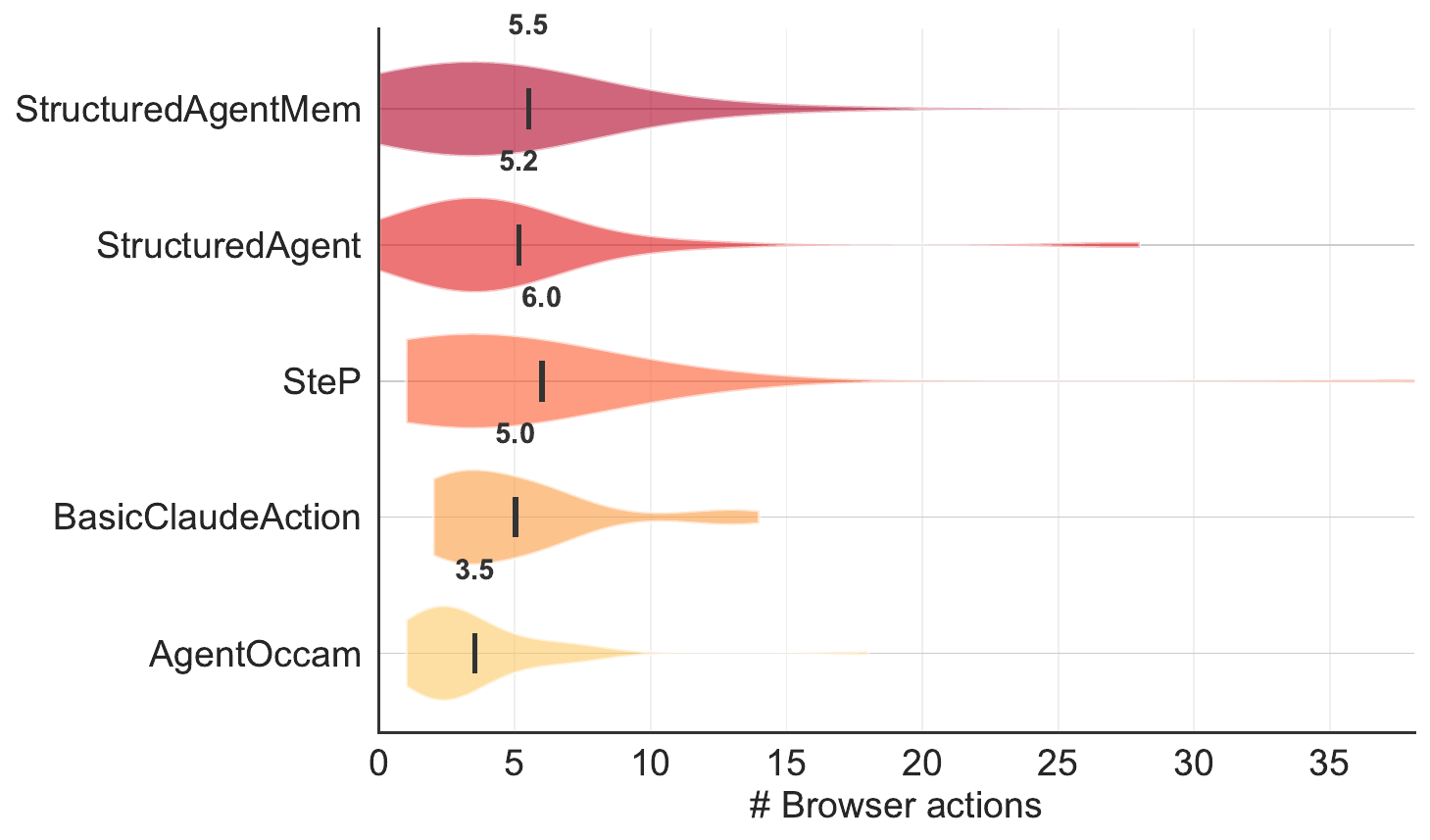}
      \caption{Amazon-easy}
      \label{fig:traj_easy}
  \end{subfigure}
  \hfill
  \begin{subfigure}[b]{0.35\linewidth}
      \centering
      \includegraphics[width=\textwidth]{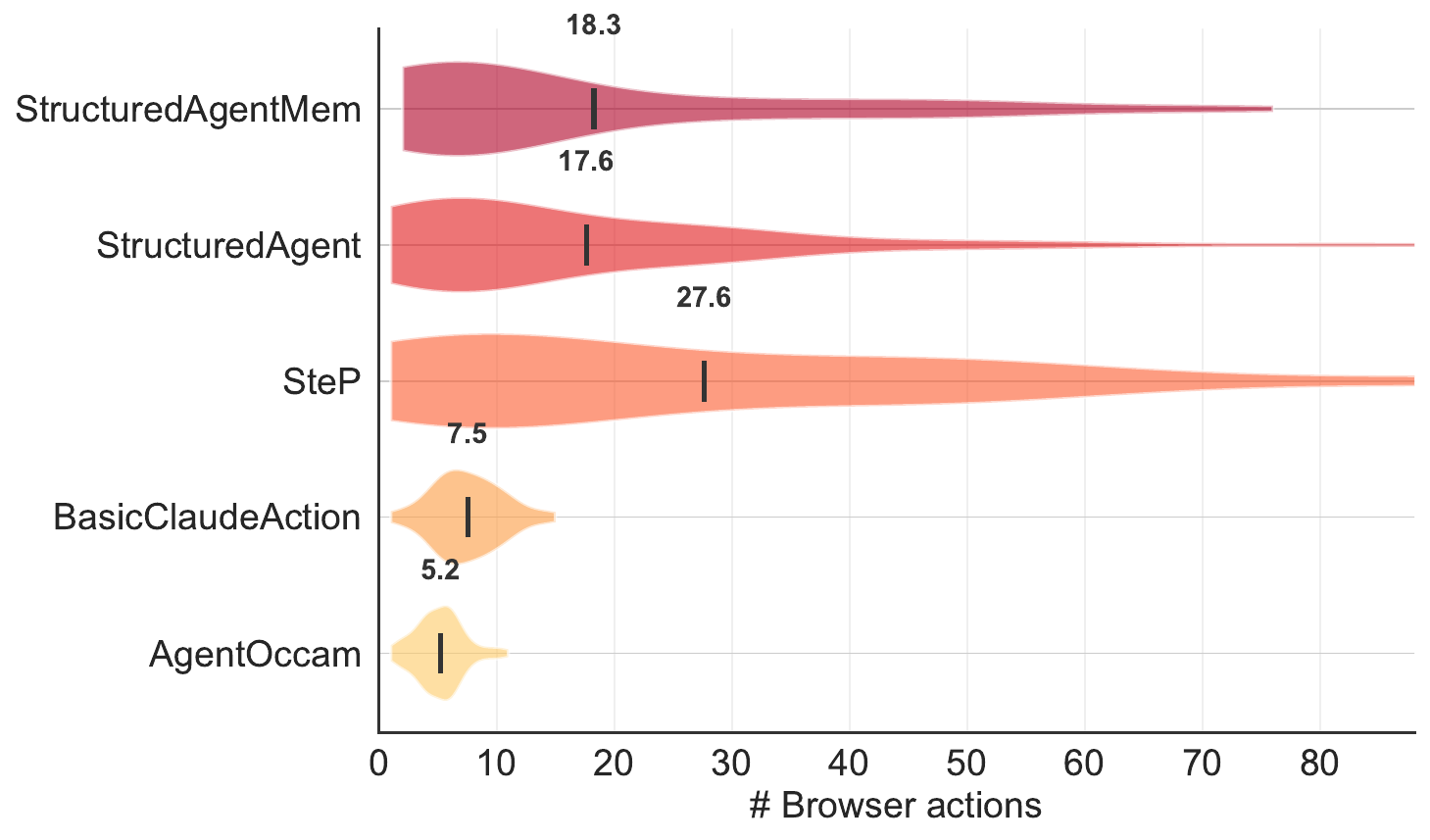}
      \caption{Amazon-hard}
      \label{fig:traj_hard}
  \end{subfigure}
  \vspace{-0.05in}
  \caption{\looseness=-1 Trajectory length distributions on Amazon tasks. On hard tasks, \SA{} explores more strategies before terminating, leading to higher success rate, while baselines terminate prematurely with shorter but less successful trajectories.}
  \label{fig:traj_amazon}
  \end{figure*}
  % \begin{table}[h!]
  % \centering
  % \small
  % \setlength{\tabcolsep}{0.9pt}
  % \renewcommand{\arraystretch}{0.8}
  % \begin{tabular}{lccccc}
  % \toprule
  % \multirow{2}{*}{\textbf{Method}} & {Map} & {Shopping} & {Reddit} & {GitLab} & {Overall} \\
  % & ($\uparrow$) & ($\uparrow$) & ($\uparrow$) & ($\uparrow$) & ($\uparrow$) \\
  % \midrule
  % \textbf{\sa{}} & 0.450 & \textbf{0.528} & \textbf{0.726} & 0.458 & \textbf{0.526} \\
  % \midrule
  % \ao{} & \textbf{0.463} & 0.380 & 0.601 & \textbf{0.467} & 0.464 \\
  % \bca{} & 0.370 & 0.303 & 0.303 & 0.281 & 0.310 \\
  % \wrep{} & 0.315 & 0.313 & 0.197 & 0.288 & 0.285 \\
  % \aogpt{} & 0.446 & 0.352 & 0.623 & 0.329 & 0.413 \\
  % \srgpt{} & 0.333 & 0.269 & 0.564 & 0.225 & 0.321 \\
  % \bottomrule
  % \end{tabular}
  % \caption{Success rates on WebArena across four task categories: Map (128), Shopping (192), Reddit (114), and GitLab (196). All methods use Claude 3.5 Sonnet unless noted (GPT suffix indicates GPT-4 backbone).
  % \SA{} achieves the highest overall score (52.6\%), with the largest gains on Shopping (+14.8\%) and Reddit (+12.5\%) over \ao{}.}
  % \label{res:webarena_results}
  % \end{table}
  
  \looseness=-1 \Cref{res:webarena_results} reports results across WebArena task categories. \SA{} achieves the highest overall performance (53\%), surpassing \ao{} by 7\%, with the largest gains on Shopping
  (+15\%) and Reddit (+13\%), the categories with the most complex multi-step workflows. On Map and GitLab, which involve simpler navigation patterns, \SA{} performs comparably to \ao{}. Hierarchical planning helps most where task complexity is highest.
  
  \noindent\looseness=-1\textbf{RQ2) Does \SA{} continue to benefit from hierarchical planning with a stronger backbone?} \Cref{res:claude37_amazon} compares performance on Amazon tasks using Claude 3.7. Despite Claude 3.7 being
   explicitly trained on agentic tasks, \SA{} still outperforms \ao{} by 7.8\% and \bca{} by 23.4\%, demonstrating that the benefits of hierarchical planning persist with stronger models.
  
  \noindent\looseness=-1\textbf{RQ3) Does \SA{} generalize to model families beyond Claude?} \Cref{res:webarena_kimi_results} evaluates \SA{} and \ao{} using Kimi-k2-0905 as the backbone on a subset of WebArena
  tasks. \SA{} achieves overall performance comparable to \ao{} but does not outperform it, leading only on the Shopping category. We attribute this to the weaker reasoning and planning capabilities of Kimi-k2. \SA{}'s gains depend on the backbone's ability to produce good decompositions and rank alternatives reliably, and Kimi-k2 is weaker at both, so the tree provides less useful guidance than with Claude.

 \noindent\looseness=-1\textbf{RQ4) How does \SA{} compare in terms of trajectory length?}
 We compare \SA{} and \ao{}'s trajectory lengths on 40 Amazon tasks from WebVoyager, which
 we label \emph{Amazon-easy}, and 60 Amazon tasks that we generate, each combining multiple constraints,
 which we label \emph{Amazon-hard} (\Cref{fig:traj_amazon}). \SA{} takes a comparable number
 of actions to \ao{} on Amazon-easy, while on Amazon-hard it produces markedly longer
 trajectories as it explores multiple strategies before terminating. This
systematic exploration is precisely what drives its higher success rate. On WebArena, \SA{} matches
 \ao{} on Reddit and Shopping while achieving substantially higher success rates, and takes
 moderately more actions on Map and GitLab, reflecting increased exploration once
 straightforward strategies fail. We report per-category WebArena distributions in Appendix~\ref{app:experimentalresults}
 (Figure~\ref{res:trajectory_lengths_webarena}) and demonstrate the exploration behavior
 with a qualitative example in~\Cref{tab:trajectory}.

  \noindent\looseness=-1\textbf{RQ5) How does \SA{} perform on WebArena-Verified Hard (a curated difficult subset)?} We evaluate on WebArena-Verified Hard~\citep{hattami2025webarena}, a 255-task subset selected via difficulty modeling and equipped with deterministic, human-verified evaluators that eliminate false positives and negatives. \Cref{tab:wa_hard} shows \SA{} outperforms \ao{} by 21 points on GitLab and 10 points on Reddit, the two categories with the most complex multi-step workflows, and by 7.8 points overall ($p = 0.037$). Hierarchical planning provides the largest benefit on genuinely hard tasks.

 \noindent\looseness=-1\textbf{RQ6) Which components of \SA{} drive the
  performance gains?} To isolate the contribution of each component, we disable
  the two adaptive mechanisms of the \ANDOR{} planner, \NODEREPAIR{} and \OR{}
  alternatives, individually and jointly on the Amazon-easy ($40$ tasks) and
  Amazon-hard ($60$ tasks) splits with a Haiku~4.5 backbone. This yields four
  conditions. Full enables both mechanisms, Repair only retains \NODEREPAIR{}
  but disables \OR{} alternatives, OR only retains \OR{} alternatives but
  disables repair, and Neither disables both, reducing the planner to a static
  tree that can neither recover from failed nodes nor fall back to alternative
  paths. Alongside success rate, we track tree statistics that capture plan
  structure, adaptation, and recovery (full definitions in
  Appendix~\ref{app:ablation_detail}). 

Results show that \NODEREPAIR{} drives
  recovery, which we measure as the fraction of repair attempts that restore a
  failed node to a passing state. OR only and Neither cannot recover failed
  nodes. With repair enabled, the Full planner recovers $44\%$ of failed nodes
  on Amazon-easy and $37\%$ on Amazon-hard, and Repair only reaches $70\%$ and
  $43\%$ respectively. Repair also enables deeper exploration. On Amazon-hard,
  the Full planner builds trees of $80$ nodes at depth $5.3$ with $172$
  revisions (expansions, repairs, and prunings), compared to $20$ nodes at
  depth $3.6$ without repair. At the same time, repair modifies only a small
  part of the plan. Plan Preservation, the fraction of nodes left intact by
  repair, stays between $93$ and $99\%$ on Amazon-easy and between $88$ and
  $96\%$ on Amazon-hard, so recovery reuses most of the existing plan rather
  than replanning from scratch. \OR{} nodes are exercised rarely on these tasks
  (mean count at most $0.6$ across conditions), so node-level repair carries
  most of the recovery behavior here. Consistent with this, enabling \OR{}
  alternatives without repair provides no reliable benefit over disabling both
  mechanisms. On Amazon-hard, OR only scores $13.3\%$ versus $16.7\%$ for
  Neither under the Gemini judge, and the two conditions perform comparably on
  Amazon-easy. Without repair, the additional branching goes unexploited, since
  failed alternatives cannot be restored. \OR{} alternatives matter more when
  the agent must explore and track multiple candidate paths before arriving at
  the correct solution. We report per-condition tree statistics in
  \Cref{tab:tree-analysis-easy,tab:tree-analysis-hard}.

 \noindent\looseness=-1 \textbf{Additional analyses.} Due to space constraints, we defer to the appendix our study of agreement among the judges and with a human annotator (\Cref{tab:human_eval}). \Cref{tab:diagnosis_quality_main} summarizes how the framework helps identify the cause of a failure.

\begin{table}[t]
\centering
\scriptsize
\caption{\SA{} vs.\ \ao{} diagnosis quality on the 15 shared failures; cells give \% favoring \SA{} (raw SA/AO/Tie). Details in \Cref{tab:diagnosis_quality}.}
\label{tab:diagnosis_quality_main}
\resizebox{\ifdim\width>\columnwidth \columnwidth\else\width\fi}{!}{%
\begin{tabular}{lcc}
\toprule
Dimension & Gemini 2.5 Flash & Opus 4.8 \\
\midrule
Precision & 86.8 (203/31/3) & 94.7 (160/9/70) \\
Self sufficiency & 64.3 (9/5/223) & 72.4 (155/59/25) \\
Clarity & 63.2 (91/53/93) & 76.9 (183/55/1) \\
Actionability & 78.3 (162/45/30) & 67.6 (148/71/20) \\
\midrule
Overall preference & 81.7 (192/43/2) & 75.2 (170/56/13) \\
\bottomrule
\end{tabular}}
\end{table}

  \section{Conclusion}
  
  \looseness=-1 We presented \SA{}, a hierarchical planning framework that decomposes complex web tasks into dynamic $\ANDOR$ trees, separating structural planning from LLM-based reasoning. The framework enables principled
  error recovery via node repair, systematic exploration of alternatives via OR nodes, and interpretable plans that support human intervention. On WebArena (630 tasks) and WebVoyager augmented with 60 complex
  shopping tasks, \SA{} consistently outperforms strong baselines across multiple model families, with gains of 7--10\% over \ao{} on complex tasks. Ablations confirm that both repair and OR-node mechanisms
  contribute complementary benefits.

\clearpage
% Additional references included in the bibliography without inline discussion.
\nocite{shahnovsky2026aiplanningframeworkllmbased,aghzal2026llmbasedwebagentsfail,vardanyan2025buildingbrowseragentsarchitecture,anupam2026browserarena,hattami2025webarena,lugoloobi2026knownactionsfingerprintingllm,anupam2025browserarenaevaluatingllmagents,yu2025browseragentbuildingwebagents,erdogan2025planandactimprovingplanningagents}
\bibliography{neurips}
% \bibliographystyle{plainnat}

% ============================================================================
% SUPPLEMENTARY MATERIAL (merged; starts on a new page). Strip these pages
% from the final main-paper PDF if submitting separately.
% ============================================================================
\clearpage
\begin{center}
  {\LARGE\bfseries \SA{}: Planning with AND/OR Trees for Long-Horizon Web Tasks}\\[0.6em]
  {\Large Supplementary Material}
\end{center}
\vspace{1.5em}
\appendix
% Number appendix sections (A, B, ...) so that \ref{app:...} resolves correctly.
\setcounter{secnumdepth}{3}

\section{Extended Related Work}\label{app:relatedworks}

We provide an extended discussion of fine-tuned LLMs as web agents, complementing the Related Work section in the main paper.

 \paragraph{Fine-Tuned LLMs as Web Agents:}
  Several works employ open-source language models as web
  agents~\citep{hong2024cogagentvisuallanguagemodel,Lai2024AutoWebGLM,zhang-etal-2024-android}.
  These models often fail because they are small and lack domain-specific
  knowledge. Follow-up work addresses this by fine-tuning open-source models on
  expert trajectories collected from human demonstrators or proprietary
  LLMs~\citep{murty2025nnetnavunsupervisedlearningbrowser,pareja2024unveilingsecretrecipeguide,prasad-etal-2024-adapt}.
  Fine-tuning yields meaningful gains, particularly when the evaluation tasks
  come from the same distribution as the training data.
  
  Training in this line of work follows one of two paradigms. Imitation learning
  trains agents offline on pre-collected expert
  demonstrations~\citep{zare2023surveyimitationlearningalgorithms}, while
  reinforcement learning (RL) improves agents through direct interaction with
  the environment~\citep{Bai2024DigiRL,zhou2024archer,wang2024offlinereinforcementlearningllm}.
  Imitation learning maximizes the likelihood of expert actions, which leaves
  agents brittle on complex or out-of-distribution
  tasks~\citep{simchowitz2025pitfallsimitationlearningactions}. Its success also
  depends on the quality and breadth of the demonstration data, which is costly
  to obtain, especially when proprietary models generate the trajectories.
  
  RL offers a path toward more adaptive agents but brings its own
  challenges~\citep{levine2020offlinereinforcementlearningtutorial}. Online RL
  for web agents is computationally demanding, and the reward is typically a
  sparse binary signal tied to overall task completion. This sparsity creates a
  credit assignment problem~\citep{Nachum2017PCL}, since a failed task can
  penalize intermediate actions that were in fact correct. Fine-tuning capable
  and generalizable web agents with high-quality trajectories and dense reward
  signals therefore remains underexplored.
  
\section{Experimental Details}\label{app:experimentalresults}

\subsection{Experimental Setup}\label{app:setup}
\textbf{Benchmark Details.}
  \textit{Custom Complex Shopping} contains 60 long-horizon Amazon shopping tasks. Each task is described by a natural language query that encodes detailed constraints over one or more products, and the agent must
  produce a set of recommendations satisfying these constraints. Task completion typically requires 10 to 30 interaction steps, reflecting the compositional and long-horizon nature of the goals.
  \textit{WebVoyager}~\citep{he-etal-2024-webvoyager} provides real-world tasks on diverse live websites (\eg BBC News, arXiv, Amazon, GitHub, and HuggingFace). We use a curated subset of 129 tasks. We exclude
  tasks that require login credentials or otherwise raise security concerns, since the agent cannot reliably bypass them in our evaluation setting.
  \textit{WebArena}~\citep{zhou2024webarenarealisticwebenvironment} contains 812 tasks, of which we evaluate on 630. We exclude tasks on the \texttt{ShoppingAdmin} site as well as tasks where agents consistently
  fail due to website errors or evaluator failures, which prevent reliable comparison.
  
  \textbf{Evaluation Details.} For WebVoyager, the LLM judge first decomposes the task into item-level and task-level constraints and then awards one point per satisfied constraint. A constraint counts as
  satisfied only when its satisfaction is explicitly supported by evidence extracted from the raw observations (\Cref{tab:evidence_extractor}, with the full prompt in \cref{app:prompts}). Prior works evaluate
  agents from the last $N$ observations using a GPT-4V judge. This approach is insufficient for long-horizon tasks, where the required $N$ can be very large, and the prompt used is overly simplistic. For WebArena,
  we use the benchmark's official code-based (HTML- and URL-based) evaluators without modification.
  
  \textbf{Baseline Details.} We compare against four baselines. \ao{}~\citep{yang2025agentoccam} is a planning agent that incrementally constructs and prunes an action tree. \bca{} augments a Claude-based agent
  with access to the full action history as additional context. \sr{}~\citep{sodhi2024step} is a stack-based agent that uses human-written atomic strategies designed for WebArena-style tasks. \wrep{} is the
  reference agent released with WebArena, prompted to perform chain-of-thought reasoning. We use the implementations of \sr{} and \wrep{} provided by~\citet{yang2025agentoccam}.
  
  \textbf{Decoding Configuration.} \Cref{tab:decoding_config} summarizes the decoding settings used for each model backend.
  
  \begin{table}[t]
  \centering
  \small
  \caption{Decoding configuration by model backend. The Anthropic (Bedrock), Novita, and Gemini APIs do not accept a request seed, so determinism on those paths follows from greedy decoding. A request seed (fixed
  to $0$) is honored only on the OpenAI path.}
  \label{tab:decoding_config}
  \resizebox{\ifdim\width>\columnwidth \columnwidth\else\width\fi}{!}{%
  \begin{tabular}{l|ccc}
  \toprule
  \textbf{Backend} & \textbf{Temp.} & \textbf{top-$p$} & \textbf{Seed} \\
  \midrule
  Claude (Bedrock \texttt{invoke\_model})   & 0.0 & NA   & NA \\
  Bedrock Converse (Kimi/Qwen/Nova/\ldots)  & 0.0 & NA   & NA \\
  OpenAI GPT                                & 0.0 & 0.01 & 0 \\
  Kimi (Novita)                             & 0.1 & 0.95 & NA \\
  Gemini                                    & 0.0 & NA   & NA \\
  \bottomrule
  \end{tabular}%
  }
  \end{table}
  
  \textbf{Statistical Inference.} For WebArena-Verified Hard we report results on the 255 valid-shared tasks. Confidence intervals use the percentile bootstrap. For each site and for the overall score, we resample
  the binary per-task scores with replacement ($B = 10{,}000$ resamples, each of size equal to the number of tasks), take the mean of each resample, and report the 2.5th and 97.5th percentiles as the 95\% CI.
  Significance uses a one-sided permutation test of $\text{SA} > \text{AO}$. We pool both agents' per-task scores, randomly reassign the agent labels $10{,}000$ times, and report as $p$ the fraction of permuted
  mean differences that meet or exceed the observed difference. Both procedures use a fixed random seed of $42$.

% \begin{figure}
%     \centering
%     \includegraphics[width=0.5\linewidth]{steps.pdf}
%   \caption{Average number of steps taken by the agents on Complex Shopping tasks using \textit{Claude 3.5} as the backbone model.}

%     \label{fig:steps}
% \end{figure}

\section*{Amazon Shopping Tasks}

% \subsection*{Medium-Level Tasks (600--610)}

\begin{itemize}
  \item Recommend 3 standing desks under \$350 that support lift ranges from $\leq 24"$ to $\geq 48"$ and include a cable management tray. Each recommendation should differ in brand or surface finish. Do not add to cart--list lift range, tray type, and price.
  \item Recommend 3 pressure cookers under \$100 with stainless steel inner pots and yogurt mode. Each must be from a different brand. Do not add to cart--list material, features, rating, and price.
  \item Recommend 3 wireless printers under \$150 that support both AirPrint and borderless 4x6 photo printing. Do not add to cart--list brand, print support, and price.
  \item Recommend 3 mechanical keyboards under \$400 with hot-swappable keys. Ensure variety in switch type or layout (e.g., 75\%, TKL). Do not add to cart--list hot-swap support, switch type, and price.
  \item Recommend 3 laptop backpacks under \$200 that support 17-inch laptops, include a padded sleeve, and have a hidden anti-theft pocket. Ensure variation in material, port support, or design. Do not add to cart--list capacity, features, and price.
  \item Recommend 3 noise-canceling headphones under \$300 with multi-device pairing and passive listening support. Ensure brand diversity. Do not add to cart--list battery specs, pairing support, and price.
  \item Recommend 3 air purifiers under \$350 with washable pre-filters and a night mode that disables display lights. Do not add to cart--list pre-filter type, noise level, and price.
  \item Recommend 3 external SSDs (1TB) under \$120 that use USB-C and offer hardware encryption. Ensure diversity in brand or ruggedness. Do not add to cart--list encryption support, interface, and price.
  \item Recommend 3 espresso machines under \$450 that include a milk frother and removable water tank. Ensure different frother types or build designs. Do not add to cart--list milk system, tank size, and price.
  \item Recommend 3 portable projectors under \$300 with 1080p native resolution, tripod screw mount, and built-in Bluetooth speakers. Ensure model diversity. Do not add to cart--list resolution, mounting, audio features, and price.

  \item Find the cheapest projector and screen set under \$450: (1) portable projector with native 1080p, Bluetooth 5.0+, keystone correction, and 300+ ANSI lumens; (2) 70--100 inch outdoor screen. Recommend 3 combinations varying by projector brand. Choose the set with the highest total reviews. List model names, Bluetooth version, brightness, and total price.
  \item Find the cheapest pair under \$350 that includes: (1) over-ear wireless ANC headphones with 40h+ battery life and USB-C charging; (2) premium Bluetooth 5.0+ speaker with stereo pairing and IPX5 water resistance. Both must be rated 4.3+ stars. Recommend 3 brand-distinct combos. List battery specs, pairing and water-resistance features, rating, and total cost.
  \item Recommend 3 lightweight work kits under \$400 with fast delivery: (1) adjustable aluminum laptop stand supporting 10kg+; (2) rechargeable wireless keyboard + mouse; (3) USB-C hub with HDMI, SD, and Ethernet. Choose the combo with the fastest shipping (2 days or less). List ports, delivery times, and prices.
  \item Under \$450, find the most-reviewed premium travel set: (1) 17 inch anti-theft backpack with TSA lock; (2) 8-piece compression packing cube set made from water-resistant fabric; (3) digital luggage scale with auto-off, tare, and backlight. All rated 4.3+ stars. Recommend most-reviewed option. List features, review counts, and total price.
  \item Recommend the cheapest high-end ergonomic combo under \$900: (1) dual-motor standing desk (dual 275+ lb capacity, cable tray, programmable presets); (2) office chair with adjustable headrest, and lumbar support. Both must be rated 4.5+ stars and have 500+ reviews. List brands, specs, and combined price.
  \item Find the fastest-delivery smart home bundle under \$500: (1) smart speaker with built-in voice assistant and premium audio; (2) smart bulb pack with color plus tuneable white and 10K+ hour lifespan; (3) smart plug with energy monitoring and USB port. All must be rated 4.5+ stars and offer two-day delivery. List assistant, bulb specs, plug features, delivery ETA, and price.
  \item Recommend 3 pro-level chef’s prep kits under \$400: (1) 8" stainless chef’s knife with full tang and thermo-transfer handle; (2) digital food thermometer with $\pm 0.1^\circ$C accuracy, fast-read (<5s), and auto-off; (3) bamboo cutting board set with juice groove and non-slip feet. Choose highest-rated combination. List specs, ratings, and price.
  \item Find the most-reviewed outdoor adventure set under \$500 each: (1) waterproof hiking backpack (30--40L) with rain cover; (2) rechargeable headlamp (500+ lumens) with adjustable beam; (3) vacuum-insulated stainless steel bottle (32 oz) with sweat-proof exterior. All rated 4.5+ stars. List capacities, feature specs, review counts, and total price.
  \item Recommend 3 advanced artist kits under \$250: (1) LED desk lamp with CRI$>90$ and adjustable 3000K--6500K color; (2) sketchbook A4 with 200gsm acid-free paper; (3) graphite pencil set including 2H--8B and charcoal. Kits must differ by lamp or sketchbook brand. List specs, brand, review counts, and price.
  \item Find the fastest-delivery premium pet care bundle under \$250: (1) hands-free dog leash with reflective stitching, padded waist belt, and shock absorber; (2) collapsible BPA-free silicone travel bowl; (3) no-pull harness with five-point control and breathable mesh. All rated 4.5+ stars with two-day delivery. List materials, feature highlights, shipping ETA, review counts, and combined price.

  \item Recommend 2 cordless stick vacuums under \$200 with detachable batteries, HEPA filtration, at least 25 minutes runtime, and listed runtime and battery replacement ease.
  \item Find 3 premium gaming chairs under \$300 with adjustable lumbar support, at least 150-degree recline, verified user weight support of at least 300 lbs, and listed weight capacity and user review rating.
  \item List 2 portable espresso makers under \$100 compatible with Nespresso capsules, at least 18 bar pressure, BPA-free certification, and clearly state compatibility, pressure, and BPA-free status. If criteria aren't met, explain the closest match.
  \item Recommend 3 noise-cancelling earbuds under \$120 with at least 24-hour total playtime, IPX5 or higher waterproof rating, transparency mode, and clearly state sound quality ratings and battery life duration.
  \item Find 2 smart thermostats under \$150 compatible with Alexa and Google Assistant, supporting multi-zone control, energy-saving certification, and clearly state assistant compatibility, multi-zone capability, and certification details. If unavailable, suggest best alternatives with limitations.
  \item List 3 robot vacuums under \$250 with mapping capability, no-go zones, voice assistant support, runtime, and bin capacity. Identify models that fulfill most criteria if any feature is missing.
  \item Recommend 2 air fryers under \$100 with at least five-quart capacity, dishwasher-safe basket, preset cooking functions, and listed cooking presets and ease of cleaning. Highlight trade-offs between price, capacity, and features if necessary.
  \item Find 3 fitness trackers under \$80 with continuous SpO2 monitoring, swim-proof rating of at least 50 meters, sleep-tracking, battery life, and additional health monitoring features clearly listed.
  \item List 2 electric toothbrushes under \$60 with pressure sensors, at least 30-day battery life, ADA acceptance, charging type, and brush head replacement availability. Recommend based on user reviews and clinical backing if ADA acceptance is missing.
  \item Recommend 3 webcam models under \$70 with autofocus, at least 1080p resolution, 60 frames per second streaming, built-in privacy cover, and microphone quality. Clearly identify compromises on frame rate or privacy cover if necessary.
  \item Recommend 2 wireless routers under \$200 with Wi-Fi 6 support, dual-band, gigabit Ethernet ports, and listed max coverage area. If no routers meet all, choose closest and note missing attribute.
  \item Find 3 kitchen stand mixers under \$300 with at least 10 speeds, 5-quart bowl, and metal construction. List speed count, bowl size, and material. If fewer than three, include models that miss one requirement and indicate which.
  \item List 2 Bluetooth speakers under \$150 with waterproof rating of IPX7, at least 12-hour battery life, and built-in voice assistant support. State battery life and assistant type. If criteria aren’t met, explain closest fit.
  \item Recommend 3 DSLR-style mirrorless cameras under \$700 with interchangeable lenses, 4K video, and in-body image stabilization. Clearly state sensor resolution, video spec, and stabilization type. If none, suggest best trade-offs.
  \item Find 2 cordless electric lawn mowers under \$400 with at least 45 minute runtime, 20-inch deck, and mulching capability. List runtime, deck size, and whether mulching kit included.
  \item List 3 external SSDs under \$150 with USB-C connection, at least 1 TB capacity, and read speed over 1000 MB/s. Provide capacity, interface, and read speed. If fewer than three, note closest specs.
  \item Recommend 2 smartwatches under \$200 with built-in GPS, NFC payments, and ECG or heart-rate variability tracking. State GPS, NFC, and health feature. If none have ECG, list HRV instead.
  \item Find 3 midsize camping tents under \$200 with capacity for at least four people, full rainfly, and tent weight under 15 pounds. List capacity, rainfly type, and weight. If criteria not fully met, show closest option.
  \item List 2 home security cameras under \$100 with 1080p resolution, night vision, two-way audio, and local storage option. Provide each feature’s status. If no local storage, note cloud-only limitation.
  \item Recommend 3 pair of running shoes under \$120 with carbon-fiber plate or equivalent propulsion tech, neutral support, and weight under 10 ounces. State plate tech, support type, and weight. If none, note closest feature set.
  \item Recommend 2 DSLR lenses under \$500 for wildlife photography that have at least 300mm focal length, image stabilization, and autofocus under 0.5s. List focal length, stabilization type, and measured autofocus time. If none, suggest closest alternatives and note compromises.
  \item Find 3 camping stoves under \$150 that support propane and butane, boil one liter of water in under four minutes, and have built-in wind protection. Provide fuel type compatibility, boil time, and wind guard design details. If fewer than three meet all, include near matches with missing features.
  \item List 2 insulated tumblers under \$40 with 30-hour cold retention, 12-hour hot retention, and dishwasher-safe lid. State retention times, lid type, and size. If criteria are missing, recommend closest and explain trade-offs.
  \item Recommend 3 external monitors under \$300 with at least 27-inch size, IPS panel, 75 Hz refresh rate, and USB-C power delivery. Provide screen size, panel type, refresh rate, and wattage delivered via USB-C. If power delivery is absent, note fallback features.
  \item Find 2 portable power stations under \$400 with AC outlets, solar charging support, and at least 500 Wh capacity. List AC output wattage, solar input type, and capacity. If solar charging not supported, mention alternative recharge methods.
  \item List 3 noise-monitoring baby monitors under \$200 with temperature display, lullaby/music playback, and two-way talk. Provide screen size (or app), temperature reporting, and audio features. If lullaby feature is missing, note which are closest.
  \item Recommend 2 countertop ice makers under \$250 that produce at least 26 lbs of ice per day, have self-clean function, and use bullet-shaped ice. List daily output, cleaning cycle, and ice type. If none match exactly, propose closest and trade-offs.
  \item Find 3 inflatable paddle boards under \$700 with maximum load of at least 300 lbs, included pump, and thickness of at least 6 inches. Provide max load, pump type, and board thickness. If load capacity slightly lower, note it.
  \item List 2 smart jump ropes under \$100 with integrated fitness tracking, Bluetooth app syncing, and rechargeable battery. State tracking metrics, app availability, and battery runtime. If rechargeable is not available, recommend suitable near match.
  \item Recommend 3 gaming keyboards under \$150 with per-key RGB lighting, hot-swappable switches, and dedicated macro keys. Provide switch type, lighting software, and macro implementation details. If hot-swap is not present, note it.
  \item Recommend 3 sulfate-free shampoos under \$25 that are color-safe, contain at least 2\% argan oil, and have a pH between 5 and 6. List ingredient percentages, color-safe claims, and pH value. If pH is not listed, note this clearly.
  \item Find 2 paraben-free retinol serums under \$50 with at least 0.5\% retinol, added vitamin C, and cruelty-free certification. Clearly state retinol and vitamin C concentrations, and certification status.
  \item List 3 mineral sunscreens under \$30 that are zinc-oxide based, at least SPF 30, reef-safe (no oxybenzone/octinoxate), and water-resistant for at least 80 minutes. Provide SPF, zinc-oxide percentage, and water resistance time.
  \item Recommend 2 fragrance-free facial moisturizers under \$40 with hyaluronic acid, non-comedogenic, and dermatologist-tested. List hyaluronic acid percentage, comedogenic rating, and dermatologist testing claims.
  \item Find 3 sulfate-free cleansing oils under \$35 with at least two plant oils, vitamin E, and eco-cert organic certification. Clearly state oil types, vitamin E content, and certification status.
  \item List 2 paraben-free body lotions under \$20 with at least 10\% shea butter, fast-absorbing formula, and allergy-tested status. Provide shea butter percentage, absorption claim, and testing certification.
  \item Recommend 3 vegan lipsticks under \$25 that are cruelty-free, provide at least six-hour wear, and include SPF 15. List wear time, SPF rating, and certification details. If SPF is absent, clearly state the trade-off.
  \item Find 2 alcohol-free facial toners under \$30 with probiotics or niacinamide, non-pore-clogging, and a pH between 4 and 5. List active ingredients and pH; if pH is not listed, clearly note it.
  \item List 3 caffeine-infused eye creams under \$35 that are fragrance-free, paraben-free, and claim to reduce puffiness within 15 minutes. State caffeine concentration, effectiveness claim timing, and ingredient exclusions.
  \item Recommend 2 multifunctional balm sticks under \$30 that are petroleum-free, contain SPF, and can be used on lips, cheeks, and cuticles. Provide ingredient list, SPF rating, and specified usage areas. Clearly note if multifunction use is limited.
\end{itemize}

\section{Experiment Results}\label{app:results}

\begin{figure*}[h]
\centering
\begin{subfigure}[b]{0.49\textwidth}
    \centering
\includegraphics[width=\textwidth]{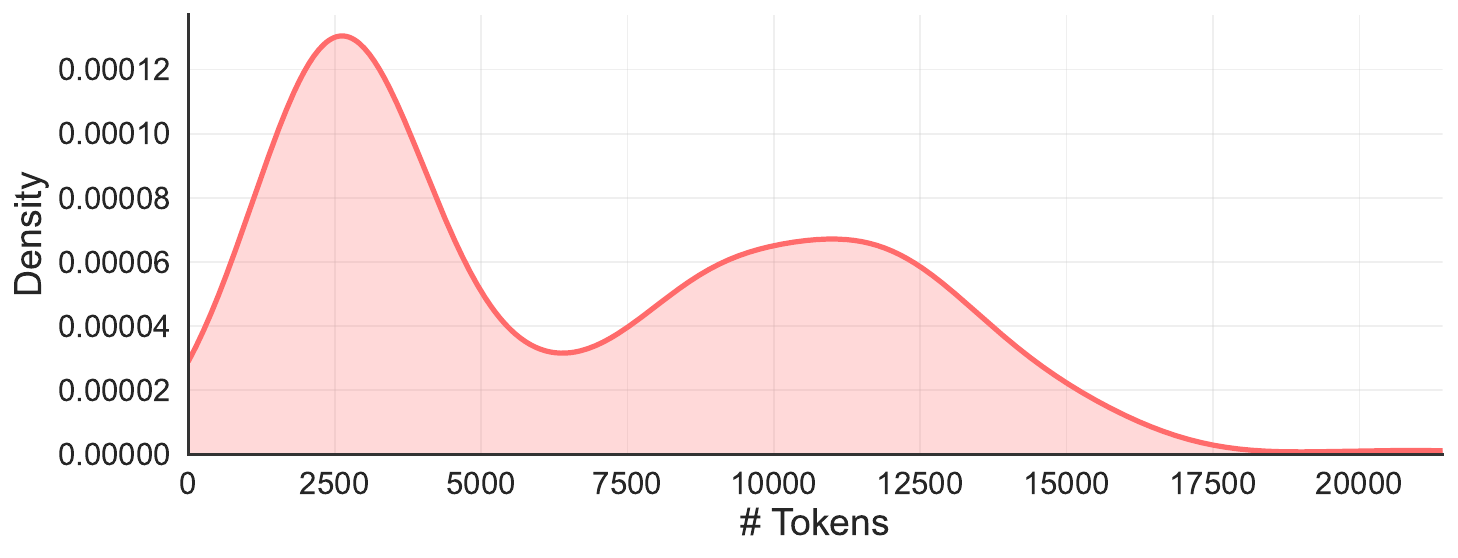}
    \caption{WebVoyager}
    \label{fig:token_webvoyager} 
\end{subfigure}
\hfill
\begin{subfigure}[b]{0.49\textwidth}
    \centering
    \includegraphics[width=\textwidth]{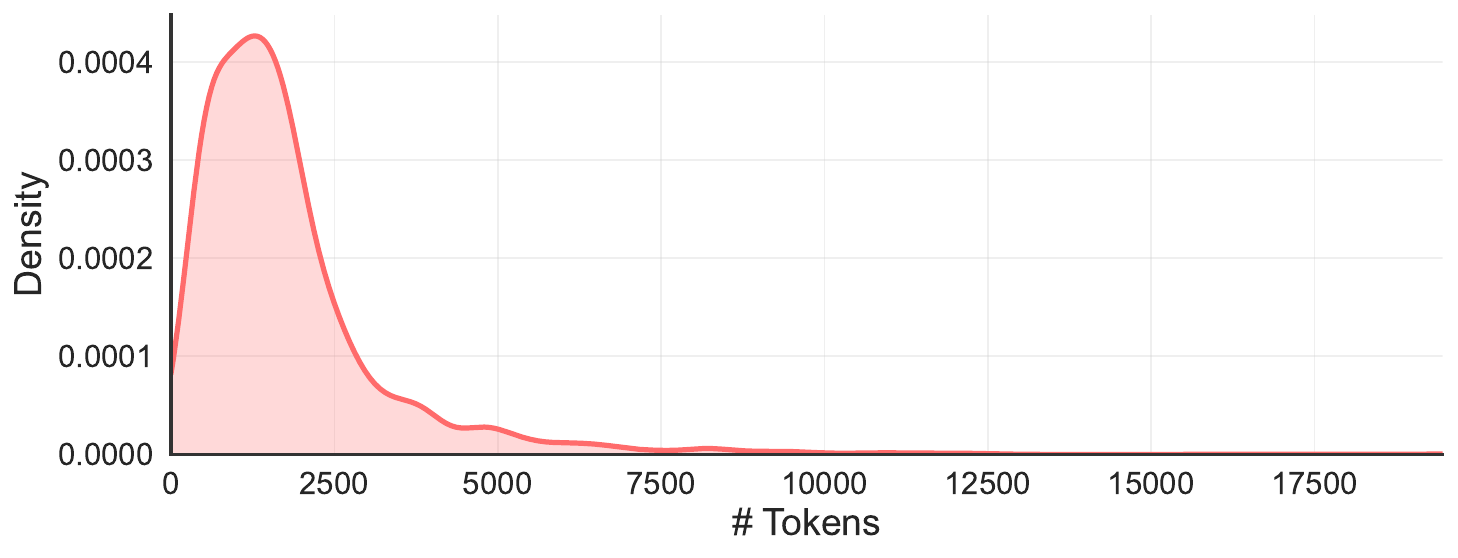}
    \caption{WebArena (Shopping)}
    \label{fig:token_webarena} 
\end{subfigure}
\caption{Distribution of average HTML observation lengths (in tokens) per timestep across agent trajectories on WebVoyager and the WebArena Shopping category. Token counts reflect only raw web page content, excluding any additional context provided to the agent.}
\label{res:tokendist}
\end{figure*}

\begin{table}[h]
\centering
\footnotesize
\setlength{\tabcolsep}{2.5pt}
\renewcommand{\arraystretch}{0.85}
\begin{tabular}{lcccccc}
\toprule
\textbf{Method} & \textbf{Map} & \textbf{Shop.} & \textbf{Reddit} & \textbf{GitLab} & \textbf{Overall} \\
 & (81) & (126) & (11) & (63) & (281) \\
\midrule
\sa{} & 0.348 & \textbf{0.432} & 0.545 & 0.383 & \textbf{0.401} \\
\ao{} & \textbf{0.410} & 0.370 & \textbf{0.591} & \textbf{0.410} & 0.399 \\
\bottomrule
\end{tabular}
\caption{\looseness=-1 Success rates on WebArena using Kimi-K2 as the backbone model, evaluated on the subset of tasks with URL and string evaluators.}
\label{res:webarena_kimi_results}
\end{table}

\begin{figure*}[h]
\centering
\begin{subfigure}[b]{0.24\textwidth}
    \centering
    \includegraphics[width=\textwidth]{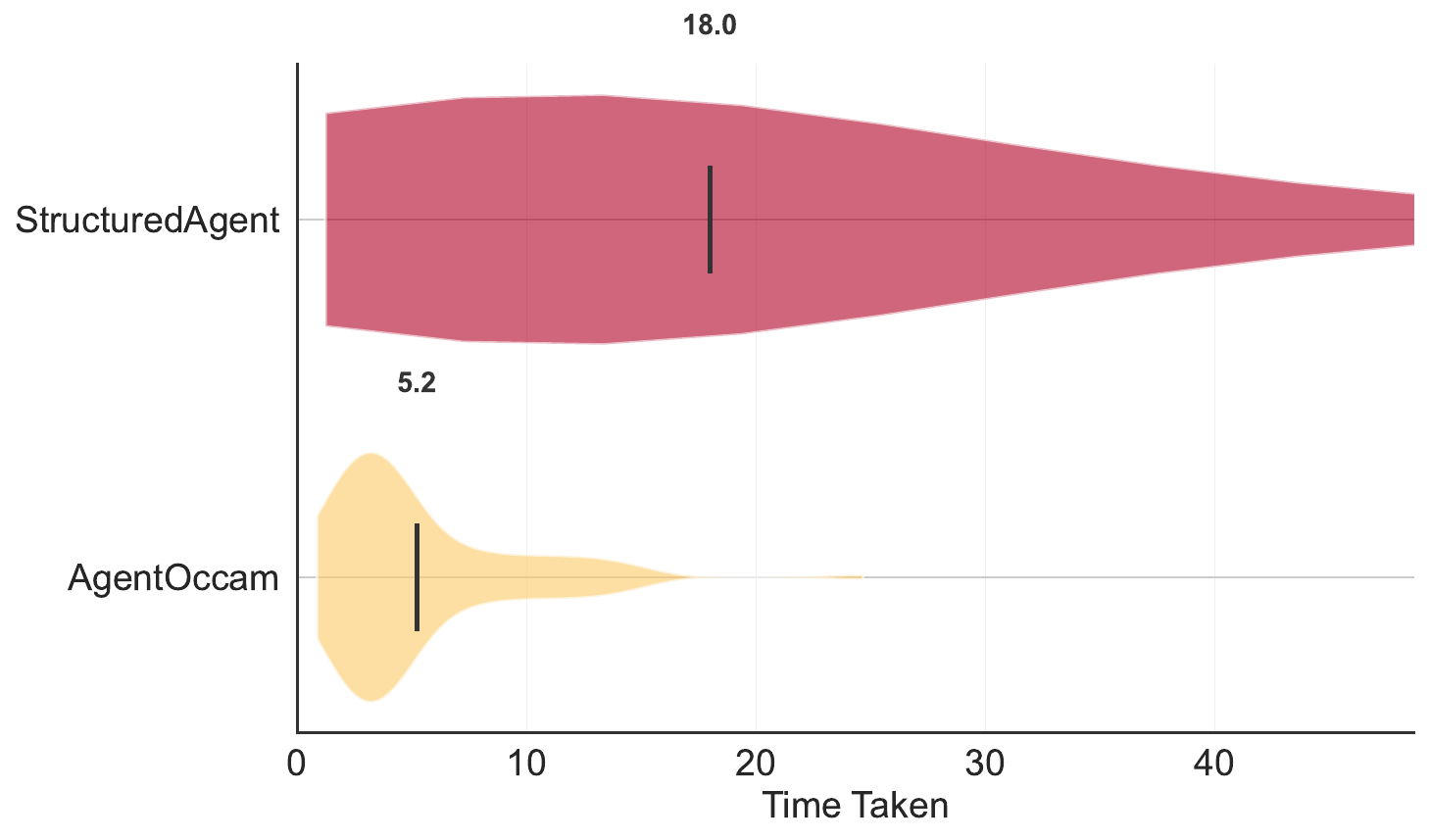}
    \caption{GITLAB}
    \label{fig:time_gitlab_app}
\end{subfigure}
\hfill
\begin{subfigure}[b]{0.24\textwidth}
    \centering
    \includegraphics[width=\textwidth]{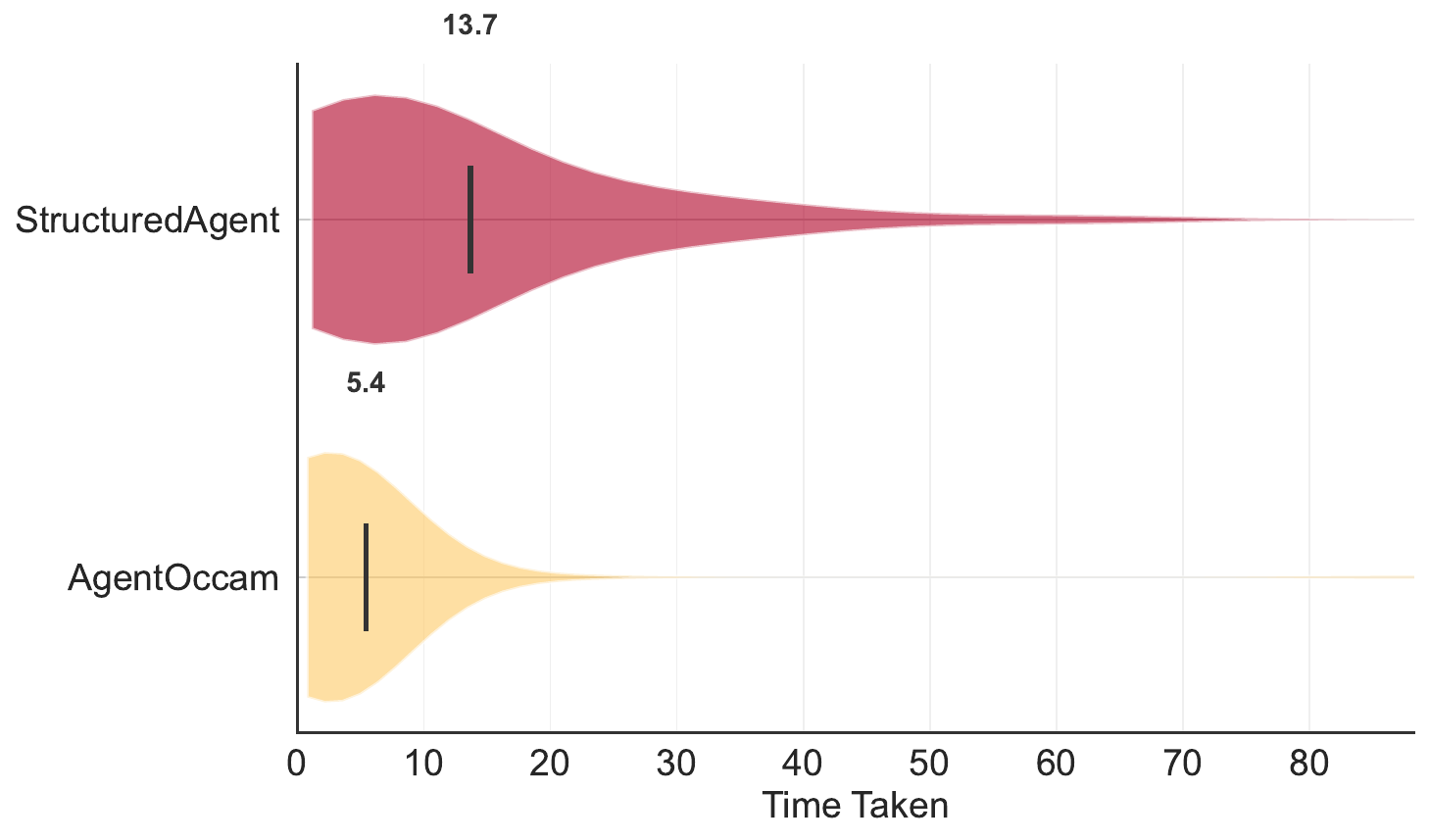}
    \caption{MAP}
    \label{fig:time_map_app}
\end{subfigure}
\hfill
\begin{subfigure}[b]{0.24\textwidth}
    \centering
    \includegraphics[width=\textwidth]{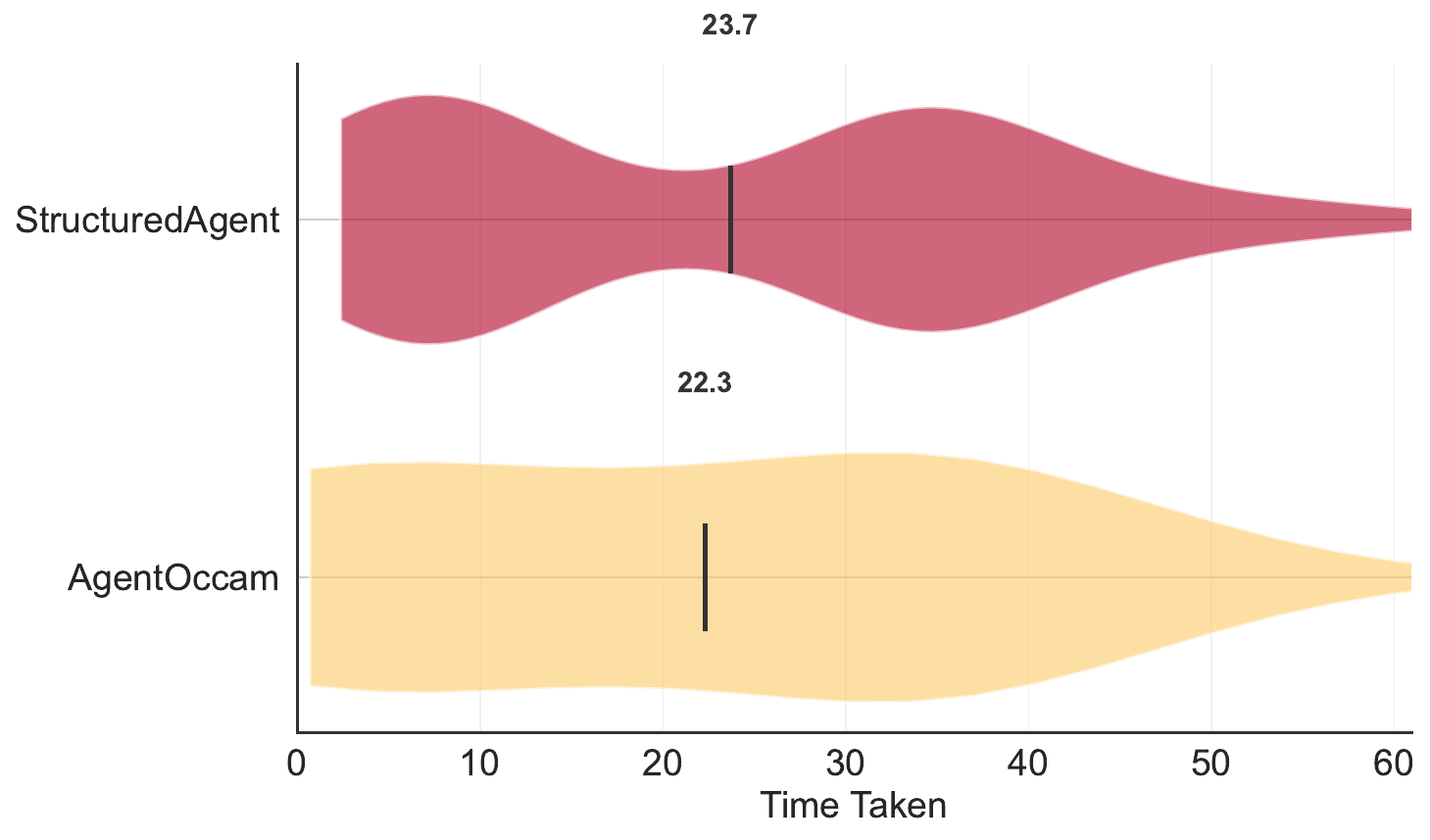}
    \caption{REDDIT}
    \label{fig:time_reddit_app}
\end{subfigure}
\begin{subfigure}[b]{0.24\textwidth}
    \centering
    \includegraphics[width=\textwidth]{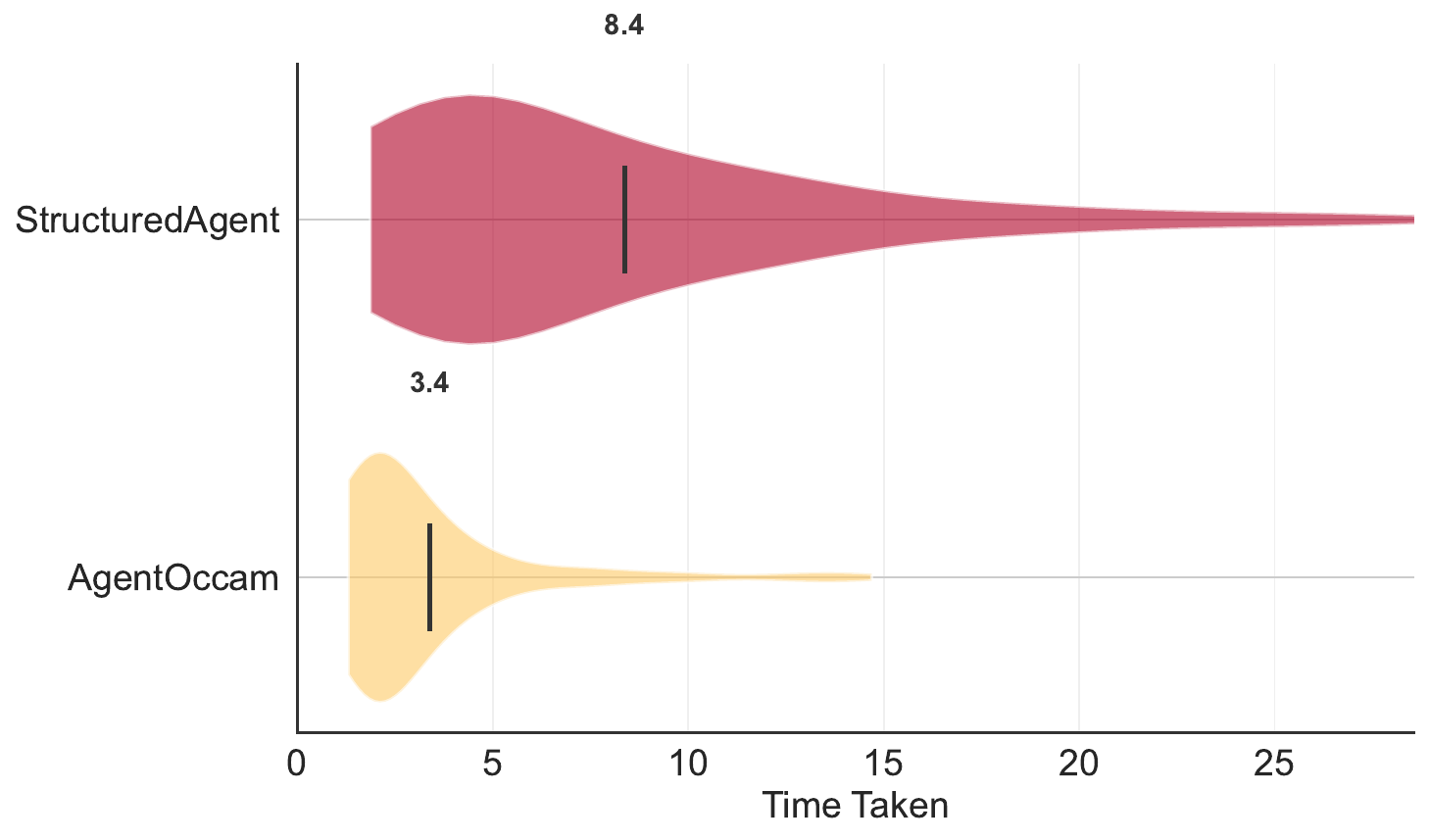}
    \caption{SHOPPING}
    \label{fig:time_reddit_app2}
\end{subfigure}
\caption{Completion time distributions (in minutes) for \sa{} and \ao{} across four WebArena task categories, with the mean marked per violin. \sa{} exhibits longer completion times and heavier right tails, reflecting greater variance in task duration due to its deliberate planning and dynamic replanning upon encountering failures.}
\label{res:time_taken_webarena}
\end{figure*}

\begin{figure*}[h!]
\centering

\begin{subfigure}[b]{0.33\textwidth}
    \centering
    \includegraphics[width=\textwidth]{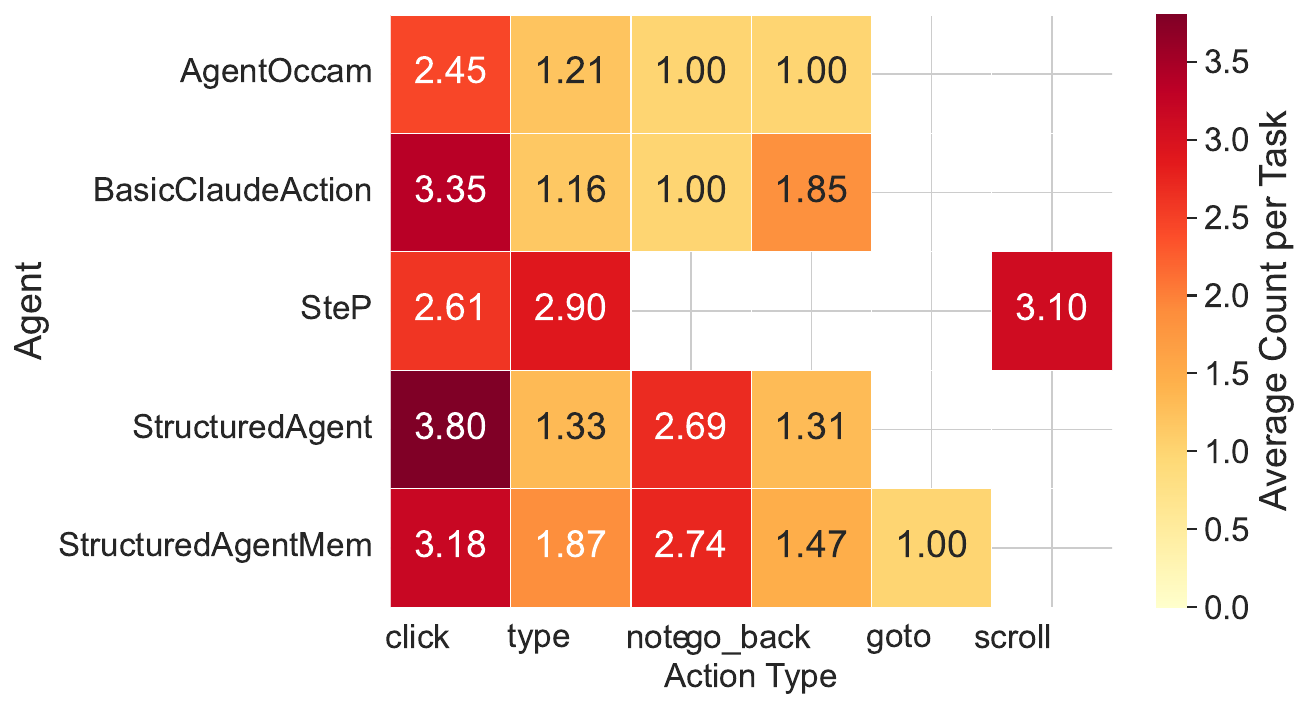}
    \caption{Amazon-easy}
    \label{fig:heatmap_amazon_easy}
\end{subfigure}
\hfill
\begin{subfigure}[b]{0.33\textwidth}
    \centering
    \includegraphics[width=\textwidth]{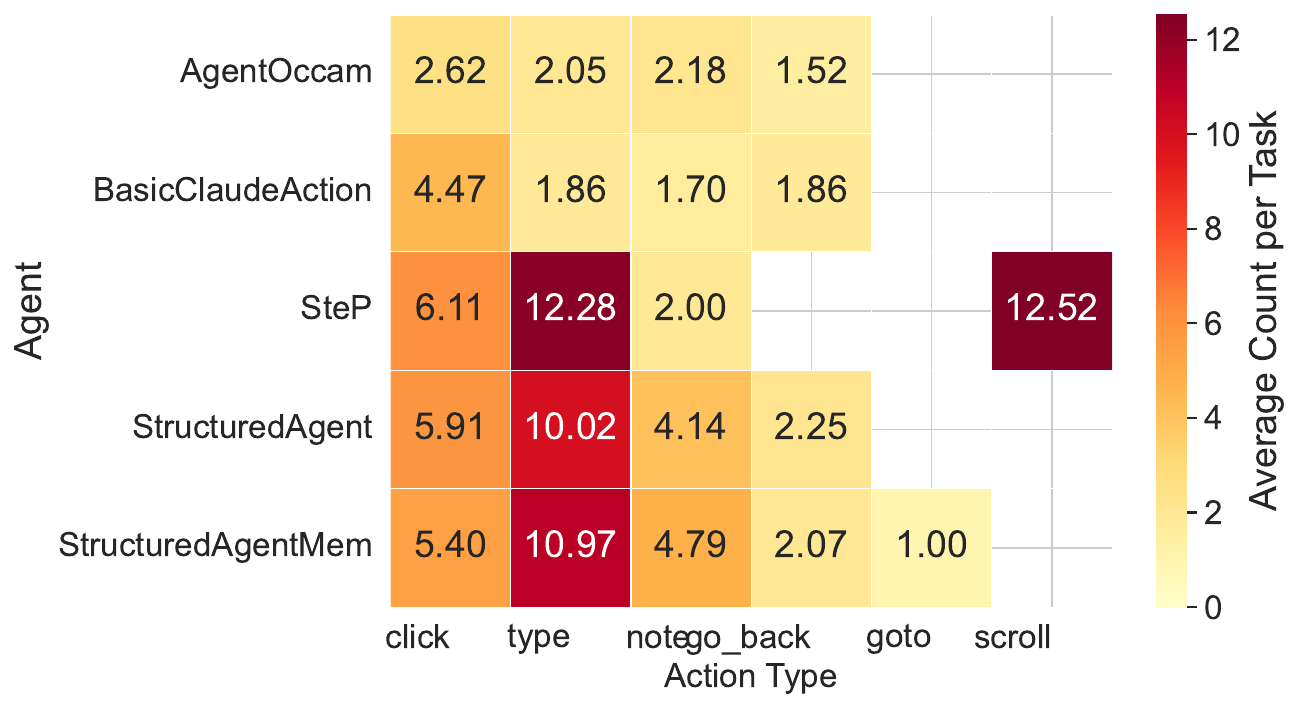}
    \caption{Amazon-hard}
    \label{fig:heatmap_amazon_hard}
\end{subfigure}
\hfill
\begin{subfigure}[b]{0.33\textwidth}
    \centering
    \includegraphics[width=\textwidth]{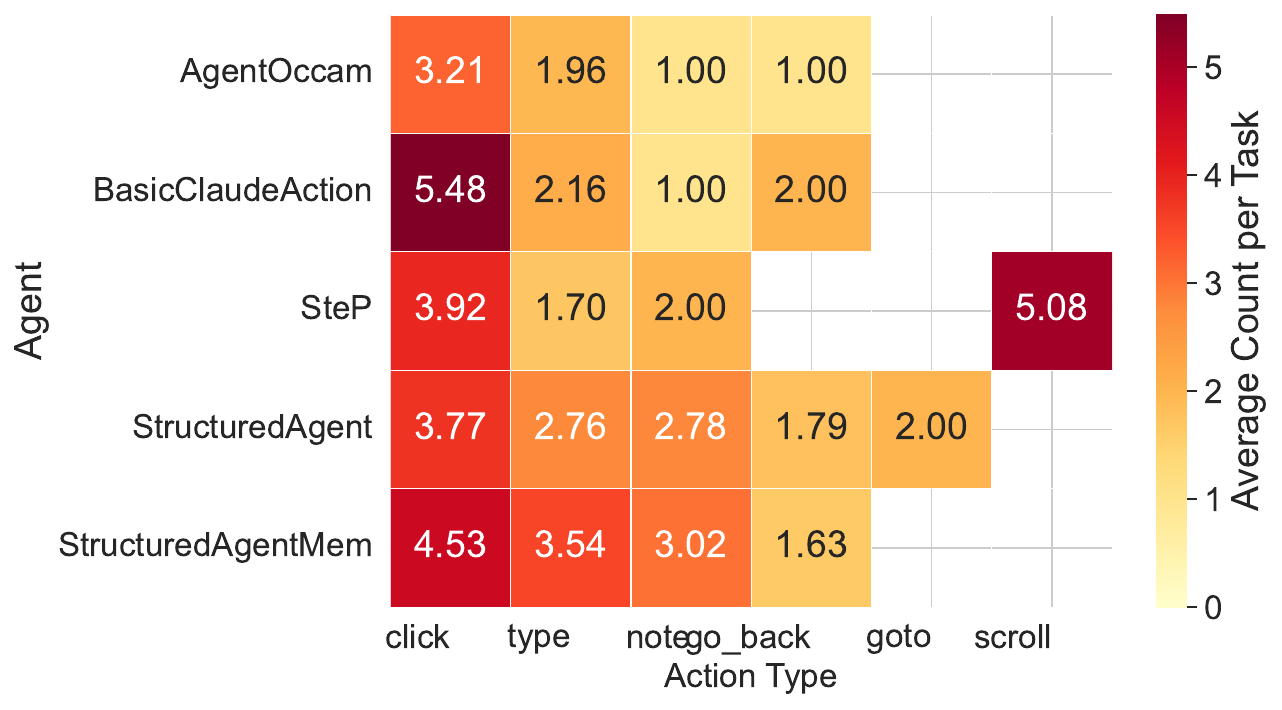}
    \caption{WebVoyager Other}
    \label{res:heatmap_webvoyager}
\end{subfigure}

\caption{Average action counts per task for each agent on Amazon and WebVoyager Other tasks, broken down by action type (click, type, note, go\_back). \sa{} and \samem{} issue more \texttt{type} and \texttt{note} actions on harder tasks because they gather and record more information about candidate items.}
\label{res:heatmaps_comparison_webvoyager}
\end{figure*}

  \begin{figure*}[h!]
  \centering
  \hfill
  \begin{subfigure}[b]{0.24\textwidth}
      \centering
      \includegraphics[width=\textwidth]{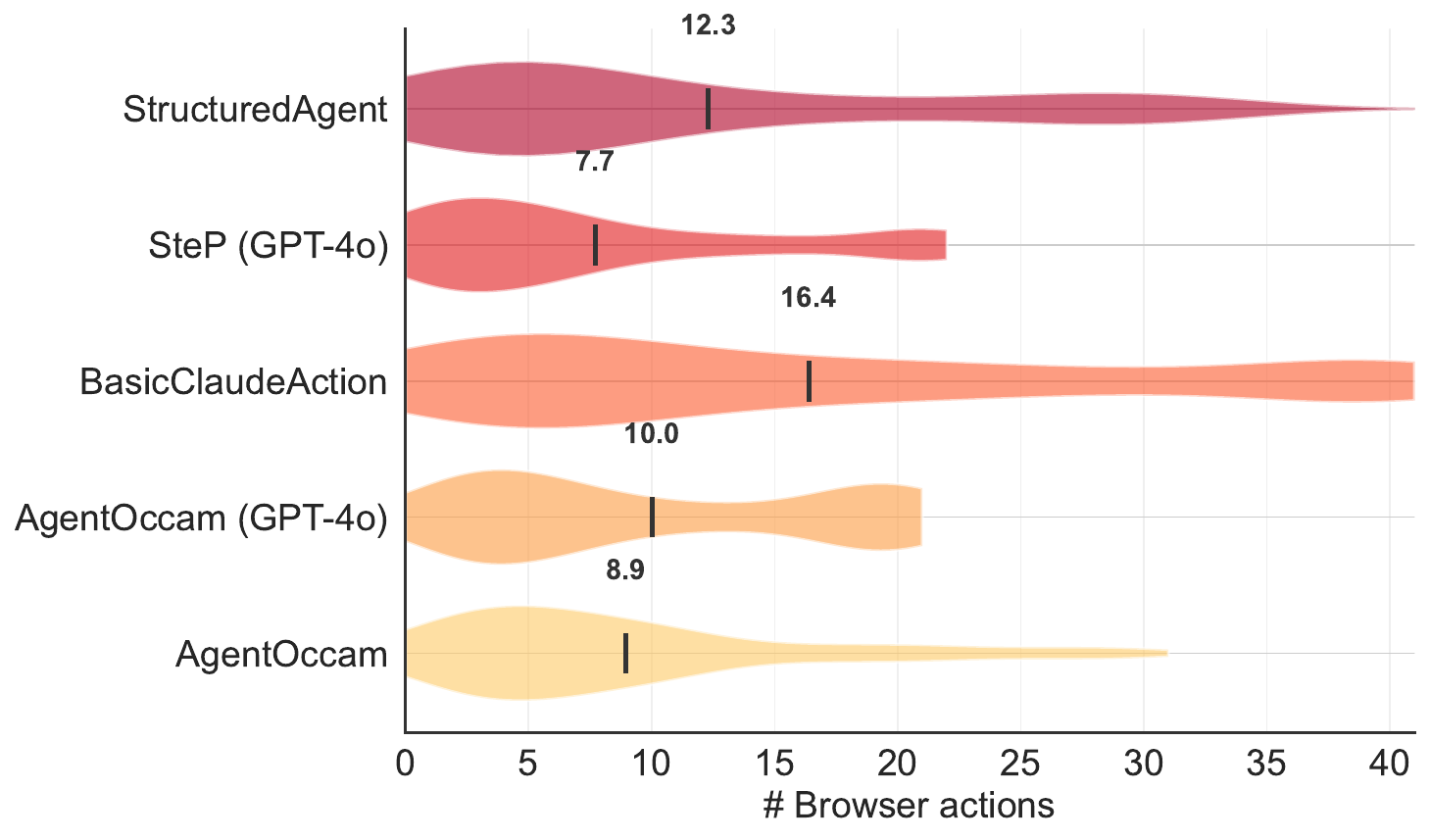}
      \caption{GitLab}
      \label{fig:time_gitlab_app3}
  \end{subfigure}
  \hfill
  \begin{subfigure}[b]{0.24\textwidth}
      \centering
      \includegraphics[width=\textwidth]{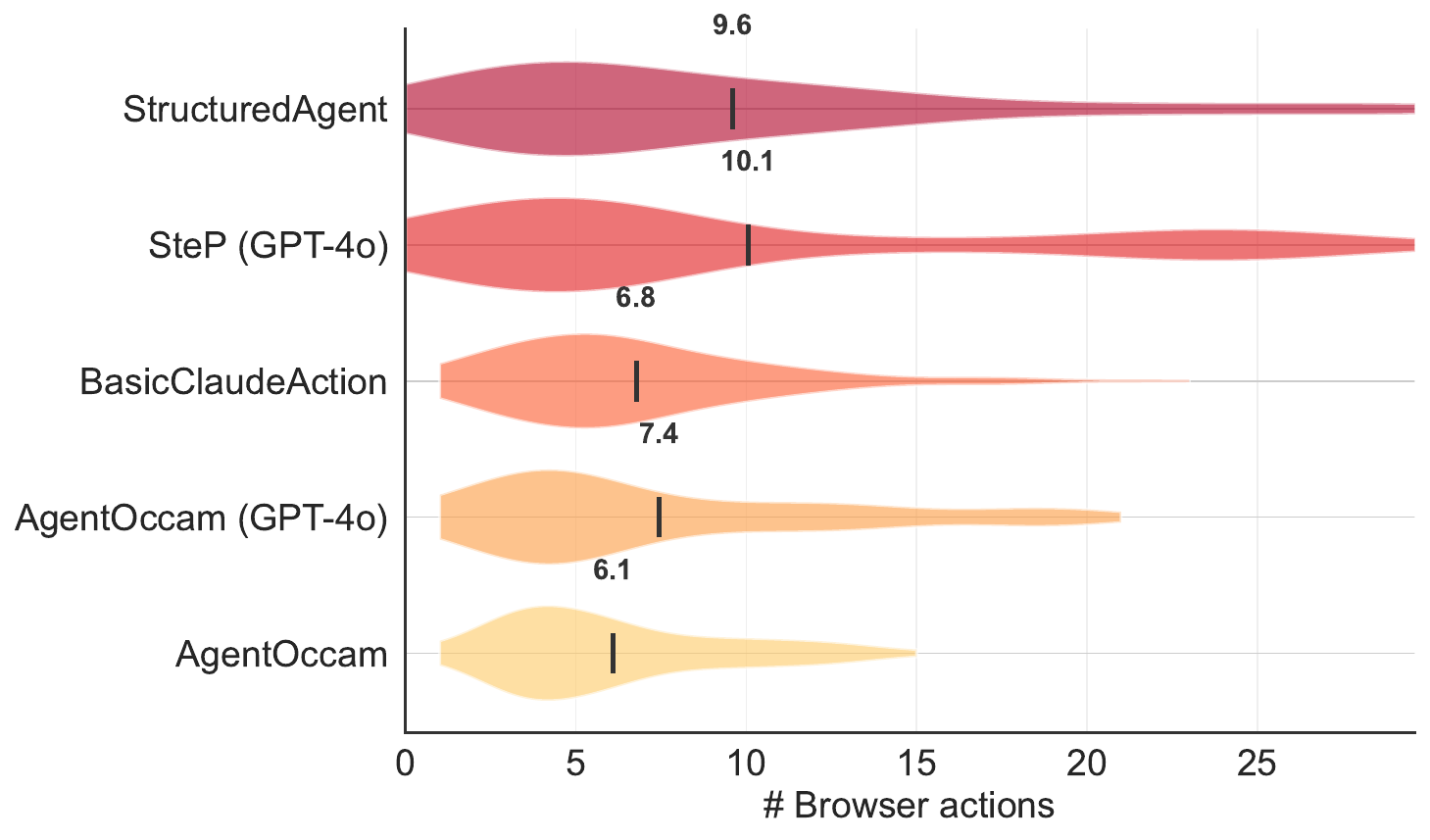}
      \caption{Map}
      \label{fig:time_map_app3}
  \end{subfigure}
  \hfill
  \begin{subfigure}[b]{0.24\textwidth}
      \centering
      \includegraphics[width=\textwidth]{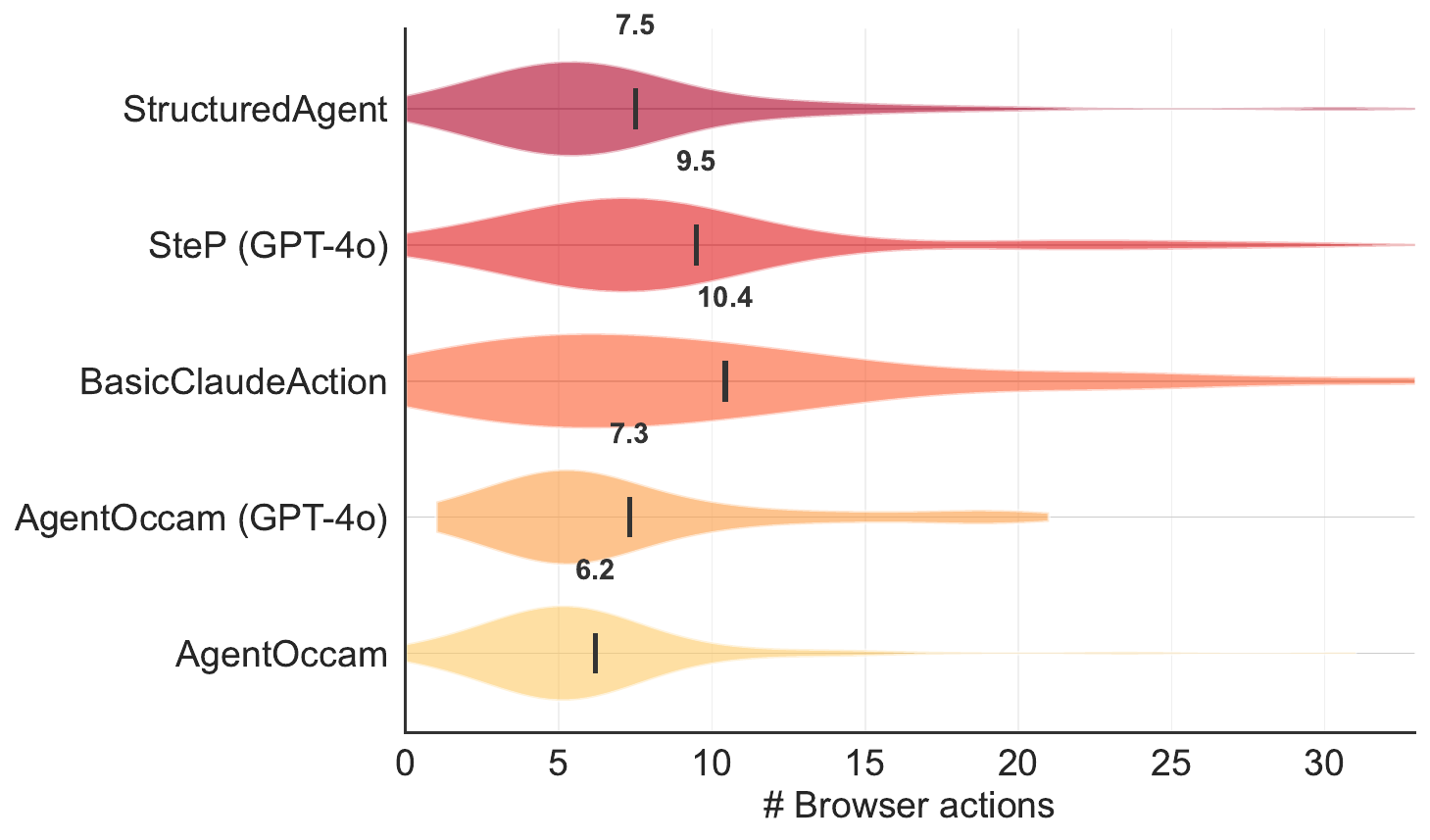}
      \caption{Reddit}
      \label{fig:time_reddit_app3}
  \end{subfigure}
  \begin{subfigure}[b]{0.24\textwidth}
      \centering
      \includegraphics[width=\textwidth]{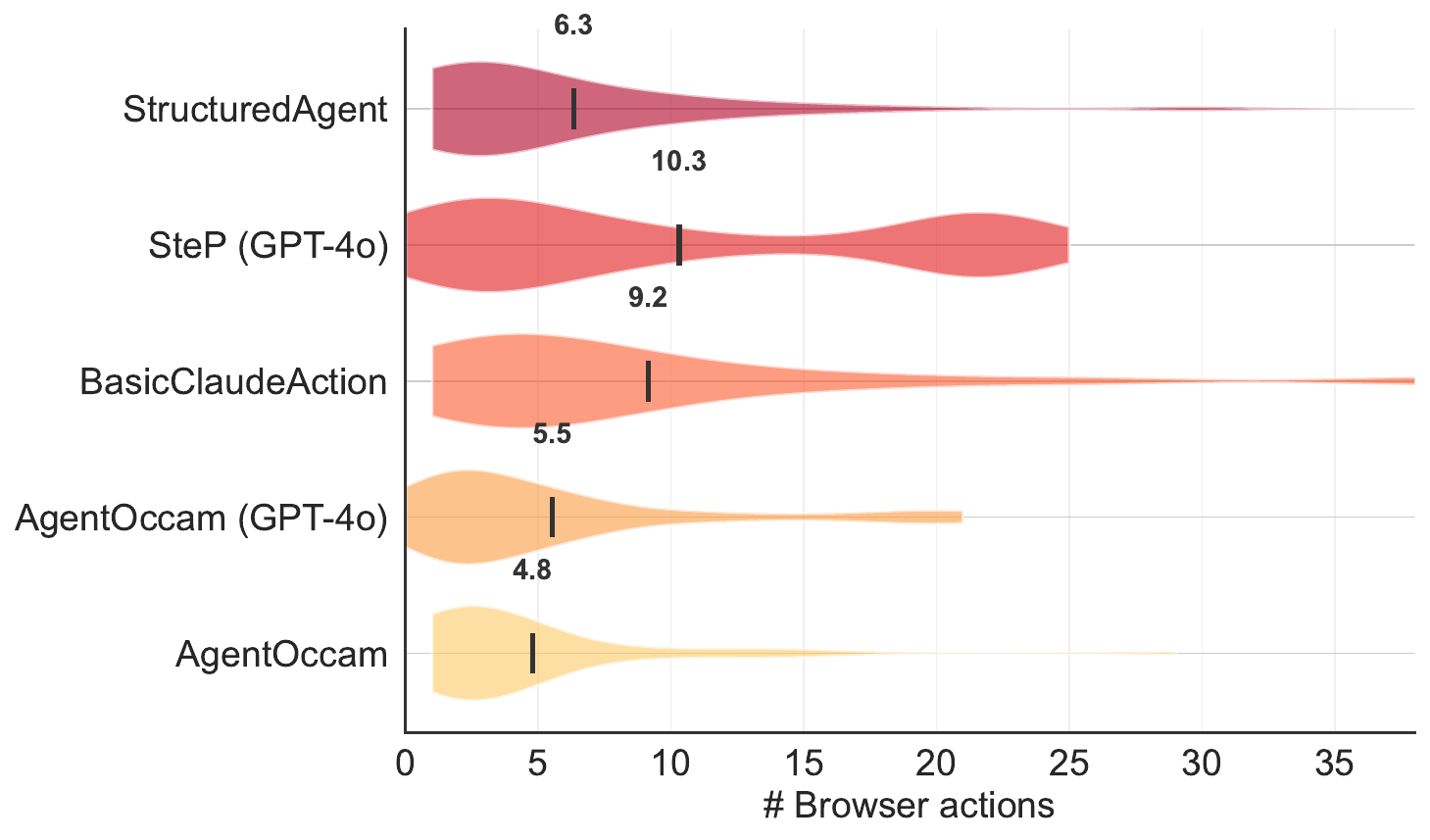}
      \caption{Shopping}
      \label{fig:time_shopping}
  \end{subfigure}
  \caption{\SA{}'s longer trajectories on harder tasks (Map, GitLab) reflect systematic exploration of alternative strategies, which translates to higher overall success rates (\cref{res:webarena_results}). On
  Shopping and Reddit, where \SA{} achieves the largest gains, trajectory lengths remain comparable to \ao{}.}
  \label{res:trajectory_lengths_webarena}
  \end{figure*}
  
%%% Amazon trajectory-length figure moved to the main paper (Figure~\ref{fig:traj_amazon}). %%%

\begin{figure*}[h]
\centering
\begin{subfigure}[b]{0.33\textwidth}
    \centering
    \includegraphics[width=\textwidth]{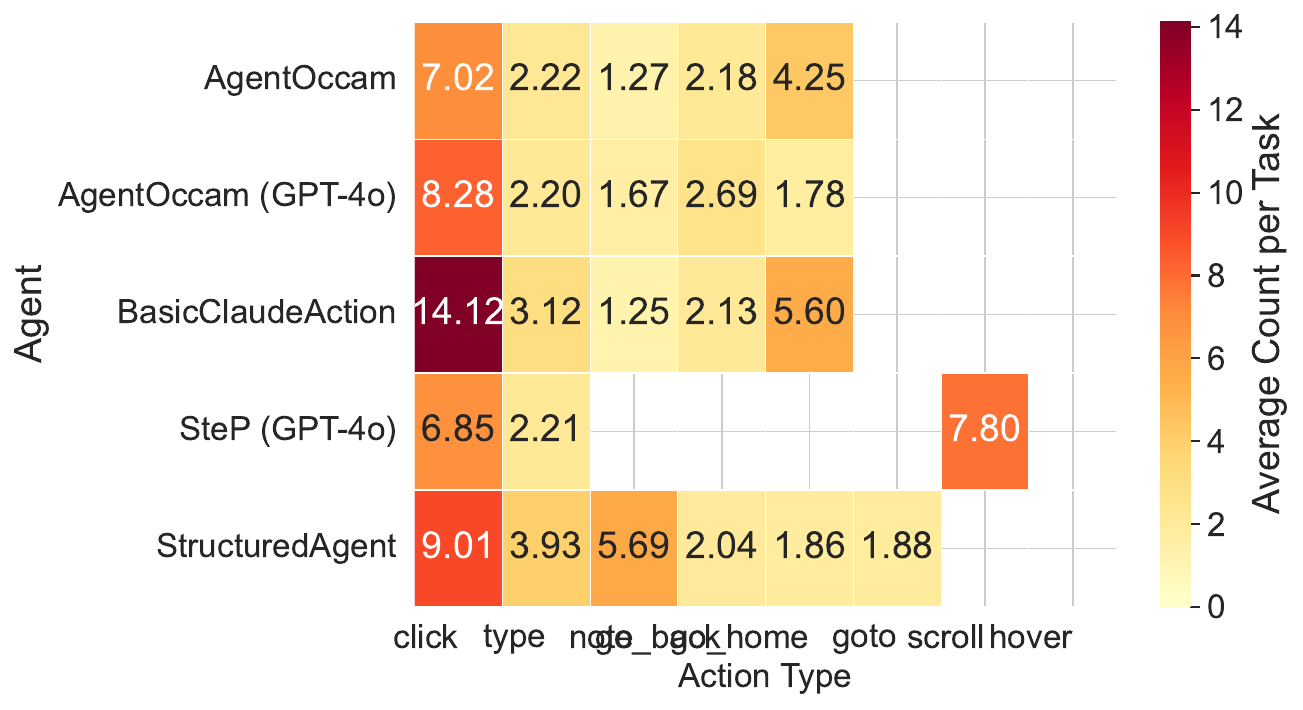}
    \caption{GITLAB}
    \label{fig:heatmap_gitlab}
\end{subfigure}
\hfill
\begin{subfigure}[b]{0.33\textwidth}
    \centering
    \includegraphics[width=\textwidth]{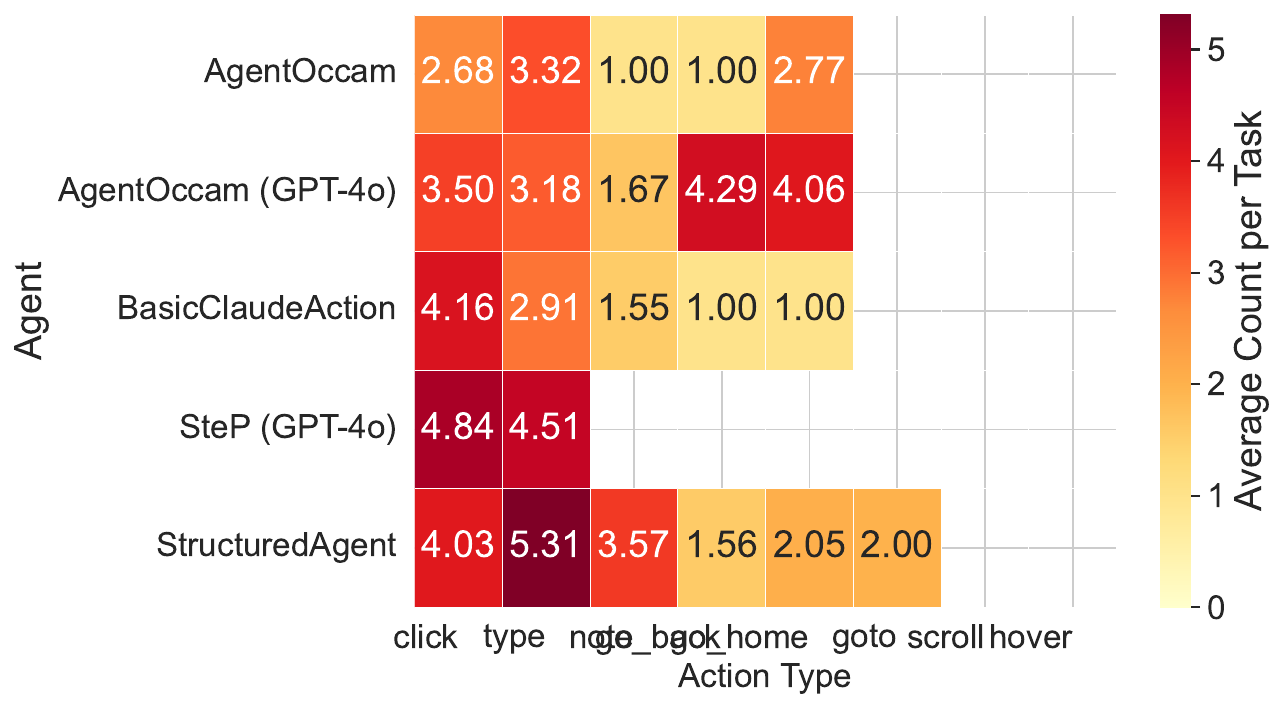}
    \caption{MAP}
    \label{fig:heatmap_map}
\end{subfigure}
\hfill
\begin{subfigure}[b]{0.33\textwidth}
    \centering
    \includegraphics[width=\textwidth]{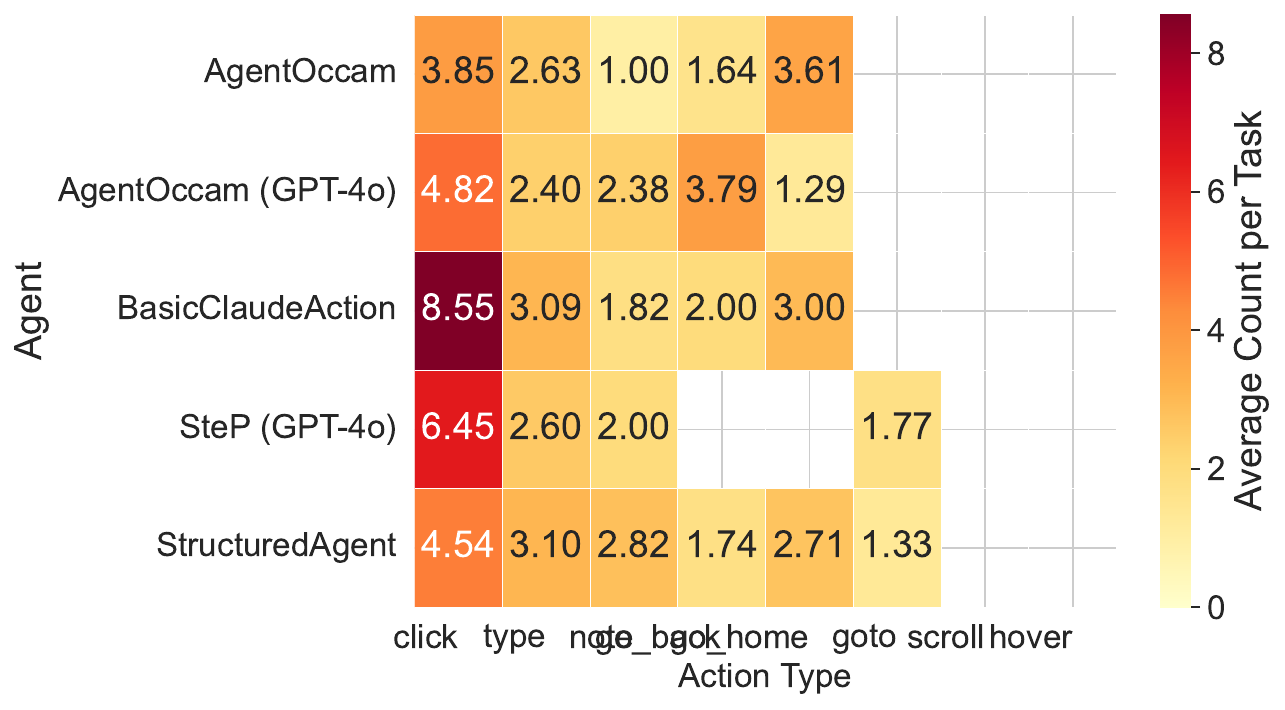}
    \caption{REDDIT}
    \label{fig:heatmap_reddit}
\end{subfigure}
% \begin{subfigure}[b]{0.33\textwidth}
%     \centering
%     \includegraphics[width=\textwidth]{images/heatmap_actions_SHOPPING.pdf}
%     \caption{SHOPPING}
%     \label{fig:heatmap_shopping}
% \end{subfigure}
\caption{Average action counts per task for each agent across three WebArena task categories (GitLab, Map, Reddit), broken down by action type (click, type, note, go\_back, go\_home). \sa{} issues more \texttt{note} actions because it records structured task information. Its click and type counts are similar to the baselines.}
\label{res:heatmaps_comparison_webarena}
\end{figure*}

\section{Detailed Ablation Analysis}\label{app:ablation_detail}
  \Cref{tab:tree-analysis-easy,tab:tree-analysis-hard} report \ANDOR{} tree
  statistics for all four ablation conditions on the Amazon-easy and Amazon-hard
  splits.
  
  \paragraph{Metric Definitions.}
  \begin{itemize}[leftmargin=0.4cm, itemsep=1pt]
  \item \textbf{Success Rate} is the task success rate as judged by the
  specified LLM evaluator (Gemini or Opus).
  \item \textbf{Constraint Satisfaction} is the fraction of user-specified
  constraints satisfied across all tasks, including partial successes.
  \item \textbf{Browser Steps} is the mean number of browser actions executed
  per task.
  \item \textbf{Max Depth} is the mean maximum depth of the \ANDOR{} tree.
  \item \textbf{Total, AND, OR, and Action Nodes} are mean node counts by type.
  \item \textbf{Pruned Nodes} is the mean number of nodes marked as pruned
  because they are irrelevant or irrecoverable.
  \item \textbf{Tree Efficiency} is the fraction of nodes that contributed to
  the final solution, meaning nodes that were neither pruned nor deleted.
  \item \textbf{Repair Trigger Rate} is the fraction of tasks with at least one
  node failure that invokes repair when repair is enabled.
  \item \textbf{Recovery Rate} is the fraction of repair attempts that restored
  the failed node to a passing state.
  \item \textbf{Plan Preservation} is the fraction of originally planned nodes
  that remained in the final tree without deletion.
  \item \textbf{Total Revisions} is the mean total number of tree modification
  operations, counting expansions, repairs, and prunings.
  \item \textbf{Nodes Revised} is the mean number of distinct nodes modified by
  at least one revision.
  \end{itemize}
  
  \begin{table*}[t]
  \centering
  \small
  \caption{\ANDOR{} tree analysis on the Amazon-easy split with a Haiku~4.5
  backbone. Full uses both \NODEREPAIR{} and \OR{} alternatives, OR only
  disables repair, Repair only disables \OR{} nodes, and Neither disables both.
  All rates are percentages and count-valued metrics report means. Repair
  Trigger Rate counts node failures that would invoke repair, whether or not
  repair is enabled. The no-repair conditions make no repair attempts, so their
  Recovery Rate is reported as 0 by convention.}
  \label{tab:tree-analysis-easy}
  \setlength{\tabcolsep}{5pt}
  \begin{tabular}{@{}l cccc@{}}
  \toprule
   & \textbf{Full} & \textbf{OR only} & \textbf{Repair only} & \textbf{Neither} \\
  \midrule
  \multicolumn{5}{l}{\textit{Task Performance}} \\[2pt]
  \quad Success Rate (Gemini) & 68.3 & 65.0 & 75.0 & 60.0 \\
  \quad Success Rate (Opus) & 67.5 & 50.0 & 57.5 & 50.0 \\
  \quad Constraint Satisfaction & 89.3 & 83.5 & 87.7 & 77.2 \\
  \midrule
  \multicolumn{5}{l}{\textit{Tree Structure}} \\[2pt]
  \quad Browser Steps & 19 & 8 & 22 & 8 \\
  \quad Max Depth & 3.6 & 3.5 & 4.1 & 3.1 \\
  \quad Total Nodes & 34 & 13 & 34 & 11 \\
  \quad AND Nodes & 3.4 & 3.0 & 4.0 & 2.4 \\
  \quad OR Nodes & 0.2 & 0.1 & 0.0 & 0.1 \\
  \quad Action Nodes & 18 & 7 & 21 & 7 \\
  \midrule
  \multicolumn{5}{l}{\textit{Repair \& Pruning}} \\[2pt]
  \quad Pruned Nodes & 12.2 & 3.0 & 9.2 & 2.2 \\
  \quad Tree Efficiency & 74 & 73 & 78 & 76 \\
  \quad Repair Trigger Rate & 22 & 32 & 25 & 18 \\
  \quad Recovery Rate & 44 & 0 & 70 & 0 \\
  \quad Plan Preservation & 99 & 93 & 98 & 96 \\
  \midrule
  \multicolumn{5}{l}{\textit{Tree Revisions}} \\[2pt]
  \quad Total Revisions & 64 & 24 & 68 & 20 \\
  \quad Nodes Revised & 33 & 13 & 34 & 11 \\
  \bottomrule
  \end{tabular}
  \end{table*}
  
  \begin{table*}[t]
  \centering
  \small
  \caption{\ANDOR{} tree analysis on the Amazon-hard split with a Haiku~4.5
  backbone. Conditions and conventions are as in \Cref{tab:tree-analysis-easy}.
  All rates are percentages and count-valued metrics report means.}
  \label{tab:tree-analysis-hard}
  \setlength{\tabcolsep}{5pt}
  \begin{tabular}{@{}l cccc@{}}
  \toprule
   & \textbf{Full} & \textbf{OR only} & \textbf{Repair only} & \textbf{Neither} \\
  \midrule
  \multicolumn{5}{l}{\textit{Task Performance}} \\[2pt]
  \quad Success Rate (Gemini) & 38.3 & 13.3 & 35.0 & 16.7 \\
  \quad Success Rate (Opus) & 14.5 & 8.5 & 20.0 & 6.8 \\
  \quad Constraint Satisfaction & 82.8 & 59.8 & 77.8 & 57.3 \\
  \midrule
  \multicolumn{5}{l}{\textit{Tree Structure}} \\[2pt]
  \quad Browser Steps & 43 & 11 & 46 & 12 \\
  \quad Max Depth & 5.3 & 3.6 & 5.6 & 3.8 \\
  \quad Total Nodes & 80 & 20 & 84 & 20 \\
  \quad AND Nodes & 6.1 & 3.2 & 7.0 & 3.4 \\
  \quad OR Nodes & 0.6 & 0.1 & 0.0 & 0.1 \\
  \quad Action Nodes & 42 & 10 & 45 & 11 \\
  \midrule
  \multicolumn{5}{l}{\textit{Repair \& Pruning}} \\[2pt]
  \quad Pruned Nodes & 29.3 & 6.2 & 32.1 & 5.0 \\
  \quad Tree Efficiency & 62 & 59 & 64 & 65 \\
  \quad Repair Trigger Rate & 63 & 58 & 70 & 60 \\
  \quad Recovery Rate & 37 & 0 & 43 & 0 \\
  \quad Plan Preservation & 95 & 88 & 96 & 88 \\
  \midrule
  \multicolumn{5}{l}{\textit{Tree Revisions}} \\[2pt]
  \quad Total Revisions & 172 & 36 & 190 & 36 \\
  \quad Nodes Revised & 79 & 20 & 83 & 20 \\
  \bottomrule
  \end{tabular}
  \end{table*}

  \section{Error Localization Experiment}\label{app:error_localization}
  
  We test whether the hierarchical structure of \SA{} makes its failures easier to diagnose than those of \ao{}. We run the experiment on the 15 Amazon tasks that both agents failed, where the \SA{} root node is
  marked as pruned and the \ao{} task remains incomplete. The design controls for prompt phrasing, judge identity, and forced choice scoring.
  
  \paragraph{Diagnosis.} We pass the full trajectory of each agent to a fixed diagnosis model, Claude Opus 4.6, which returns a structured diagnosis. The diagnosis states the root cause, the critical steps where
  the agent went wrong, a recommended fix, a failure category, and a confidence score. Each agent is represented by the trajectory format it natively produces. \ao{} receives its flat sequence of steps together
  with the interaction history notes it records at each step. \SA{} receives its \ANDOR{} decomposition tree, in which every node carries an execution status of \texttt{success}, \texttt{fail}, \texttt{pruned}, or
  \texttt{visited}, followed by the same sequence of steps and notes. We require every claim in a diagnosis to rest on the observations, actions, and notes in the trajectory, and we pass the trajectory in full
  without truncation.
  
  \paragraph{Blinded comparison.} We present the two diagnoses to a judge as ``Agent A'' and ``Agent B'' and randomize their order for each task under a fixed seed, so the judge never learns which architecture
  produced which diagnosis. The judge rates four dimensions. Precision measures how exactly the diagnosis locates the failure. Self sufficiency measures whether the diagnosis lets an engineer repair the agent
  without rereading the trajectory. Clarity measures how easily a reader understands the failure. Actionability measures how directly the recommended fix can be implemented. The judge also states an overall
  preference. On every item the judge chooses $A$, $B$, or Tie. We map the verdicts back to \SA{} and \ao{} only after collecting them. We use two judges that differ from the diagnosis model, Gemini 2.5 Flash and
  Claude Opus 4.8.
  
  \paragraph{Prompt variants.} To rule out sensitivity to a single phrasing, we run the diagnosis stage under four paraphrases (detailed, terse, chain of thought, and free category). We run the comparison stage
  under four paraphrases that cross two question wordings with two tie tolerance framings. Each judge therefore produces $15 \times 4 \times 4$ verdicts for every dimension.
  
  \paragraph{Results.} \SA{} wins on every dimension for both judges (\Cref{tab:diagnosis_quality}). Pooled across judges and paraphrases, the judges prefer the \SA{} diagnosis on $78.5\%$ of the decided overall
  preference verdicts at $p<10^{-36}$. The largest margin is on Precision, where \SA{} wins $90.1\%$ of decided verdicts pooled across judges. This is consistent with the mechanism we expect, since the tree with
  per-node execution statuses points the diagnosis model directly at the sub-goal that failed, whereas a flat trajectory forces it to reconstruct the failure point from the step sequence alone.
  
  \begin{table}[H]
  \centering
  \scriptsize
  \caption{Blinded comparison of \SA{} and \ao{} diagnoses across 15 tasks that both agents failed, 4 diagnosis paraphrases, and 4 comparison paraphrases. Each cell reports the percentage of decided verdicts that
  favor \SA{}, followed by the raw counts SA/AO/Tie. The diagnosis model, Claude Opus 4.6, differs from both judges. We randomize the presentation order for each task under a fixed seed. We test significance with
  a one sided binomial test that compares \SA{} against \ao{} and excludes ties. The markers $^{*}$, $^{**}$, and $^{***}$ denote $p<0.05$, $p<0.01$, and $p<0.001$.}
  \label{tab:diagnosis_quality}
  \resizebox{\ifdim\width>\columnwidth \columnwidth\else\width\fi}{!}{%
  \begin{tabular}{lccc}
  \toprule
  Dimension & Gemini 2.5 Flash & Opus 4.8 & Pooled \\
  \midrule
  Precision & 86.8 (203/31/3)$^{***}$ & 94.7 (160/9/70)$^{***}$ & 90.1 (363/40/73)$^{***}$ \\
  Self sufficiency & 64.3 (9/5/223) & 72.4 (155/59/25)$^{***}$ & 71.9 (164/64/248)$^{***}$ \\
  Clarity & 63.2 (91/53/93)$^{***}$ & 76.9 (183/55/1)$^{***}$ & 71.7 (274/108/94)$^{***}$ \\
  Actionability & 78.3 (162/45/30)$^{***}$ & 67.6 (148/71/20)$^{***}$ & 72.8 (310/116/50)$^{***}$ \\
  \midrule
  Four dimension pooled & 77.6 (465/134/349)$^{***}$ & 76.9 (646/194/116)$^{***}$ & 77.2 (1111/328/465)$^{***}$ \\
  Overall preference & 81.7 (192/43/2)$^{***}$ & 75.2 (170/56/13)$^{***}$ & 78.5 (362/99/15)$^{***}$ \\
  \bottomrule
  \end{tabular}}
  \end{table}

\begingroup
\begin{table*}[!htp]
\tiny
\centering
\begin{tabular}{p{0.9\linewidth}}
\toprule
You are an expert evaluator analyzing web browsing agent trajectories to diagnose why a task failed.

You will be given: the task objective, which constraints were NOT satisfied, and the agent's step-by-step trajectory.

Your job is to perform a precise root-cause analysis of why the agent failed.

You MUST respond in this exact JSON format:

\{
  "root\_cause": "<one sentence: the primary reason the task failed>",
  "critical\_steps": [
    \{"step\_number": <int>, "what\_happened": "<what the agent did>", "what\_went\_wrong": "<why this was the critical error>"\}
  ],
  "what\_should\_have\_been\_done": "<specific alternative action(s)>",
  "failure\_category": "<one of: wrong\_search\_strategy | missed\_information | navigation\_error | premature\_stop | constraint\_violation | verification\_failure | other>",
  "confidence": <1-5 integer>
\}

EXAMPLE of a valid response (structure only -- your content will differ):

\{
  "root\_cause": "The agent searched with the wrong keyword and never applied the price filter.",
  "critical\_steps": [
    \{"step\_number": 2, "what\_happened": "Typed 'cheap laptop' into search", "what\_went\_wrong": "Ignored the required 'under \$500' constraint, returning irrelevant results"\},
    \{"step\_number": 5, "what\_happened": "Clicked the first result and stopped", "what\_went\_wrong": "Never verified the price met the constraint"\}
  ],
  "what\_should\_have\_been\_done": "Search 'laptop', then apply the <\$500 price filter and sort by rating before selecting.",
  "failure\_category": "wrong\_search\_strategy",
  "confidence": 4
\}

Output ONLY the JSON object -- no prose before or after, no markdown code fences. Use exactly these keys, once each.

GROUNDING (mandatory): Every claim in your diagnosis MUST be grounded in or directly implied by the observations, actions, and notes in the trajectory. Do NOT assume, infer, or invent any fact that is not present in the trajectory. If the trajectory does not make the root cause clear, say so explicitly in "root\_cause" and set a LOW "confidence" (1-2) rather than guessing. Do not use hedging like "likely" or "probably" to smuggle in unsupported claims.

Guidelines:
- Be SPECIFIC. Name the wrong search term, the missed filter, the unexplored page.
- Point to exact step numbers where the critical error occurred.
- The fix should be concrete and actionable.
- Confidence 5 = root cause obvious; 1 = unclear. \\
\bottomrule
\end{tabular}
\caption{Diagnosis prompt (detailed variant) that we use in the error localization experiment. We append the agent trajectory below this system prompt. \ao{} receives its flat step sequence with the notes it records at each step. \SA{} receives its AND/OR tree with per-node status followed by its step sequence with the notes it records at each step.}
\label{tab:prompt_diagnosis}
\end{table*}
\endgroup

\section{Human Evaluation}\label{app:humaneval}

To validate the reliability of our LLM-based evaluation, we conducted a human evaluation study on the Amazon-hard task set. We recruited one human annotator who was not an author of this paper. We provided the annotator with the description of the AND/OR tree structure used by each agent and the full agent trajectories, including all browser observations and actions taken. For each of the 60 Amazon-hard tasks, the annotator judged whether each method fully completed the task, and we report the percentage of tasks that each method completed.

\begin{table}[h]
\centering
\scriptsize
\caption{Human evaluation task-success rate (\%) on the 60 Amazon-hard tasks. A single independent annotator (non-author) judged whether each method fully completed each task from its full trajectory. \Cref{tab:case603} shows an example where the annotator overlooked unmet constraints and marked a partially correct answer as complete.}
\label{tab:human_eval}
\begin{tabular}{lc}
\toprule
\textbf{Method} & \textbf{Human} \\
\midrule
\textsc{StructuredAgent} & \textbf{56.7} \\
AgentOccam & 55.0 \\
BasicClaudeAction & 40.0 \\
\bottomrule
\end{tabular}
\end{table}

The human evaluation scores (\Cref{tab:human_eval}) agree with the LLM judge averages in \Cref{res:combined}. \textsc{StructuredAgent} achieves the highest human score, followed by AgentOccam, and BasicClaudeAction trails both, which is the same ordering the automated judges produce.

\Cref{tab:evaluator_reliability} quantifies how closely each LLM judge aligns with the human annotator on the 60 Amazon-hard tasks, per agent.

\begin{table}[t]
\centering
\scriptsize
\setlength{\tabcolsep}{4pt}
\caption{Evaluator reliability on the 60 Amazon-hard tasks, per judge and agent. Agr.\ and Cohen's $\kappa$ measure how well each judge's per-task success labels match the human annotator's. Sat.\ is the judge's mean constraint-satisfaction rate, the mean of $(\text{satisfied}+1)/(\text{total}+1)$ across tasks.}
\label{tab:evaluator_reliability}
\resizebox{\ifdim\width>\columnwidth \columnwidth\else\width\fi}{!}{%
\begin{tabular}{llccc}
\toprule
\textbf{Judge} & \textbf{Agent} & \textbf{Agr.\ (\%)} & \textbf{$\kappa$} & \textbf{Sat.\ (\%)} \\
\midrule
\multirow{3}{*}{GPT-4.1-mini} & \SA{} & 81.7 & 0.64 & 87.2 \\
 & \ao{} & 75.0 & 0.50 & 90.8 \\
 & \bca{} & 86.4 & 0.71 & 82.7 \\
\midrule
\multirow{3}{*}{Gemini-2.5-Flash} & \SA{} & 71.7 & 0.45 & 82.1 \\
 & \ao{} & 68.3 & 0.38 & 84.8 \\
 & \bca{} & 86.7 & 0.71 & 78.9 \\
\midrule
\multirow{3}{*}{Kimi-K2} & \SA{} & 76.7 & 0.54 & 81.5 \\
 & \ao{} & 75.0 & 0.50 & 87.8 \\
 & \bca{} & 86.7 & 0.71 & 80.5 \\
\bottomrule
\end{tabular}}
\end{table}

\begin{table*}[t]
\centering
\small
\setlength{\tabcolsep}{5pt}
\renewcommand{\arraystretch}{1.25}
\caption{Case study of task Amazon--DeepMulti-603. The task asks the agent to recommend three wireless printers under \$150 that support both AirPrint and borderless 4$\times$6 photo printing, to list brand, print support, and price, and not to add to cart. The printer rows reproduce each agent's returned answer. \yes\ and \no\ mark whether each requested spec is grounded in the page observations rather than only asserted by the agent. \ao{} asserts that all three of its printers meet the specs, but the observations confirm this for only the TS5320. \SA{} states in its answer that none of its printers fully meet all requirements. The human annotator nevertheless marks \ao{} as a success, while the LLM judge, which requires every spec to be grounded in the observations, marks both agents as failures.}
\label{tab:case603}
\resizebox{\linewidth}{!}{%
\begin{tabular}{@{}l l c c c@{}}
\toprule
\textbf{Agent} & \textbf{Recommended printer (price)} & \textbf{Wireless,\,$<$\$150} & \textbf{AirPrint} & \textbf{Borderless 4$\times$6} \\
\midrule
\multirow{3}{*}{\ao{}}
& Canon TS5320 (\$85.99)  & \yes & \yes & \yes \\
& Epson XP-5200 (\$79.99) & \yes & \no\ \textit{(not observed)} & \yes \\
& Canon TS702a (\$109.00) & \yes & \yes & \no\ \textit{(not observed)} \\
\midrule
\multirow{4}{*}{\SA{}}
& Canon PIXMA TS6420a (\$44.99--109.95) & \yes & \yes & \no\ \textit{(not observed)} \\
& Canon PIXMA TR4720 (\$64.00)          & \yes & \no\ \textit{(not observed)} & \yes \\
& HP DeskJet 2827 ($\cdots$)            & \yes & \no & \no \\
\cmidrule(l){2-5}
& \multicolumn{4}{l}{\footnotesize From the agent's answer: ``none fully meet all requirements.''} \\
\midrule
\multicolumn{2}{@{}l}{\textbf{Judge score}}   & \multicolumn{3}{l}{\ao{}: $0.75$ \quad \SA{}: $0.50$} \\
\multicolumn{2}{@{}l}{\textbf{Human verdict}} & \multicolumn{3}{l}{\ao{}: \textbf{Yes} \quad \SA{}: \textbf{No}} \\
\multicolumn{2}{@{}l}{\textbf{LLM judge verdict}} & \multicolumn{3}{l}{\ao{}: \textbf{Failure} \quad \SA{}: \textbf{Failure}} \\
\bottomrule
\end{tabular}}
\end{table*}

\section{Prompts}\label{app:prompts}

\begingroup
\begin{table*}[!htp]
\tiny
\centering
\begin{tabular}{p{0.9\linewidth}}
\toprule
You are a web-browsing assistant designed to extract structured constraint data from user queries. Given a natural language query, your task is to identify and return:

1. A list of **distinct items** the user wants to retrieve (if any).
2. For each item, a list of **item-level constraints** that are explicitly mentioned in the query.
3. A separate list of **task-level constraints** that apply to the overall task.

Definitions:

- **Item**: A specific product, object, or entity that the user wants to find or retrieve information about.
- **Item-level constraint**: Any attribute or specification that is directly tied to an individual item. This may include (but is not limited to): brand, model, price, rating, features, color, size, dimensions, delivery time, material, and any other **explicitly** stated property of the item.
- **Task-level constraint**: A condition that affects the overall search or task execution rather than a specific item. 

    Instructions:

- You must extract **only those task-level and item-level constraints that are explicitly stated** in the query.
- You must **not assume, infer, or interpret** any implicit constraints, even if they seem logically implied or contextually likely.
- Do **not include any reasoning, commentary, or explanation** in your output.
- Each item type must appear only once in the `ITEM\_CONSTRAINTS` list.
- If the query contains no valid constraints or items, return empty lists using the required format.

    Output Format:

The result must be returned strictly in the following JSON format:

\{
  "TASK\_CONSTRAINTS": ["{explicit\_task\_constraint1}", "{explicit\_task\_constraint2}", "..."],
  "ITEM\_CONSTRAINTS": [
    \{
      "{item\_type}": ["{explicit\_item\_constraint1}", "{explicit\_item\_constraint2}", "..."]
    \},
    ...
  ]
\}

    EXAMPLE:

QUERY: Find a black laptop bag under $40 and a wireless mouse under $25. Both should be delivered within 2 days.  
URL: www.amazon.com

OUTPUT:

\{
 \"TASK\_CONSTRAINTS\": [
  ],
  \"ITEM\_CONSTRAINTS\": [
    \{
      "laptop bag": [
        "Color: black",
        "Price: under \$40",
        "Delivery time: within 2 days"
      ]
    \},
    \{
      "wireless mouse": [
        "Price: under \$25",
        "Type: wireless",
        "Delivery time: within 2 days"
      ]
    \}
  ]
\}

QUERY: \{task\_objective\} \\
\bottomrule
\end{tabular}
\caption{Prompt for extracting task constraints for task completion evaluation.}
\label{tab:constraints}
\end{table*}
\endgroup

\begingroup
\begin{table*}[!htp]
\tiny
\centering
\begin{tabular}{p{0.9\linewidth}}
\toprule
You are an evaluator assessing the performance of a web-browsing agent on a specific task.

You will be provided with the following components:

- Task Description  
   A natural-language instruction outlining the goal the agent is expected to accomplish. This may involve retrieving, summarizing, comparing, or verifying information from the web.

- Constraints  
   A list of specific item-level and task-level requirements that must be satisfied for the task to be considered successful.

- Trajectory Summary  
   A step-by-step log of the agent’s interaction with the environment. Each time step includes:
   - Observation: Summary of the web-page the agent perceives at that moment.
   - Action: The behavior or command the agent executes (e.g., clicking, searching).
   - Interaction History Summary: Summary of the agent's progress (if available)

- The agent's final response to the task

- Evidence extracted from raw observations  
   A digest of what was actually observed across the agent's full trajectory.

    Evaluation Guidelines

- **Evaluate each item-level constraint once per item type**.  
  Do not repeat the constraint evaluation for each instance of the same item type.  
  Instead, assess whether the constraint is satisfied **across all relevant items of that type**, and justify the evaluation with evidence from the observations, notes, or interaction history.  
  **For example:** If the task requires finding 3 laptops under $1000, and the agent recommends 3 laptops, evaluate the constraint "Price: under $1000" only once. Mark it as **satisfied only if all 3 laptops** meet the price constraint.

- Only use information that is **explicitly visible** in the agent’s observations, actions, notes, and final response.  
  Do not rely on external knowledge, domain expertise, or assumptions made by the agent.
  
- Do **not assume that a constraint is satisfied** based on the agent’s search terms, reasoning, or keyword usage (e.g., searching for "size 6").  
  A search query does **not guarantee** that the results meet the constraint.  
  A constraint is only considered satisfied if the specific value (e.g., "Size: 6") is explicitly shown in the observations, actions or agent’s notes. 

- If the task requires finding the cheapest, most rated, or most reviewed item, only verify that appropriate filters or sorting options were applied - but **only if those options are actually available on the website**.

- Evaluate only based on the **final response** of the agent.  
  Intermediate steps or partially explored options should not be considered unless they directly contribute to the final result.

- Do **not award partial credit** for incomplete constraints.  
  A constraint is either **fully satisfied** or **not satisfied**. All requested attribute values (e.g., price, brand, size) are treated as **separate constraints**.

- Do **not evaluate** the agent based on constraints that are **not explicitly stated** in the task description.  
  For example, if the task does not mention "size," "material," or "capacity," do not treat them as constraints — even if the agent includes or discusses them during the task.

- If there is no final response or the task is incomplete, assign a score of 0.

- You do not need to consider constraints that require the agent to find items with a given delivery period as the agent does not have access to the current date and time.

    Scoring Instructions

- Assign **1 point** for each item-level or task-level constraint that is fully and explicitly grounded in the observations, notes and actions. **Confirm that the agent is not hallucinating.**
  **Note:** If the recommended item costs **less than** the specified price, it still satisfies the price constraint.
    Evidence for constraints satisfied must be grounded only in observations, notes and actions. 
    **Do not** ground evidence in the final response of the agent.

- Assign **0 points** for constraints that are:
  - Not satisfied or Partially Satisfied  
  - Assumed or inferred without explicit evidence in observations, notes, or final response

- Assign **1 point** if the task was completed successfully and the final response is likely **not hallucinated**.  
  A task is considered **completed successfully** if the agent provides a clear and specific final response that directly fulfills the task objective, and is **explicitly supported** by observed content or notes.

- Normalize the score:  
  **Score = (Number of Satisfied Constraints + 1 point if task completed) / (Total Number of Constraints + 1)**  
  This yields a final score between **0.0 and 1.0**.

    Final Verdict

- **SUCCESS** – Only if **all** explicitly stated item and task-level constraints are fully satisfied (score = 1.0)  
- **FAILURE** – If **any** constraint is not satisfied, or if the task is incomplete (score < 1.0)

    Your response must strictly follow this format:

1. **Task Overview**
- **Task Description**:  
  (Copy the task as provided.)

- **Constraints**:  
  (List the item-level and task-level constraints exactly as stated in the task description.  
  **Do not infer or add any additional constraints.**)

2. **Evaluation of Each Constraint**  
Evaluate each constraint using one of the following labels:  
- Satisfied  
- Partially Satisfied  
- Not Satisfied

Constraint 1:   
Evidence: 
Evaluation: 
Reason: 

Constraint 2:   
Evidence:
Evaluation: 
Reason: 

(Continue for all constraints.)
**Note**: Evidence must be grounded in observations, notes and actions only. Evidence cannot be grounded in the final response of the agent.

3. **Agent Behavior Assessment**
- **Action Effectiveness**:  
  Was the agent’s sequence of actions appropriate and goal-directed?

- **Interpretation Accuracy**:  
  Did the agent correctly interpret the content it observed?

- **Handling of Multi-step Tasks (if applicable)**:  
  Did the agent address all required components of the task?

4. **Final Scoring**
- Total Satisfied Constraints: X  
- Total Constraints: Y  
- Score (X / Y): Z  
- Final Verdict: SUCCESS / FAILURE

5. **Summary Comments (Optional)**  
(Provide any additional insights or suggestions about the agent’s behavior or reasoning.)

OBJECTIVE:
\{objective\}

CONSTRAINTS:
\{constraints\}

FINAL RESPONSE:
\{notes\}

TRAJECTORY
\{trajectory\} \\
\bottomrule
\end{tabular}
\caption{Prompt for evaluating completion of task based on observation summaries, actions, notes and final response of the agent.}
\label{tab:evaluation}
\end{table*}
\endgroup

\begingroup
\begin{table*}[!htp]
\tiny
\centering
\begin{tabular}{p{0.9\linewidth}}
\toprule
You are a meticulous evidence-EXTRACTION assistant. You are NOT the judge. You are shown the raw text of ONE web page OBSERVATION (an accessibility tree / page dump) captured during a browsing session, plus the OBJECTIVE and CONSTRAINTS. Your only job is to COPY OUT, verbatim, every snippet in THIS observation that bears on any constraint or on overall task completion, and return it as JSON.

VERBATIM RULE (most important): every "quote" you emit MUST be an EXACT, character-for-character substring of THIS observation's raw text -- copy-pasted. Do NOT reword, summarize, normalize, round, translate, or reformat. If a price appears as "-31 109.98 List:159.99", copy exactly that; never rewrite it as "\$109.98". If a title spans a long string, copy the whole literal string. A reader must be able to Ctrl-F your quote and find it unchanged in this observation.

BANNED -- never report content in your own words ("the page shows", "it lists", "priced at", etc.). Every "quote" is a raw verbatim substring; nothing else.

OUTPUT FORMAT -- return ONLY a JSON object, no prose, no markdown fences:
\{"items": [\{"constraint": "<the constraint label from CONSTRAINTS this snippet bears on, verbatim>", "quote": "<exact verbatim substring copied from THIS observation>", "step\_id": <the integer step id given below>\}]\}
- One entry per distinct relevant snippet. If several constraints share one snippet, emit one entry per constraint.
- If NOTHING in this observation bears on the task, return exactly: \{"items": []\}

Rules:
- Copy real text from THIS observation only. Never invent. Never infer. Never reference other steps.
- Do NOT read or rely on any agent's summary or answer -- extract only from the raw observation text.
- Do NOT judge. No verdict/status/agreement words. Only the JSON object.

EXAMPLE (illustrative only -- a made-up museum page, unrelated to your actual task):
CONSTRAINTS: ["Ticket price: under 20 EUR", "Open on Mondays"]
THIS OBSERVATION (step\_id=2): "link 'Rijks Gallery' StaticText 'Adult admission EUR 18,50' StaticText 'Closed on Mondays' button 'Buy'"
Correct output:
\{"items": [\{"constraint": "Ticket price: under 20 EUR", "quote": "Adult admission EUR 18,50", "step\_id": 2\}, \{"constraint": "Open on Mondays", "quote": "Closed on Mondays", "step\_id": 2\}]\}
Note the quotes are copied character-for-character (e.g. "EUR 18,50", not "18.50 euros"). \\
\bottomrule
\end{tabular}
\caption{Prompt for extracting grounded, verbatim evidence from each raw observation. The per-observation snippets are consolidated into the evidence digest supplied to the evaluator.}
\label{tab:evidence_extractor}
\end{table*}
\endgroup

\begingroup
\begin{table*}[!htp]
\tiny
\centering
\begin{tabular}{p{0.9\linewidth}}
\toprule
You are an efficient $\ANDOR$ Tree Constructing Agent specialized in web-browsing tasks. You solve complex problems using $\ANDOR$ planning trees. You dynamically construct $\ANDOR$ planning trees from observations of the webpage's HTML DOM structure for efficient and robust task execution.  

You are provided: 

- root-level task description

- task constraints at item level
 
- HTML DOM structure as the observation  

- node\_id and desription of the node to analyze

- summary of current task progress

- summary of notes taken so far

- Information about node's siblings $\AND$ node's parent's siblings.

Node definitions: 

Node Types <AND/ OR / Atomic>

- $\AND$ node: Represents an ordered list of logical sub-goals required to achieve the node's objective.

- $\OR$ node: Represents alternative sub-strategies (which can be other $\ANDOR$ nodes)

- Atomic Action: Single executable action strictly matching {{navigation\_specifications}} described below.     

Node status indicators: 

- Unvisited nodes marked UNVISITED  

- Pruned nodes marked PRUNED  

- Completed nodes marked SUCCESS  

- Temporarily failed nodes marked FAIL

NAVIGATION\_SPECIFICATIONS: <LIST OF BROWSER ACTIONS>

Your Task

For the given node:

1. First, briefly analyze the node type:  
   Based on the available information and the node description, determine whether the node is an $\AND$ node, an $\OR$ node, or an Atomic node. Justify your reasoning before taking any action.

2. Then, choose ONE of the following options for the node:

   A. MARK node as Atomic if:  
   - the goal can be achieved using a single atomic action that strictly matches the {{NAVIGATION\_SPECIFICATION}}.
   OR 
   - the goal does not require any browser navigation actions like `type`, `click`, and `go\_back` and is information retrieved from thr webpage that can be noted down using the `note` action.

   B. EXPAND the node if:  
   - the goal is not atomic and requires executing a sequence of atomic actions from {{NAVIGATION\_SPECIFICATIONS}}.
   Clearly state the node type (AND/OR):  
    - Do not consider sub-goals and sub-strategies that cannot be executed on the webpage using one action or a sequence of actions from {{NAVIGATION\_SPECIFICATIONS}}.
    - For $\AND$ nodes, provide ordered list of logical sub-goals that are necessary for satisfying the node's objective. 
    - For $\OR$ nodes, provide observable list of alternative sub-strategies only.  
    - For $\OR$ nodes, order alternatives by likelihood of success, assigning scores within the range (0,1).
    - Do not add speculative or redundant sub-goals.

VERY IMPORTANT RULES:
- **Focus on expanding nodes in a way that will result in completing the task faster with high probability (Don't take very risky actions)**.

- Make sure that the temporal order of the children of an $\AND$ node is correct, detailed, and efficient. 
- Make sure to not split notes into multiple atomic actions.

- Strictly do not use `note` for noting future actions or sub-goals. For example, this is not a note -> `[Need to examine each search result for: 1. Publication date 2024 2. Review count 20+ 3. Confirm it's a Japan travel guide book]`

- Use `go\_back` action to return to previous page especially when you have consecutive atomic actions like `click`, `type`, `goto`.

- Atomic actions like `type`, `click` must be accompanied with valid IDs from the HTML DOM observation.

- Use atomic action `note` strictly to note information that is necessary for completing the task or for actions that cannot be executed on the webpage.
- Expand only non-atomic steps. Do not decompose goal when it is directly achievable using an action from {{NAVIGATION\_SPECIFICATIONS}}.

- Create $\OR$ nodes to consider different alternative strategies in cases where failure is probable.

- Structure plans as a knowledgeable efficient human would.

Output Format:

Begin by briefly describing the key elements from the input (overall task description, task progress, task feedback, task notes, node description, observation). Then analyze the node within the given context and clearly identifying its type. 

Do not change the node's intended objective or assume a different node objective. Your highest priority is satisfying the node's objective.

You can borrow notes from the notes summary if it helps in achieving the node's objective. 
Use the local tree information to ensure that you are not repeating any sub-goals that have already been considered.

If the node is an Atomic Action, determine and briefly analyze the 3 best possible actions it could represent. For each, explain why it fits the context. Then, select the single best action among them that is not very risky and will contribute to faster completion of the task and justify your choice. Conclude your response using ONE of the following formats:

FORMAT 1  (Enclose all values in <<>> on a single line):  

Node ID: <<node\_id>>  

Node Description: <<Describe what does the atomic action do>>  

Node Type: <<Atomic>>  

Expansion:  <<Exact Atomic action from 
{{NAVIGATION\_SPECIFICATIONS}}>>

Reasoning: <<Brief justification explaining why the node is classified as Atomic>> \enspace OR

FORMAT 2  (Enclose all values in <<>> on a single line, colon (;) separated):  

Node ID: <<node\_id>>  

Node Description: <<node\_description>>  

Node Type: <<AND / OR>>  

Expansion: <<1. First sub-goal; 2. Second sub-goal; 3. Additional sub-goals>> OR <<1. Alternative strategy; 2. Alternative strategy; Additional alternatives as necessary;...>>  

Reasoning: <<Brief justification explaining why the node is classified as AND, OR, or Atomic, based on current observations.>>

Example 1:
Node ID: <<2.1>>

Node Description: <<Find recipe with visible ratings and review counts>>

Node Type: <<Atomic>>

Expansion: <<click [123]>>

Example 2:

Node ID: <<2>>

Node Description: <<Filter results by rating and number of reviews>>

Node Type: <<AND>>

Expansion: <<1. click [Recipe with 69 ratings to check star rating]; 2. go\_back; 3. click [Check page 2 if first recipe doesn't meet criteria]; 4. go\_back; 5. click [Check page 3 if needed]>>

Example 3:

Node ID: <<0.5.1>>

Node Description: <<Filter recipes by rating and review count>>

Node Type: <<OR>>

Expansion: <<1. Manually scan recipe listings for review counts and ratings (score: 0.9), 2. Click through to individual recipes to check ratings and reviews (score: 0.8)>>

Do not output anything outside of the specified format. Atomic actions must strictly match format specified in {{NAVIGATION\_SPECICATIONS}}.

NOTE: Atomic actions `click` and `type` must be accompanied with valid IDs from the HTML DOM observation.

ROOT-LEVEL TASK DESCRIPTION: 

\{task\_description\}

ITEM LEVEL CONSTRAINTS:

\{item\_constraints\} \{additional\_context\}

OBSERVATION: 
\{observation\}

NODE\_ID : Description:
{node\_id} : \{node\_description\}

LOCAL TREE INFORMATION:
\{local\_tree\_info\} \\
\bottomrule
\end{tabular}
\caption{Prompt for node expansion.}
\label{nodeexpansion}
\end{table*}
\endgroup

\begingroup
\begin{table*}[!htp]
\tiny
\centering
\begin{tabular}{p{0.9\linewidth}}
\toprule
You are an $\ANDOR$ Tree Completeness Evaluator Agent specialized in web-browsing tasks. Your role is to assess whether a given $\AND$ node's objective has been successfully executed by the child nodes.

Node definitions: 

Node Types <AND/ OR / Atomic>

- $\AND$ node: Represents an ordered list of logical sub-goals required to achieve the node's objective.

- $\OR$ node: Represents alternative sub-strategies (which can be other $\ANDOR$ nodes)

- Atomic Action: Single executable action strictly matching {{navigation\_specifications}} described below.     

Node status indicators:  

- Unvisited nodes marked UNVISITED  

- Pruned nodes marked PRUNED  

- Completed nodes marked SUCCESS  

- Temporarily failed nodes marked FAIL  

NAVIGATION\_SPECIFICATIONS:  

<LIST OF BROWSER ACTIONS>

You are provided:  

- The root-level task description  

- Item level constraints if any  

- node\_id and description of the node to analyze  

- Task progress summary  

- Notes summary: notes taken by the agent so far  
- A list of its children: each with an ID, description, and status (SUCCESS, VISITED, UNVISITED, or PRUNED)  

- Task feedback  

Your task is to determine whether a given $\AND$ node is complete.

Follow the steps below to make this determination:

1. Examine the current children of the $\AND$ node and their statuses.  

   - Identify each child sub-goal or atomic action.  
   
   - Check whether each is marked as SUCCESS, FAILURE, or INCOMPLETE.

2. Review the task progress, task feedback, and notes summary.  

   - Look for any additional information that confirms or contradicts the completion of the node’s objective.

3. Determine whether all required sub-goals or actions are present.  

   - Ensure no required sub-goals are missing. 
   
   - Confirm that the sub-goals appear in the correct order, as expected based on the task tree and the overall task objective.

4. Verify that each required sub-goal or atomic action is marked as SUCCESS.  

   - Do not assume completion unless it is explicitly indicated.  
   
   - Partial or INCOMPLETE statuses are not sufficient.

5. Conclude that the $\AND$ node is COMPLETE only if:  
   - All required sub-goals or atomic actions are present, in the correct order, and marked SUCCESS; OR  
   - The successful child nodes, task progress summary, and task feedback clearly indicate that the node’s objective has been fully achieved.  
   
   - **You are only required to satisfy explicitly stated constraints. Do not assume or enforce any implicit constraints.**

6. Mark the $\AND$ node as INCOMPLETE if:  

   - The node's objective is not fully achieved, or 
   
   - Any required sub-goal, step, or relevant 
   information is missing or not marked SUCCESS.

Special Case - Root Node Evaluation:  

If the node being evaluated is the root node, you must additionally verify whether the overall task objective is fully satisfied.  

- If the task requires recommending or evaluating items, then:  

  - You must iterate over each item that is being considered or presented as a recommendation.  
  
  - For each item, check whether it satisfies all task constraints explicitly provided in the task description or item-level constraints. 
  
  - Note that constraints that are not explicitly mentioned in the task need not be satisfied.  
  
  - The node is incomplete if any recommended item fails to meet required constraints or if evaluation is missing or partial.
  
  - The node is incomplete if the task is incomplete and some information explicitly required by the task is missing. 
  
  - Do not assume any additional implicit constraints than the ones explicitly specified. 

Begin by analyzing all the information provided (task progress summary, notes summary, node description, children description).  

Provide a brief explanation for your decision, citing any missing, out-of-order, or incomplete sub-goals if relevant.

End your response in strictly one of the following formats, using a single line (reasoning $\AND$ node id must be enclosed in <<>>):

If node's objective is achieved:

COMPLETE <<node\_id>>  

Reasoning: <<Reason why node's objective is achieved by the agent. Cite references from task progress summary, notes summary and successful children's description>>

OR

If node's objective is not achieved:

INCOMPLETE <<node\_id>>  

Reasoning: <<Reason why node's objective is not achieved by the agent, Cite references from task progress summary, notes summary and successful children's description>>

Example 1:  

COMPLETE <<0.1>>  

Reasoning: <<All required sub-goals for finding a compact digital camera with specified requirements are present, in logical order, and marked as SUCCESS.>>

Example 2:  

INCOMPLETE <<0.2>>  

Reasoning: <<The sub-goal to compare prices is marked as PRUNED, and the sub-goal to check reviews is missing. The $\AND$ node's objective is not fully achieved.>>

You must strictly follow the above format.

\{additional\_context\}

NODE ID : DESCRIPTION
{node\_id} : \{node\_description\}

CHILDREN
\{children\}\\
\bottomrule
\end{tabular}
\caption{Prompt for evaluating if an $\AND$ node's objective has been achieved.}
\label{tab:nodecompletion}
\end{table*}
\endgroup

\begingroup
\begin{table*}[!htp]
\tiny
\centering
\begin{tabular}{p{0.9\linewidth}}
\toprule
You are a Global $\ANDOR$ Tree Update Agent revising an existing planning tree for a web-browsing task based on current available information so that the task is executed more efficiently. Do not change the ordering of the sub-plans.

Node definitions: 
Node Types <AND/ OR / Atomic>
- $\AND$ node: Represents an ordered list of logical sub-goals required to achieve the node's objective.
- $\OR$ node: Represents alternative sub-strategies (which can be other $\ANDOR$ nodes)
- Atomic Action: Single executable action strictly matching {{navigation\_specifications}} described below.      
Node status indicators: 
- Unvisited nodes marked UNVISITED  
- Pruned nodes marked PRUNED  
- Completed nodes marked SUCCESS 
- Temporarily failed nodes marked FAIL

NAVIGATION\_SPECIFICATIONS:
<LIST OF BROWSER ACTIONS>

You are provided:
-The root-level task description
-Current $\ANDOR$ tree description
-HTML DOM structure as the observation  
- Task progress summary
- Notes summary: Summary of notes taken by the agent during the task

Your task is

to carefully analyze all the information provided and determine which nodes to prune and which nodes to update.

Apply changes in this strict order:

1. PRUNE nodes that are no longer relevant or are duplicates.
2. UPDATE node descriptions if intent is unchanged but content needs minor revision.

VERY IMPORTANT RULES:
- You are only allowed to PRUNE or UPDATE nodes that have not been deleted, pruned, or marked successful. 
- Do not add status of the node while updating the description.
- Only use existing node IDs from the current tree; do not create new node IDs or subtrees.
- Do not PRUNE important observations relevant to the task made in Atomic nodes with action `note`.
- Do NOT PRUNE children that are necessary for satisfying the parent node's objective.
- Do not change node types.
- Do not change the ordering of the sub-plans.

First reason about the update to the tree based on the information given (task description, item constraints, task progress summary, notes summary, current $\ANDOR$ tree description, observation) to you and then given your answer in the following format. 
Use the $\ANDOR$ tree description to ensure that you are not repeating any sub-goals that have already been considered.
Format each instruction on a new single line in the format given below (All children and description must be enclosed in <<>>. Children should be colon (;) separated):

PRUNE [node\_id]
UPDATE [node\_id] <<Describe the node's new objective (sub-goal/strategy/description of action)>>

Example:

PRUNE [1.2]
PRUNE [1.3]
UPDATE [0.1.2] <<type eggless cake in search bar>>
UPDATE [0.2] <<Find an eggless cake recipe with over 60 votes and at least 4.5 star rating by examining search results>>

Do not split the list across multiple lines. 

 ROOT-LEVEL TASK DESCRIPTION:
 \{task\_description\}

\{additional\_context\}

AND/OR TREE DESCRIPTION: 
\{and\_or\_tree\_description\}

OBSERVATION:
\{observation\} \\
\bottomrule
\end{tabular}
\caption{Prompt for global tree update.}
\label{tab:globalupdate}
\end{table*}
\endgroup

\begingroup
\begin{table*}[!htp]
\tiny
\centering
\begin{tabular}{p{0.9\linewidth}}
\toprule
You are a Web-browsing Agent specialized in web-browsing tasks. You have taken notes while completing a given task and now you need to generate an output for the task based on the notes you have collected.  
You are provided: 
- Task description
- Sequence of notes taken during the execution of task
- Planning tree representation of the task

Node definitions: 
Node Types <AND/ OR / Atomic>
- $\AND$ node: Represents an ordered list of logical sub-goals required to achieve the node's objective.
- $\OR$ node: Represents alternative sub-strategies (which can be other $\ANDOR$ nodes)
- Atomic Action: Single executable action strictly matching {{navigation\_specifications}} described below.    
Node status indicators: 
- Unvisited nodes marked UNVISITED  
- Pruned nodes marked PRUNED  
- Completed nodes marked SUCCESS  
- Temporarily failed nodes marked FAIL

Your Task is to:

Give a detailed response that directly addresses the task description, using only the information from the notes and the planning tree presentation. 
Do not add any additional information or context that is not present in the notes. Even if the selected notes indicate that the task was not complete or certain information is missing or constraints are not met, provide the best possible response based on the available information. 
Do not make assumptions or inferences beyond what is explicitly stated in the notes.

VERY IMPORTANT RULES:
- Do not assume any details or context that is not explicitly mentioned in the notes.

Output Format:

Begin by analyzing the notes, then give your detailed final response in the following format.:

Task Response: <<Your detailed response to the task based on the notes. If the task was not completed, this may include incomplete information or recommendations that do not meet the constraints or requirements, if applicable. If the task was completed successfully, ensure all required details are specified. Do not skip any information that is required by the task.>>

Example:
Task Response: <<I successfully found a recipe for an eggless chocolate cake with over 60 votes and a rating of 4.5 stars. The recipe includes ingredients such as flour, sugar, cocoa powder, and baking soda. I also noted that the recipe requires 30 minutes of preparation time....>>

TASK DESCRIPTION:
\{task\_description\}

NOTES TAKEN DURING TASK EXECUTION: 
\{notes\}

\{additional\_context\}\\
\bottomrule
\end{tabular}
\caption{Prompt for generating final task response from notes taken during execution.}
\label{tab:taskresponse}
\end{table*}
\endgroup

\begingroup
\begin{table*}[!htp]
\tiny
\centering
\begin{tabular}{p{0.9\linewidth}}
\toprule
You are a Context Summarization and Critiquing Agent for web-browsing tasks.

Your job is to maintain an accurate, up-to-date understanding of task progress by analyzing:

- Task description  

- Item-level constraints (if any)  

- Task progress summary (if provided)  

- Observation history (if provided) 

- Action history (if provided)

- Notes summary: summary of notes taken by the agent so far (if provided)  

- Current observation (HTML DOM)

INSTRUCTIONS:

0. Do not infer any details or make assumptions.
1. Analyze all available inputs, especially the current HTML DOM.  

2. Ensure all summaries remain aligned with the main task objective.

3. Do not omit any information that may influence decisions or navigation.  

4. Make summaries detailed, well-structured, and actionable.  

5. Use the format below exactly.  

6. Enclose the response to each section in double angle brackets: `<< >>`.

Begin by analyzing the given context in detail. Then given your response stricly in the following format.
Your output must ensure continuity, preserve all essential details, and contribute to successful task completion.

RESPONSE FORMAT:

OBSERVATION SUMMARY

<<Describe the information from the CURRENT OBSERVATION. Emphasize elements and features that are relevant or potentially useful for fulfilling the task objective. Include all important detail.>>

OBSERVATION HIGHLIGHTS 

<<Single list of integer element IDs (i.e., [123, 8765, 345]) from the current DOM that are relevant for interaction or for revisiting this step later. Only include elements from the current page.  
Sort by relevance, with the most important listed first.>> 

TASK PROGRESS

<<Analyze the task progress using the task description, item-level constraints, past progress summaries, observation history, action history, and notes. 
First, summarize the key high-level steps the agent has actually taken toward completing the goal, focusing only on actions that were executed. 
Next, evaluate whether each explicitly stated task requirement or constraint has been satisfied—if the task involves item selection or recommendation, assess each item individually for compliance. 
Do not assume or infer unstated constraints. 
Finally, diagnose why the task has not yet been completed: identify any unsatisfied constraints, incorrect actions, or gaps in execution. Conclude with key takeaways or insights from the agent's exploration that are relevant to achieving successful task completion.>>

TASK FEEDBACK  

<<Based on the task progress summary, provide an outline of what the agent should focus on next to successfully complete the task. (2 sentences)>>

Example:

OBSERVATION SUMMARY
<<The page shows search results for "iPhone 12 Pro Blue 128GB" on Amazon. The first relevant listing is an Apple iPhone 12 Pro, 128GB in Pacific Blue (Renewed) for \$314.39. Multiple other iPhone models are also shown including iPhone 12, 13, and 14 in various colors and storage configurations....>>

OBSERVATION HIGHLIGHTS

<<[6028, 6033, 6042, 7204, 9277, 9280, 7242]>>

NEW NOTES

<<Refer to the task description and constraints, and identify all new information that can be used to completing the task. This should support future reference and continuity.
Do not assume any information or make up new details. All details must be grounded in observation and actions.
>>

TASK PROGRESS

<<The agent has successfully navigated to an Amazon search results page and identified a relevant product listing matching the task requirements. Specifically, the agent located an Apple iPhone 12 Pro with 128GB of storage in Pacific Blue, listed for \$314.39 in renewed condition and fully unlocked. 
This matches the explicitly stated constraints for model, color, and storage capacity. The observation and notes confirm that the identified product has a strong customer rating (4.1 out of 5 from over 12,000 ratings), further supporting its relevance. 
All explicitly defined task constraints appear to be satisfied: model (iPhone 12 Pro), storage (128GB), color (Pacific Blue), fully unlocked, and condition (renewed). 
However, the task has not yet been completed. The agent has not taken action to click on the product listing, verify the full product details on the product page, or add the item to the cart. 
These remaining steps are essential for completing the task, as product details may vary across pages. 
The main takeaway is that the search and identification phase has been completed successfully, and the agent should now focus on verifying the product and adding it to the cart to fulfill the task requirements.>>

TASK FEEDBACK

<<The next step should be to click on the product listing to verify all details on the product page and then add it to the cart. 
The agent should ensure the exact model and specifications are confirmed before proceeding with the purchase. 
Focus on completing the final step of adding the item to cart.>>

Do not output anything outside of the specified format.

TASK DESCRIPTION:

\{task\_description\}

\{additional\_context\}

CURRENT OBSERVATION:

\{observation\} \\
\bottomrule
\end{tabular}
\caption{Prompt for generating observation summary and context summary.}
\label{tab:observationsummary}
\end{table*}
\endgroup

\begingroup
\begin{table*}[!htp]
\tiny
\centering
\begin{tabular}{p{0.9\linewidth}}
\toprule
You are an advanced web-browsing agent that solves web-browsing related tasks.

Your job is to generate notes base on current observation as well as formulate a response to the task query based on the notes and available information.

You are given:

- Task description  

- Item-level constraints (if any)  

- Task progress summary (if provided)  

- Action history (if provided)

- Notes: Notes taken so far (if provided)  

- Current observation (HTML DOM)

INSTRUCTIONS:

0. Do not infer any details or make assumptions.

1. Analyze all available inputs, especially the current HTML DOM.  

3. Do not omit any information that may influence decisions or navigation.  

5. Use the format below exactly.  

6. Enclose the response to each section in double angle brackets: `<< >>`.

Begin by analyzing the context in detail. Given your final answer strictly in the following format.

RESPONSE FORMAT:

NEW NOTES

<<Refer to the task description and constraints, and identify all new information that can be used to completing the task. This should support future reference and continuity.
Do not assume any information or make up new details. All details must be grounded in observation. Do not skip any important notes relevant to the task.
>>

TASK RESPONSE 

<<Give a detailed response (you may repeat information from the above notes) that directly addresses the task description and constraints (if any), using only the information from the previous notes,new notes, and the action history. 
Do not add any additional information or context that is not present in the available information. Even if the selected notes indicate that the task was not complete or certain information is missing or constraints are not met, provide the best possible response based on the available information. 
Do not make assumptions or inferences beyond what is explicitly stated in the notes.>>

Example:

NEW NOTES

<<Found relevant iPhone 12 Pro listing matching requirements:
- Model: iPhone 12 Pro
- Storage: 128GB
- Color: Pacific Blue
- Price: \$314.39
- Condition: Renewed
- Fully Unlocked
- Rating: 4.1/5 stars from 12,669 ratings>>

TASK RESPONSE

<<Based on the task requirements, I have found an Apple iPhone 12 Pro with 128GB of storage in Pacific Blue, fully unlocked, in renewed condition, and priced at \$314.39. The listing has a solid customer rating of 4.1 out of 5 from 12,669 reviews.>>

Do not output anything outside of the specified format. 

TASK DESCRIPTION:
\{task\_description\}

\{additional\_context\}

PREVIOUS NOTES:
\{notes\}

CURRENT OBSERVATION:
\{observation\} \\
\bottomrule
\end{tabular}
\caption{Prompt for generating notes summary.}
\label{tab:notessummary}
\end{table*}
\endgroup

\begingroup
\begin{table*}[!htp]
\tiny
\centering
\begin{tabular}{p{0.9\linewidth}}
\toprule
You are an advanced web-browsing agent. Your task is to maintain a table of candidate items by parsing the current webpage's HTML DOM and applying 'ADD', 'UPDATE', or 'DELETE' actions. Your goal is to keep the table accurate, up to date, and free of duplicates, based only on information explicitly visible in the DOM.

You are given:

-HTML DOM of the current webpage

- task description

- item-level constraints

VERY IMPORTANT

- Record all important details that are relevant to the task and can be used for completing the task. Include any additional relevant details.

- Only include keys whose values are **clearly visible** in the HTML DOM. Record the exact value corresponding to each key.

- Keys must match attributes listed under ITEMS AND CONSTRAINTS.

    IMPORTANT RULES

Use this table to keep track of potential candidate solutions that can be used in the future for task completion or if search fails.

    1. Matching and Deduplication

- Before using `ADD`, check whether the item already exists in the candidate table.

- An item is considered a duplicate if all its visible fields match an existing entry (even if the ID differs).

- If a match is found:

  - Use `UPDATE` with the existing ID.
  
  - Only include fields that are newly visible or changed.
  
  - Do **not** use `ADD` for duplicates.

    2. Field Visibility
- Only include fields with values explicitly visible in the HTML DOM.

- Do **not** guess or infer missing values.

- Add exact values instead of simply checking if constraints is satisfied.

- Do **not** include `"unknown"` or empty-string values for missing fields.

- Omit any field that is not explicitly visible.

3. Constraint Checking

    3.1 Basic Constraint Handling
- For each item, evaluate which required constraints are satisfied.

- If any are violated or missing:

  - Add `constraints\_not\_met` listing the missing/invalid fields (e.g., `lift range, tray type`)
  
  - Set `status: uncertain`
  
  - Add a brief `comment` (e.g., "missing lift range")

    3.2 Add Only When Most Constraints Are Met
- Only use `ADD` if the item satisfies **most** required constraints (at least 60\%).

- If fewer than half are satisfied:
  - Do **not** add the item.
  
  - If it already exists, consider using `UPDATE` or `DELETE`.

    4. When to Use `UPDATE`

Only use `UPDATE` when **at least one of the following is true**:

- A previously missing field has now been found (and was not already present)

- The item now meets all required constraints and can be marked `status: complete`

Do **not** update items:

- If nothing has changed

- If the values are already present and correct

- If the item is already marked `deleted`

    5. Deletion
- Use `DELETE` only if the item is **clearly invalid** — for example:

  - It is irrelevant
  
  - It is severely incomplete (missing most required fields)
  
  - It does not satisfy constraints and is not fixable
  
- Never issue `UPDATE` for an item that has already been deleted.

    6. ID Assignment

- When using `ADD`, assign a unique ID not already present in the table (e.g., `S102`, `CH014`).

    7. Item Validity

- The `{{item}}` name must match a known type defined under ITEMS AND CONSTRAINTS.

- Skip any item with an invalid or unrecognized type.

RESPONSE FORMAT

- Return only valid `ADD`, `UPDATE`, or `DELETE` actions — one per line.

- Do **not** return the HTML DOM or candidate table.

- Do **not** include explanation or commentary.

- If no action is needed, return an empty string.

Your response must strictly follow the following format.

Use one or more of the following actions, one per line:

ADD {{item}}:ID:{{UniqueID}}; key1:value1; key2:value2; ...

UPDATE {{item}}:{{ExistingID}} key1:value1; key2:value2; ...

DELETE {{item}} [{{ExistingID}}]

Note: {{item}} must match exactly to the items in {{item-level constraints}}. 

While adding new items, always include the name or the title of the item.

 EXAMPLES

ADD standing desk:ID:S102; Title:ErgoRise Desk; Price:\$299.99; surface finish:Matte; status:uncertain; comment:"missing tray type and lift range"

UPDATE standing desk:S101 lift range:"24\"-48\""; status:complete; comment:"lift range confirmed"

DELETE standing desk [S099]

TASK DESCRIPTION
\{task\_description\}

ITEMS AND CONSTRAINTS  
\{items\_and\_constraints\}

CURRENT CANDIDATE TABLES  
\{current\_table\}

WEB PAGE CONTENT 
\{current\_observation\}  \\
\bottomrule
\end{tabular}
\caption{Prompt for extracting candidate items to add to structured memory}
\label{tab:structuredmemory}
\end{table*}
\endgroup

\begingroup
\begin{table*}[!htp]
\tiny
\centering
\begin{tabular}{p{0.9\linewidth}}
\toprule
You are a web-browsing assistant. Given a user query, extract a list of items to retrieve and, for each item, list all relevant item-specific constraints such as brand, price, rating, features, delivery time, etc.

Include any other item attributes explicitly or implicitly mentioned in the query that affect the desirability, quality, or functionality of the item (e.g., surface finish, memory presets, power type, compatibility). Treat these as additional constraints.

Include all intrinsic item-specific constraints, even if mentioned implicitly or as part of comparison criteria (e.g., “differ in brand or surface finish” implies that brand and surface finish are relevant constraints). Treat such attributes as constraints to be listed.

Ignore any source-specific constraints (e.g., “from [a specific website]”) and general task-level instructions (e.g., “compare prices”, “top three results”). Only include intrinsic item-specific constraints.

Only include items to retrieve—ignore any actions to perform. Ensure each item listed is unique. If there are no items to retrieve, return an empty list.

Output your response strictly in the following JSON format:

{{
  "ITEMS": [
    {{
      "item": "<description of item>",
      "constraints": ["<constraint1>", "<constraint2>", "..."]
    }},
    ...
  ]
}}

Do not include any additional explanation. Always follow this format exactly.

TASK DESCRIPTION

\{task\_description\}

ITEMS AND CONSTRAINTS  

\{items\_and\_constraints\}

CURRENT CANDIDATE TABLES  

\{current\_table\}

WEB PAGE CONTENT 

\{current\_observation\}  \\
\bottomrule
\end{tabular}
\caption{Prompt for extracting item-level constraints and attributes from task description.}
\label{tab:itemconstraints}
\end{table*}
\endgroup

\begingroup
\begin{table*}[!htp]
\tiny
\centering
\begin{tabular}{p{0.9\linewidth}}
\toprule
You are a validation assistant for a candidate item table.

Your task is to verify whether each item satisfies the required constraints for its item type.  
If any constraint is violated or missing, issue an `UPDATE` or `DELETE` command to adjust the table accordingly.

INSTRUCTIONS

1. For each item:

   - Issue an `UPDATE` only if all of the following are true:
   
     - The item is not marked as deleted
     
     - One or more constraints are violated or missing
     
     - The current values of `constraints\_not\_met`, `status`, or `comment` are incorrect, incomplete, or missing
     
     - Use the following format:
     
       ```
       UPDATE {{item}}:{{ID}} constraints\_not\_met: <key1> <key2> ..., status: uncertain or complete; comment: "brief explanation"
       ```
       
     - You are only allowed to update the following fields:
       - `constraints\_not\_met`
       - `status`
       - `comment`
       
     - Do not issue duplicate updates. Skip updates if no changes are needed.

   - Issue a `DELETE` only if:
   
     - The item is marked as complete but fails to meet any constraint
     
     - Or the item is mostly complete but violates most required constraints
     
     - Use this format:
       ```
       DELETE {{item}} [{{ID}}]
       ```

   - Do not delete items that are mostly incomplete or missing most of the required fields. These should be updated instead.

2. Item Validity
   - The `{{item}}` in every `UPDATE` or `DELETE` command must exactly match a valid item type defined under `ITEMS AND CONSTRAINTS`.
   
   - Skip any action for items with invalid or unrecognized types.

3. If the item satisfies all constraints and is already marked `status: complete`, or if the item is incomplete but does not require any changes, return nothing.

RESPONSE FORMAT  

- Return only `UPDATE` and `DELETE` commands, one per line.  

- Do not return the full item rows or any explanation.  

- If no action is needed, return an empty string.

- Use one or more of the following actions, one per line.

- Note: {{item}} must match exactly to the items in {{item-level constraints}}.

Your response must strictly follow the following format.

Use one or more of the following actions, one per line:

UPDATE {{item}}:{{ExistingID}} key1:value1; key2:value2; ...

DELETE {{item}} [{{ExistingID}}]

EXAMPLES

UPDATE standing desk:SD002 constraints\_not\_met: <lift range> <tray type>; status: uncertain; comment: "Missing lift range and tray type"  

DELETE standing desk [SD010]

TASK DESCRIPTION

\{task\_description\}

ITEMS AND CONSTRAINTS

\{items\_and\_constraints\}

CANDIDATE TABLE: 

\{current\_table\} \\
\bottomrule
\end{tabular}
\caption{Prompt for updating structured memory}
\label{tab:updatememory}
\end{table*}
\endgroup

% \begingroup
% \begin{table*}[!htp]
% \tiny
% \centering
% \begin{tabular}{p{0.9\linewidth}}
% \toprule

% \end{tabular}
% \caption{Test}
% \label{tab:example1}
% \end{table*}
% \endgroup

\begingroup
\begin{table*}[!htp]
\tiny
\centering
\begin{tabular}{p{0.9\linewidth}}
\toprule
You are an $\ANDOR$ Tree Repairer Agent specialized in web-browsing tasks. Your job is to repair a node in an $\ANDOR$ planning tree derived from HTML DOM structure observations.

Node definitions: 

Node Types <AND/ OR / Atomic>

- $\AND$ node: Represents an ordered list of logical sub-goals required to achieve the node's objective.

- $\OR$ node: Represents alternative sub-strategies (which can be other $\ANDOR$ nodes)

- Atomic Action: Single executable action strictly matching {{navigation\_specifications}} described below.      

Node status indicators: 

- Unvisited nodes marked UNVISITED  

- Pruned nodes marked PRUNED  

- Completed nodes marked SUCCESS  

- Temporarily failed nodes marked FAIL

NAVIGATION\_SPECIFICATIONS:

<LIST OF BROWSER ACTIONS>

You are provided: 

- root-level task description

- task constraints at item level
- HTML DOM structure as the observation  
- node\_id and desription of the node to analyze
- A list of its children: each with an ID, description, and status (SUCCESS, VISITED, UNVISITED, or PRUNED)
- A success$\OR$ node with id and description, if present
- Task progress summary of the task
- Notes summary of the task : Notes taken by the agent so far

Your task:

Your task is to review all the information provided including reasoning for repair (if available) and use it to repair the node. Make sure you prune infeasible sub-goals (which have not been marked successful or pruned) and add sub-goals that are necessary for completing the node's objective:

Repair the node by pruning or adding children to the node: Remove old infeasible children and add new children that were not previously used.

VERY IMPORTANT:
-Assess node repairability using node type, child status, and remaining unmet goals.
-Add only new, untried children , not already present or marked SUCCESS or PRUNED. These children will be appended to existing children.
-Each child must be minimal, concrete, and justified by the current DOM.
-For $\OR$ nodes: Suggest only new, obvious alternatives from the DOM.
-For $\AND$ nodes: Suggest only missing or incomplete sub-goals in correct order.
-Do NOT PRUNE children that are necessary for satisfying the parent node's objective.

First reason about the repair and then give the final answer in the following format (Enclose all values in <<>> on a single line, colon (;) separated. Ordering: Prune first, then add children):

PRUNE [node\_id of child]
ADD [repair\_node\_id] repair\_node\_type : <<1. First added sub-goal; 2. Second added sub-goal; 3. Additional added sub-goals as necessary>> OR ADD [node\_id] node\_type : <<1. Added alternative strategy 1 (score: 0.x); 2. Added alternative strategy 2 (score: 0.y); Added additional alternatives as necessary>> 
Reasoning <<Reason for the nodes pruned and added.>>

Example 1:
PRUNE [1246]
PRUNE [1245]
ADD [1244] AND : <<1. Go north for 5 meters; 2. Turn left at the intersection; 3. Walk for 10 minutes>>>
OR
ADD [1244] OR : <<1. Open the door (score: 0.9); 2. Pick up the key (score: 0.8)>>

Do not re-add processed or pruned children.
Note that you can only prune children of the node to repair and add children to the node to repair. 
Added children will be appended to existing successful children.
Respond using only the above. No extra text. Valid node\_types are AND/OR/Atomic . 

ROOT-LEVEL TASK DESCRIPTION: 
\{task\_description\}

\{additional\_context\}

OBSERVATION: 
\{observation\}

NODE ID : Description
\{node\_id\} \{node\_type\} : \{node\_description\}

NODE\_CHILDREN\_REPRESENTATION:
{children}

REASON FOR REPAIR
{reason\_for\_repair}

LOCAL TREE INFORMATION:
\{local\_tree\_info\} \\
\bottomrule
\end{tabular}
\caption{Prompt for node repair}
\label{tab:noderepair}
\end{table*}
\endgroup

\section{Limitations}\label{app:limitations}
\SA{} incurs higher computational overhead than flat agents due to additional LLM calls for node expansion, repair, and verification. The quality of the hierarchical plan depends on the backbone LLM's ability to produce meaningful decompositions; poor expansions can cascade into subtree failures. Our framework currently operates on text and HTML observations only, without visual understanding of rendered web pages.

\section{Ethical Considerations}\label{app:ethics}
This work presents a planning framework for web-browsing agents evaluated on publicly available benchmarks and commercial websites. We do not collect, store, or process personal user data. All experiments are conducted in sandboxed or read-only browsing environments and do not involve human subjects. The human annotator participated voluntarily and was compensated fairly for their time. Autonomous web agents can take real actions on live websites, such as submitting forms or completing purchases. Deployments outside sandboxed settings should therefore restrict the action space and require human confirmation for irreversible steps.

\section{Future Work}\label{app:futurework}
Several directions remain open. LLM-based scoring of \OR{} node alternatives could be replaced with learned value functions. Integrating vision-language models would extend the framework to visually rich interfaces. The framework could also be applied to other domains such as code generation and OS-level tasks. Finally, caching and policy distillation could reduce computational cost.

\section{POMDP Formalization}\label{app:pomdp}

We model the agent-browser interaction as a Partially Observable Markov Decision Process (POMDP) $\langle\mathcal{O},\mathcal{S},\mathcal{A},R,T, p_0,\gamma\rangle$~\cite{kaelbling1998planning}, where $o \in \mathcal{O}$ is an \emph{observation} comprising the task description, agent-constructed plans, the HTML DOM, and interaction metadata; $s \in \mathcal{S}$ is the true underlying \emph{state} of the browser and agent; and $a \in \mathcal{A}$ is an \emph{action}, either an atomic browser operation (click, scroll, type, go back) or a token sequence describing a sub-goal or strategy. The \emph{reward function} $R: \mathcal{S} \times \mathcal{A} \to \mathbb{R}$ assigns a scalar reward to each state-action pair, the \emph{transition function} $T: \mathcal{S} \times \mathcal{A} \to \Delta^{|\mathcal{S}|}$ defines a distribution over next states, $p_0 \in \Delta^{|\mathcal{S}|}$ is the \emph{initial state distribution}, and $\gamma \in (0,1)$ is the discount factor. The agent's \emph{policy} $\pi: \mathcal{S} \to \Delta^{|\mathcal{A}|}$ maps states to distributions over actions and is, in general, history-dependent~\citep{kaelbling1998planning}. We represent the agent's policy using an LLM. At each timestep $t$, the observation history $h_t = \{o_1, \dots, o_t\}$ is encoded as a token sequence $x_t = [x_1, \dots, x_m]$ serving as the input prompt. The action $a_t = [y_1, \dots, y_n]$ is then generated autoregressively as:
\[
    \Pr(a_t \mid h_t) = \prod_{i=1}^{n} \pi(y_i \mid x_1, \dots, x_m, y_1, \dots, y_{i-1}).
\]

\section{Detailed DFS Example Trace}\label{app:example_trace}

\begin{figure}[h]
    \centering
    \includegraphics[width=0.9\linewidth]{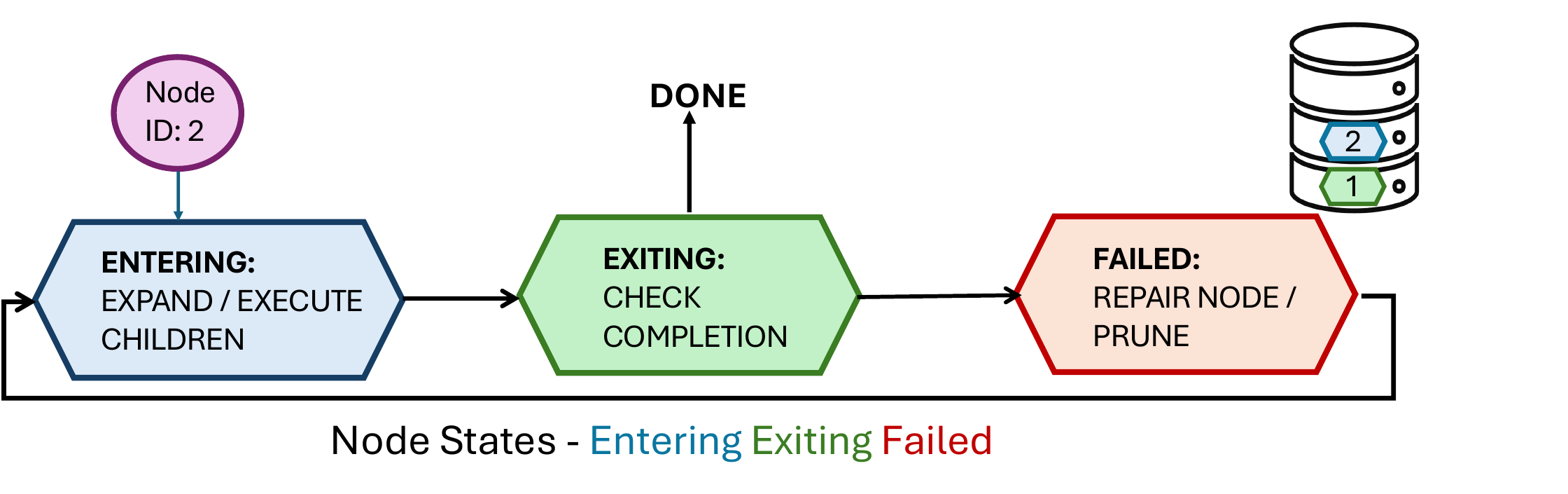}
   \caption{Illustration of node state transitions during iterative modified greedy depth-first search on AND-OR tree. Unlike traditional depth-first search, nodes may be revisited and repeatedly transition through \ENTERING{}, \EXITING{}, and \FAILED{} states.}
    \label{fig:dfs_states}
\end{figure}

We provide a step-by-step trace of the modified greedy DFS on the task \texttt{``Find a vegan chocolate brownie recipe with 4+ rating on allrecipes.com''} (illustrated in \cref{fig:framework}). The agent begins with only root node~$1$ on the stack in \ENTERING{}. After each \ACTION{} execution, \SUMMARIZER{} updates the internal context and \GLOBALTREE{} is invoked (making no tree modifications in this trace).

\textbf{Steps 1--2 [Expand root \& execute $1.1$]:}
The agent pops node~$1$ and invokes \NODEEXPANSION{}, which identifies it as an \AND{} node with three ordered children: $1.1$~(type query), $1.2$~(select recipe), and $1.3$~(note recipe), all of which must succeed. Node~$1$ is re-pushed in \EXITING{} and its children are pushed in \ENTERING{} in prescribed order. Node~$1.1$ is then popped, identified as \ACTION{}, executed, and marked \SUCCESS{}.

\textbf{Steps 3--4 [Expand $1.2$ \& execute $1.2.1$]:}
The agent pops node~$1.2$ and identifies it as an \OR{} node encoding alternative recipe-selection strategies, with children $1.2.1$~(Recipe~A, score~$1.0$) and $1.2.2$~(Recipe~B, score~$0.95$). Only the highest-ranked child~$1.2.1$ is pushed. It is popped, executed as \ACTION{} (clicking Recipe~A), and \NODECHECK{} confirms the \OR{} objective of $1.2$ is satisfied, so $1.2.2$ is never explored and $1.2$ is marked \SUCCESS{}.

\textbf{Steps 5--6 [Execute $1.3$ \& terminate]:}
Node~$1.3$ is popped, executed as \ACTION{} (noting the recipe), and marked \SUCCESS{}. Root node~$1$ is then popped in \EXITING{}; \NODECHECK{} confirms all three \AND{} children have succeeded. The stack empties and the agent terminates with \SUCCESS{}.

No node enters \FAILED{} in this trace, so repair and pruning are not triggered. Had node~$1.2$ exhausted both \OR{} alternatives without success, it would enter \FAILED{}, causing sibling~$1.3$ to be pruned and failure to propagate to root node~$1$, which would then invoke \NODEREPAIR{} if its revision budget permits.

\section{Tree and Agent Operations Details}\label{app:treeoperations}
% Duplicate of the main-body figure \ref{fig:treeoperations} (same image and
% caption) removed during the main+supplementary merge.

\begin{figure*}[h!]   
    \centering
    \includegraphics[width=0.8\textwidth]{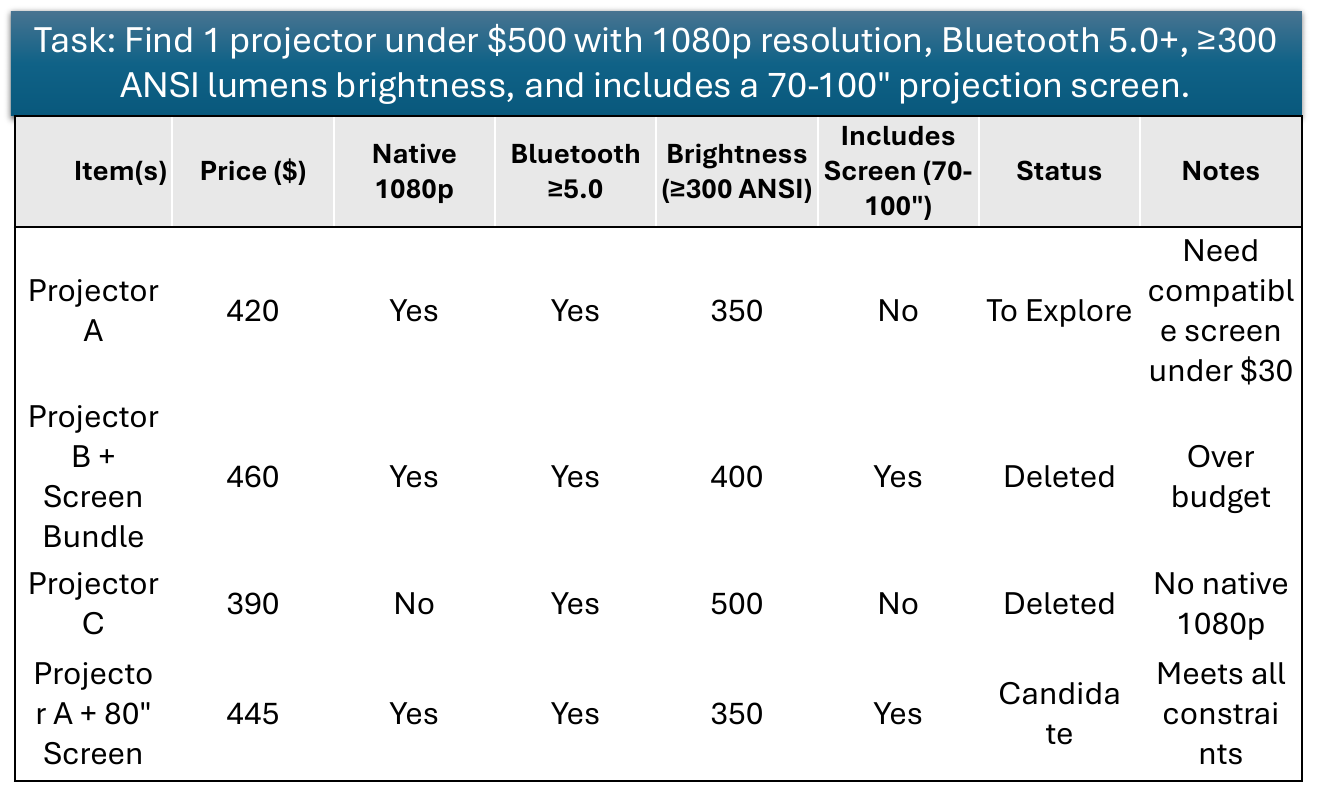}
    \caption{Example of Structured Memory used to track candidate solutions during information-seeking tasks.}
    \label{fig:memory}
\end{figure*}
 \subsection{Operator Details}\label{sec:operatordetails}
\textbf{\NODEEXPANSION:}
This operator identifies a node's type and, when appropriate, generates its children. For $\AND$ nodes, it generates sub-goal descriptions; for $\ACTION$ nodes, it produces a browser-level command; and for $\OR$ nodes, it outputs a set of candidate strategies along with their scores. The framework then updates the tree structure accordingly. If the expansion fails, the agent retries the operation up to a predefined limit. After repeated failures, the agent marks the node as \FAILED{}.

\textbf{\NODEREPAIR:}
This operator revises $\AND$ or $\OR$ nodes that have entered the \FAILED{} state. Conditioned on the node’s metadata and contextual summaries, it proposes modifications to the node’s children. When the node remains repairable, the operator recommends adding new children; otherwise, it instructs the framework to prune the node. If several repair attempts fail, the agent prunes the node.

\textbf{\GLOBALTREE:}
This operator updates the global tree structure based on newly observed information. Given the current tree and contextual metadata, it issues instructions for pruning irrelevant or infeasible subtrees and for updating node descriptions. This operator aggressively removes unpromising branches to maintain tractability. If repeated inference attempts fail, the operator recommends fallback pruning strategies.

\textbf{\NODECHECK:}
This operator evaluates whether an $\AND$ node has satisfied its objective based on the statuses of its children. The framework invokes this operator once all children have completed execution and at least one has succeeded. The operator assesses the node’s description, the task progress summary, the notes summary, and the task constraints, and returns a binary decision indicating whether the node’s objective has been achieved.

\textbf{\SUMMARIZER:}
This module maintains the agent’s internal representations and tracks its progress. This operator processes new observations to update the task progress summary, generate an observation summary, and provide guidance for subsequent actions. When the agent encounters a new webpage, this module extracts task-relevant information, updates the agent’s notes, and produces a revised notes summary. These summaries enable the agent to maintain an accurate understanding of the environment and support effective planning.

\begin{table*}[h]
\centering
\caption{Internal Representations Maintained by \SA{}}
\label{tab:internal_representations}
\begin{tabular}{lp{0.66\linewidth}}
\toprule
\textbf{Representation} & \textbf{Description} \\ \midrule
Interaction History & Logs all observation summaries and actions up to the current time step \\
Task Progress Summary & Tracks progress toward the main objective \\
Notes Summary & Aggregates information from all past observations that are relevant to the task \\
Task Constraints & Encodes global and item-level requirements from the task description \\
Observation Summary & Distills salient information from the current web page at each step \\
Action History & Lists all executed browser actions \\ \bottomrule
\end{tabular}
\end{table*}

\section{Human-in-the-Loop Intervention}\label{app:intervention}

Here we provide an extended illustration of how our framework's interpretable hierarchical structure supports human-in-the-loop oversight.

\begin{figure*}
    \centering
    \includegraphics[width=0.9\linewidth]{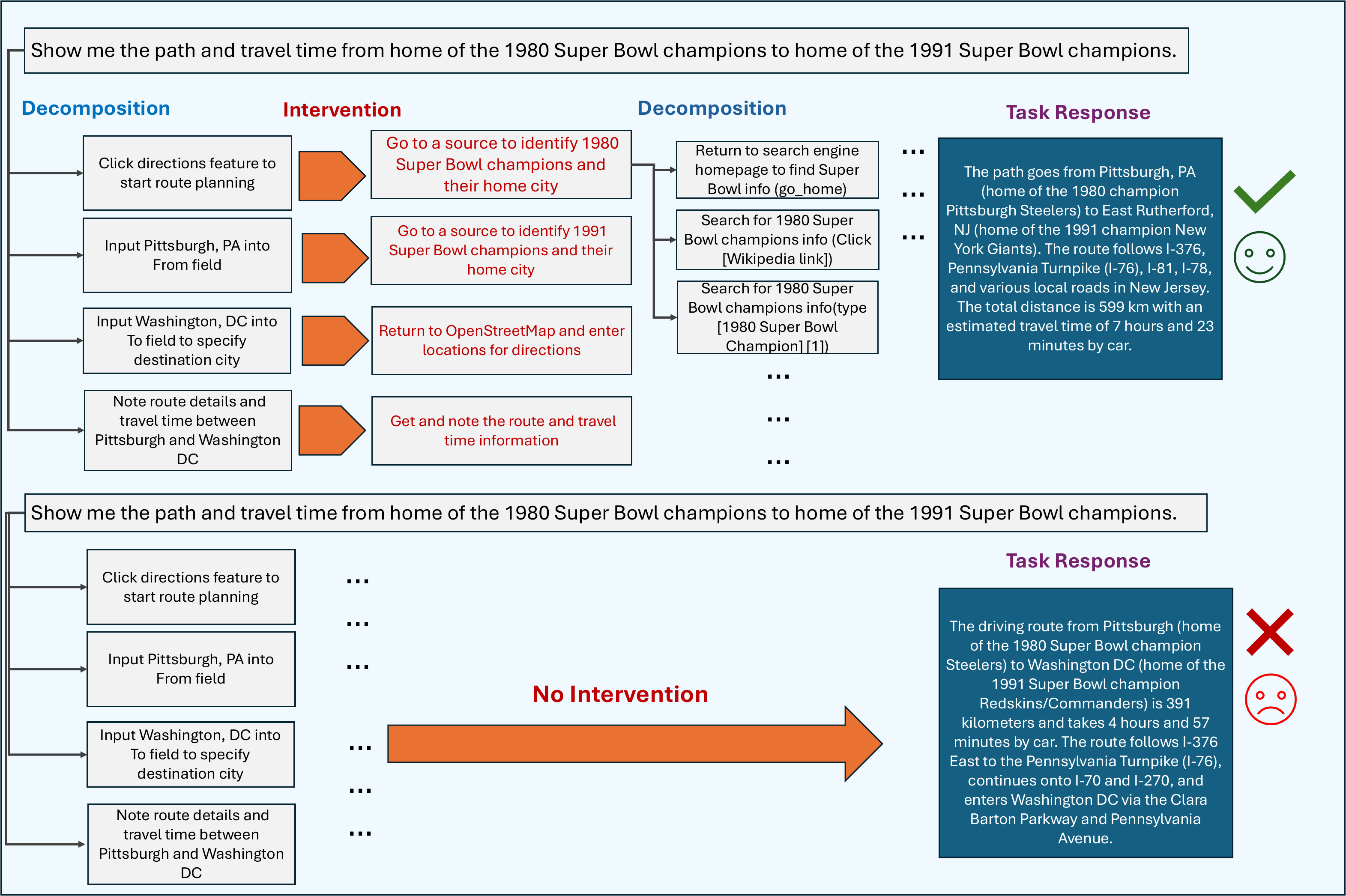}
\caption{\textbf{Human Decomposition Intervention in AND/OR Tree Planning to Correct Bad Subtask Decomposition.} 
The diagram illustrates how a human decomposition intervention can redirect an AI agent that has 
produced an incorrect sub-goal breakdown for a multi-step task. The task requires finding the 
driving route between the home cities of the 1980 and 1991 Super Bowl champions. 
\textit{Without intervention} (bottom), the agent incorrectly assumes destinations a priori, defaulting 
to Pittsburgh, PA and Washington, DC, without first verifying which teams won in those 
respective years, resulting in a hallucinated but confidently stated incorrect answer. 
\textit{With intervention} (top), a human intercepts the agent's flawed decomposition and injects 
three corrective sub-goals into the AND/OR tree: (1) identify the 1980 Super Bowl 
champions and their home city, (2) identify the 1991 Super Bowl champions and their home 
city, and (3) return to the mapping tool with the verified locations. This human-guided correction 
grounds the plan in factual information before execution continues, yielding the correct route from Pittsburgh, PA 
(Pittsburgh Steelers) to East Rutherford, NJ (New York Giants) at 599\,km and approximately 
7 hours 23 minutes. For simplicity, we illustrate the human intervention at the first level of the 
AND/OR planning tree; however, such human decomposition interventions can be applied 
at \textit{any level} of the tree, allowing a human operator to catch and correct errors in 
sub-goal formulation at whatever level of abstraction they arise.}
    \label{fig:intervention}
\end{figure*}

\section{Algorithm Details}\label{app:algorithm}
\subsection{Planning and Execution Algorithm}\label{sec:planning_detail}
The agent begins by pushing the root node which represents the given task, onto the DFS stack in the \emph{ENTERING} state. As in classical iterative DFS, the algorithm then iteratively pops the top node from the stack and processes it according to its current state.

When processing a node in the \ENTERING{} state, the algorithm first checks the node’s status. By default, new nodes are marked as unvisited. If the node is encountered for the first time, the algorithm invokes the Node Expansion module to determine its type and expand it. For $\AND$ nodes, the module generates an ordered list of children according to task constraints. For $\OR$ nodes, it ranks alternatives by their estimated likelihood of success, as determined by LLM scoring or a value function. For $\ACTION$ nodes, it outputs the precise browser command to execute. After expansion, or if the node was previously expanded, the agent marks the node as visited and re-adds it to the stack in the \emph{EXITING} state.

Next, the agent proceeds according to node type. For $\AND$ nodes, it pushes each child onto the stack in the \ENTERING{} state for further processing. For $\OR$ nodes, it pushes only the most promising unprocessed child in the \ENTERING{} state. If the node has no children, no further action is taken until its status is evaluated in the \EXITING{} state. For $\ACTION$ nodes, the agent immediately executes the specified browser command. If the command is successful, the agent marks the node as successful, updates its internal context via the Observation and Notes Summarization module. The Global Tree Update module prunes irrelevant branches and refines node descriptions as needed, propagating these changes upward to keep the stack and tree consistent. If the $\ACTION$ node fails, the agent marks it as \emph{FAILED}.

When processing a node in the \EXITING{} state, the agent evaluates completion criteria based on the node type. For $\AND$ nodes, the agent verifies that all valid, that is, unpruned or undeleted, child nodes have succeeded and invokes the Node Completion module to ensure that the overall objective of the $\AND$ node is met. Although this check is theoretically applicable to all $\AND$ nodes, we found it to be overly strict in practice. Therefore, to avoid unnecessary retries and failures, we apply this strict completion check only at the root node in our implementation.
If completion is verified, the agent marks the node as successful, otherwise, as \emph{FAILED}. For $\OR$ nodes, the agent checks if at least one child has succeeded, marking the node as \emph{SUCCESS} if so, and as \emph{FAILED} otherwise. For $\ACTION$ nodes, the agent sets the status to success or \emph{FAILED} depending on the outcome of the browser operation. Once a node is marked as success, it is not processed again.

If the agent processes a node in the \FAILED{} state, it attempts to repair the node or marks it as pruned. For $\ACTION$ nodes, it prunes the node and updates the stack. If a failed $\ACTION$ node belongs to an $\AND$ node, the agent deletes all remaining unexecuted siblings, since the conjunctive requirement cannot be satisfied. For $\AND$ nodes, if none of the children have achieved the node’s objective, the agent checks the node’s revision count. If the revision limit is not reached, the Node Repair module attempts to generate new sub-goals or modify the set of children. If repair fails or the revision limit is reached, the agent prunes the $\AND$ node and all its descendants, propagating failure upward and updating the stack.
For $\OR$ nodes in the \FAILED{} state, the agent first checks for any remaining unprocessed children. If such children exist, it selects and pushes the most promising one onto the stack in the \ENTERING{} state. If all alternatives are exhausted, it attempts node repair if within the revision limit, otherwise, it prunes the node. As with $\AND$ nodes, any pruning or failure of an $\OR$ node propagates upward, ensuring the global plan remains consistent and that failures at lower levels trigger repair or pruning at higher levels.

This DFS-based algorithm allows the agent to incrementally build and adapt its hierarchical plan, efficiently responding to new observations and failures, while executing the task.

% \subsection{Algorithm Details}\label{app:algorithm}

\begin{algorithm}[t]
\scriptsize
\caption{PROCESS\_NODE\_ENTERING(node, stack)}\label{alg:entering}
\SetAlgoLined

\tcp{Perform rollback when parent is an OR node}
\If{node.parent and node.parent.type == NodeType.OR}{
    success $\gets$ PERFORM\_ROLLBACK(node.parent)\;

    \If{success == true}{
        currentUrl $\gets$ node.parent.url\;
    }
}

\tcp{Increase node execution count}
node.execution\_count $\gets$ node.execution\_count + 1\;

\tcp{If node hasn't yet been visited, expand it}
\If{node.type == NodeType.UNKNOWN and node.status $\notin$ CLOSED\_STATUSES}{
    \tcp{Populate the node type}
    node, \_, ACTION $\gets$ POPULATE\_NODE\_TYPE(node)\;
}

node.status $\gets$ NodeStatus.VISITED\;

\tcp{Process node based on its type}
\If{node.type == NodeType.ACTION}{
    stack, node $\gets$ PROCESS\_ACTION\_NODE(node, stack)\; \tcp*{Algorithm~\ref{alg:action}}
}
\ElseIf{node.type == NodeType.AND}{
    stack, node $\gets$ PROCESS\_AND\_NODE(node, stack)\; \tcp*{Algorithm~\ref{alg:and}}
}
\ElseIf{node.type == NodeType.OR}{
    stack, node $\gets$ PROCESS\_OR\_NODE(node, stack)\; \tcp*{Algorithm~\ref{alg:or}}
}

\Return{stack, node}\;

\end{algorithm}

\begin{algorithm}[t]
\scriptsize
\linespread{0.9}\selectfont
\caption{PROCESS\_NODE\_EXITING(node, stack)}\label{alg:exiting}
\SetAlgoLined

\tcp{ACTION NODE IN EXITING STATE}
% \tcp{If node is atomic action, check if execution was successful}
% \tcp{If successful, mark as success; else, mark as failed and push to FAILED state}
\If{node.type == NodeType.ACTION}{
    \tcp{Put them back into the stack in failure mode}
    \If{node.status $\in$ FAILED\_OR\_PRUNED}{
        stack.append((node, NodeState.FAILED))\;
    }
    \Else{
        node.status $\gets$ NodeStatus.SUCCESS\;
    }
}

\tcp{AND NODE IN EXITING STATE}

\ElseIf{node.type == NodeType.AND}{
    % \tcp{Check if children are successful}
    success $\gets$ IS\_SUCCESSFUL\_AND(node)\;
    
    \If{success == true}{
        \tcp{If node revision has not reached max limit, check if node is complete}
        \tcp{If not complete, repair the node}
        \If{(node.revision\_count $<$ MAX\_REVISION\_COUNT)}{
            isComplete, reasoning $\gets$ CHECK\_FOR\_COMPLETION\_AND(node)\;
            node.reasoning $\gets$ reasoning\;
            
            \If{isComplete == true}{
                node.status $\gets$ NodeStatus.SUCCESS\;
                \textbf{continue}\;
            }
            \Else{
                \tcp{Is complete check failed, mark node as failed and push it on the stack in FAIL state}
                node.status $\gets$ NodeStatus.FAIL\;
                node.reasoning $\gets$ reasoning\;
                stack.append((node, NodeState.FAILED))\;
                \textbf{continue}\;
            }
        }
        \Else{
            \tcp{If node has already been revised sufficient number of times and the children are successful, mark the node as successful}
            node.status $\gets$ NodeStatus.SUCCESS\;
            \textbf{continue}\;
        }
    }
    \Else{
        % \tcp{Children don't appear to be successful}
        \tcp{Mark the node as failed and push it onto the stack in FAIL state}
        node.status $\gets$ NodeStatus.FAIL\;
        stack.append((node, NodeState.FAILED))\;
    }
}

\tcp{OR NODE IN EXITING STATE}
% \tcp{Check if any child of the OR node is successful}
% \tcp{If not, mark it as failed and push it onto the stack in FAIL state}
\ElseIf{node.type == NodeType.OR}{
    success $\gets$ IS\_SUCCESSFUL\_OR(node)\;
    
    \If{success == true}{
        node.status $\gets$ NodeStatus.SUCCESS\;
    }
    \Else{
        node.status $\gets$ NodeStatus.FAIL\;
        stack.append((node, NodeState.FAILED))\;
    }
}

\Return{stack, node}\;

\end{algorithm}

\begin{algorithm}[t]
\scriptsize
\caption{PROCESS\_NODE\_FAILED(node, stack)}\label{alg:failed}
\SetAlgoLined

\tcp{ACTION NODE IN FAILED STATE}
\If{node.type == NodeType.ACTION}{
    node.status $\gets$ NodeStatus.PRUNED\;
    BACKTRACK\_FAILURE(node)\;
}
\tcp{Process node based on its type}
\ElseIf{node.type == NodeType.AND}{
    stack, node $\gets$ PROCESS\_FAILED\_AND\_NODE(node, stack)\; \tcp*{Algorithm~\ref{alg:failed_and}}
}
\ElseIf{node.type == NodeType.OR}{
    stack, node $\gets$ PROCESS\_FAILED\_OR\_NODE(node, stack)\; \tcp*{Algorithm~\ref{alg:failed_or}}
}

\Return{stack, node}\;

\end{algorithm}

\begin{algorithm}
\scriptsize
\caption{BACKTRACK\_FAILURE(node)}
\label{alg:backtrack}
\SetAlgoLined
\If{node.parent == null}{
    \Return\; \tcp*{No parent to backtrack to}
}
\If{node.parent.type == NodeType.AND and node.parent.ordered == true}{
    \tcp{Handle ordered AND node: delete remaining siblings}
    remaining $\gets$ GET\_REMAINING\_EXCLUDING\_NODE(node, node.parent)\;
    undeletedChildren $\gets$ []\;
    \ForEach{rem in remaining}{
        \If{rem.status != NodeStatus.DELETED}{
            undeletedChildren.append(rem)\; \tcp*{Filter undeleted children}
        }
    }
    \If{undeletedChildren not empty}{
        \ForEach{rem in undeletedChildren}{
            rem.status $\gets$ NodeStatus.DELETED\; \tcp*{Mark as deleted}
        }
        % \tcp{Delete descendants via TreeManager}
        deletedChildren $\gets$ RECURSIVELY\_DELETE\_CHILDREN(undeletedChildren)\;
        \If{stack not empty}{
            stack $\gets$ PURGE\_STACK(deletedChildren, stack)\; \tcp*{Remove from stack}
        }
    }
}

\end{algorithm}

\begin{algorithm}[t]
\scriptsize
\caption{PROCESS\_ACTION\_NODE(node, stack)}\label{alg:action}
\SetAlgoLined

\tcp{If node represents an ACTION, execute it. If successful, globally update tree}
status $\gets$ null\;

\tcp{Before processing any node, push it back onto stack in EXITING state}
stack.append((node, NodeState.EXITING))\;

success $\gets$ false\;

\tcp{Execute action with retry logic}
\While{node.retry\_count $<$ MAX\_RETRY\_COUNT}{
    \tcp{Try block for attempting to perform action}
    \If{NO\_EXCEPTION}{
        action, status $\gets$ PERFORM\_ACTION(node)\;
        
        \If{env.steps $>$ MAX\_STEPS}{
            \Return{stack, node}\; \tcp*{Reached maximum steps}
        }
        
        \If{env.done()}{
            \Return{stack, node}\; \tcp*{Environment done after performing action}
        }
        
        \tcp{Check if action was successfully executed}   
            assert(status is True)\;
        
        \tcp{Extract notes from the action if there are any}
        UPDATE\_NOTES(node.action)\;
        
        \tcp{Mark ACTION node as successful}
        node.status $\gets$ NodeStatus.SUCCESS\;
        success $\gets$ true\;
        
        \tcp{If ACTION has resulted in change inn current observation}
        isNoteAction $\gets$ IS\_NOTE\_ACTION(node.action)\;
        
        \If{not isNoteAction}{
            \tcp{Update notes and observation summaries}
            ASYNC\_FULL\_UPDATE()\;
            STEP $\gets$ STEP + 1\;
            
            \tcp{Update URL of current node and current URL}
            currentUrl $\gets$ env.get\_url()\;
            UPDATE\_URL(node)\;

        }
        
        node.retry\_count $\gets$ node.retry\_count + 1\;
        \textbf{break}\;
    }
    
    \tcp{If exception occurred, update retry count and continue}
    node.retry\_count $\gets$ node.retry\_count + 1\;
}
\tcp{If success is false, update status to fail}
\If{success == false}{
    node.status $\gets$ NodeStatus.FAIL\;

}

\Return{stack, node}\;

\end{algorithm}

\begin{algorithm}[t]
\scriptsize
\caption{PROCESS\_AND\_NODE(node, stack)}\label{alg:and}
\SetAlgoLined

\tcp{For an AND node, process all unprocessed children and put them on stack}

\tcp{First push the node back onto stack in EXITING state}
\tcp{so that control returns to it once children have been executed}
stack.append((node, NodeState.EXITING))\;

children $\gets$ node.children\;

\tcp{Check if the AND node is valid}
isValid $\gets$ IS\_VALID\_AND(children)\;

\tcp{Check if the AND node is successful}
success $\gets$ IS\_SUCCESSFUL\_AND(node)\;

\tcp{If successful, mark node as successful}
\If{success}{
    node.status $\gets$ NodeStatus.SUCCESS\;
}
\ElseIf{not isValid}{
    \tcp{I there are no valid children, mark it as fail}
    node.status $\gets$ NodeStatus.FAIL\;
}
\Else{
    \tcp{Put unprocessed children in stack in reversed order}
    \ForEach{child in reverse(node.children)}{
        \If{child.status $\notin$ CLOSED\_STATUSES}{
            stack.append((child, NodeState.ENTERING))\;
        }
    }
}

\Return{stack, node}\;

\end{algorithm}

\begin{algorithm}[t]
\scriptsize
\caption{PROCESS\_OR\_NODE(node, stack)}\label{alg:or}
\SetAlgoLined

\tcp{For an OR node, process next promising child}

\tcp{First push node back onto stack in EXITING state}
\tcp{to return control to it after child has been processed}
stack.append((node, NodeState.EXITING))\;

children $\gets$ node.children\;

\tcp{Check if node is successful}
success $\gets$ IS\_SUCCESSFUL\_OR(node)\;

\tcp{There should be one valid child}
isValid $\gets$ IS\_VALID\_OR(children)\;

\tcp{Mark the node as successful}
\If{success}{
    node.status $\gets$ NodeStatus.SUCCESS\;
}
\tcp{If no children are available to process, mark node as failed}
% \tcp{and exiting state will handle this failure}
\ElseIf{not isValid}{
    node.status $\gets$ NodeStatus.FAIL\;
}
\Else{
    \tcp{Find next promising child (children ordered by estimated likelihood of success)}
    child $\gets$ FIND\_NEXT\_PROMISING(children)\;
    
    \tcp{Push promising child onto the stack}
    stack.append((child, NodeState.ENTERING))\;
}

\Return{stack, node}\;

\end{algorithm}

\begin{algorithm}[t]
\scriptsize
\linespread{0.82}\selectfont
\caption{PROCESS\_FAILED\_AND\_NODE(node, stack)}\label{alg:failed_and}
\SetAlgoLined

\tcp{For an AND node in FAILED state, revise, prune, or backtrack}

\tcp{Initialize tracking variables}
delChildren $\gets$ []\;
addedChildren $\gets$ []\;
prunedChildren $\gets$ []\;
revised $\gets$ false\;

isValid $\gets$ IS\_VALID\_AND(node.children)\;

\tcp{If node is invalid but has at least one successful child, check if the node's objective has been completed}
\If{not isValid}{
    reasoning $\gets$ ""\;
    hasOneSuccess $\gets$ HAS\_ATLEAST\_ONE\_SUCCESS(node)\;
    
    \If{hasOneSuccess}{
        isComplete, reasoning $\gets$ CHECK\_FOR\_COMPLETION\_AND(node)\;
        node.reasoning $\gets$ reasoning\;
        
        \If{isComplete and isComplete == true}{
            node.status $\gets$ NodeStatus.SUCCESS\;
            \Return{stack, node}\;
        }
    }
    
    \tcp{If node revision has not reached the maximum limit, repair it}
    node.status $\gets$ NodeStatus.FAIL\;
    
    \If{(len(node.children) $<$ MAX\_CHILDREN and node.revision\_count $<$ MAX\_REVISION\_COUNT)}{
        \tcp{Revise the node}
        node, delChildren, prunedChildren, addedChildren $\gets$ REVISE\_AND\_NODE(node)\;
        
        removedChildren $\gets$ delChildren + prunedChildren\;
        
        \tcp{Update stack and purge all the removed children}
        stack $\gets$ PURGE\_STACK(removedChildren, stack)\;
        
        \tcp{Check if node has been revised. If revised, update revision count}
        \If{node.status != NodeStatus.PRUNED and (len(addedChildren) != 0 or len(prunedChildren) != 0)}{
            revised $\gets$ true\;
            node.revision\_count $\gets$ node.revision\_count + 1\;
        }
    }
}

\tcp{If node is successful, mark the node as successful}
\If{node.status == NodeStatus.SUCCESS}{
    RECURSIVELY\_MARK\_SUCCESS(node.id)\;
}
\tcp{If the node has been revised or there are children to be processed, then push the node back onto the stack in ENTERING state}
\ElseIf{revised or isValid}{
    node.status $\gets$ NodeStatus.VISITED\;
    stack.append((node, NodeState.ENTERING))\;
}
\tcp{If node is invalid and not revised, prune the node}
\ElseIf{not revised and not isValid}{
    % \tcp{TreeManager recursively prunes descendants}
    prunedNodes $\gets$ RECURSIVELY\_PRUNE(node.id)\;
    
    \tcp{Update stack and purge all the pruned nodes}
    stack $\gets$ PURGE\_STACK(prunedNodes, stack)\;
    
    \tcp{Propagate failure}
    BACKTRACK\_FAILURE(node)\;
}

\Return{stack, node}\;

\end{algorithm}

\begin{algorithm}[t]
\scriptsize
\caption{PROCESS\_FAILED\_OR\_NODE(node, stack)}\label{alg:failed_or}
\SetAlgoLined

% \tcp{OR NODE IN FAILED STATE}
\tcp{If an OR node is in failed state, either execute the next promising child}
\tcp{(if any) or repair or prune the node. If node is pruned and the node's parent is AND,}
\tcp{then purge the rest of the children, or else if the parent is OR, then process the next promising child}

% \tcp{Revise, prune, backtrack}
revised $\gets$ false\;
addedChildren $\gets$ []\;

isValid $\gets$ IS\_VALID\_OR(node.children)\;

\If{isValid}{
    node.status $\gets$ NodeStatus.VISITED\;
    stack.append((node, NodeState.ENTERING))\;
    \Return{stack, node}\;
}
\Else{
    \tcp{If the node revision count has not reached the max limit, revise the node}
    \If{node.revision\_count $<$ MAX\_REVISION\_COUNT}{
        \tcp{Revise OR node}
        node, addedChildren, prunedChildren $\gets$ REVISE\_OR\_NODE(node)\;
        
        \tcp{Update stack}
        stack $\gets$ PURGE\_STACK(prunedChildren, stack)\;
        
        \tcp{Check if node was revised, update revision count}
        \If{len(addedChildren) $>$ 0}{
            revised $\gets$ true\;
            node.revision\_count $\gets$ node.revision\_count + 1\;
        }
    }
    
    \tcp{If node was revised, push it back onto the stack in ENTERING STATE}
    \If{revised}{
        node.status $\gets$ NodeStatus.VISITED\;
        \tcp{If new children were added, put this node back onto the stack}
        stack.append((node, NodeState.ENTERING))\;
    }
    \Else{
        \tcp{If not revised then prune the node}
        prunedNodes $\gets$ RECURSIVELY\_PRUNE(node.id)\;
        
        \tcp{Update stack}
        stack $\gets$ PURGE\_STACK(prunedNodes, stack)\;
        
        \tcp{Propagate failure and process next promising child of OR node}
        BACKTRACK\_FAILURE(node)\;
    }
}

\Return{stack, node}\;

\end{algorithm}

\section*{Internal Functions Reference}
\begin{table*}[htbp]
\centering
\caption{Node Status Definitions in \SA{} Framework }
\label{tab:node_statuses}
\label{tab:node_states}
\begin{tabular}{p{0.2\linewidth}p{0.72\linewidth}}
\toprule
\textbf{Status} & \textbf{Description} \\
\midrule
\emph{UNVISITED} & Initial state of all nodes before processing begins. \\
\addlinespace
\emph{VISITED} & Node has been processed but final outcome is not yet determined. \\
\addlinespace
\emph{SUCCESS} & Node's objective is achieved. For \AND\ nodes, all children succeed and collectively satisfy the objective; for \OR\ nodes, any single child's success suffices; for \ACTION\ nodes, the corresponding browser operation succeeds. \\
\addlinespace
\emph{FAIL} & Node's objective cannot be achieved. For \AND\ nodes, occurs if any child fails or is pruned; for \OR\ nodes, if the selected child fails; for \ACTION\ nodes, if the operation is invalid or unsuccessful. \\
\addlinespace
\emph{PRUNED} & Node becomes irrelevant or irresolvable due to new information discovered during execution. \\
\addlinespace
\emph{DELETED} & Node is marked as inapplicable due to failure or pruning of a predecessor sibling in an \AND\ node with ordering constraints. \\
\bottomrule
\end{tabular}
\end{table*}

\begin{table*}[t]
\centering
\scriptsize
\caption{Internal Functions Reference}
\setlength{\tabcolsep}{4pt}
\begin{tabular}{>{\raggedright\arraybackslash}p{0.40\linewidth}|>{\raggedright\arraybackslash}p{0.55\linewidth}}
\toprule
\textbf{Function Name} & \textbf{Description} \\
\midrule

PERFORM\_ROLLBACK(node) & 
Performs context rollback to restore the browser state to the URL of the given node. Returns success status. Used when processing OR node children to ensure clean state. \\
\hline

POPULATE\_NODE\_TYPE(node) & 
Expands an UNKNOWN node by determining its type (ACTION, AND, OR) and populating relevant fields like children or action details. Returns updated node, status, and action. \\
\hline

BACKTRACK\_FAILURE(node) & 
Handles backtracking when a node fails. For ordered AND nodes, deletes remaining siblings and their descendants. Propagates failure up the tree. \\
\hline

PERFORM\_ACTION(node) & 
Executes the action specified in the node (e.g., click, type, navigate). Interacts with the environment to perform the web automation task. Returns action and execution status. \\
\hline

UPDATE\_NOTES(action) & 
Extracts and updates notes from the action if any are present. Notes are used to record information during execution. \\
\hline

IS\_NOTE\_ACTION(action) & 
Checks if an action is a "note" action. Note actions don't change observations, so they don't require a full tree update. \\
\hline

ASYNC\_FULL\_UPDATE() & 
Performs a global asynchronous update of the agent's state, observation summaries, and tree structure after an observation-changing action. \\
\hline

UPDATE\_URL(node) & 
Updates the URL field of the given node to match the current environment URL. \\
\hline

IS\_VALID\_AND(children) & 
Checks if an AND node is valid by verifying it has at least one non-closed child that can be processed. \\
\hline

IS\_SUCCESSFUL\_AND(node) & 
Checks if an AND node is successful by verifying all its children have succeeded. \\
\hline

CHECK\_FOR\_COMPLETION\_AND(node) & 
Evaluates whether an AND node's objective has been completed even if not all children succeeded. Returns completion status and reasoning. Used to determine if partial completion is acceptable. \\
\hline

HAS\_ATLEAST\_ONE\_SUCCESS(node) & 
Checks if a node has at least one successful child. Used to determine if completion checking should be performed for AND nodes. \\
\hline

IS\_SUCCESSFUL\_OR(node) & 
Checks if an OR node is successful by verifying at least one of its children has succeeded. \\
\hline

IS\_VALID\_OR(children) & 
Checks if an OR node is valid by verifying it has at least one child available to process. \\
\hline

FIND\_NEXT\_PROMISING(children) & 
Finds the next most promising child to explore from a list of OR node children. Children are ordered by estimated likelihood of success. \\
\hline

REVISE\_AND\_NODE(node) & 
Revises a failed AND node by analyzing failures, potentially pruning unsuccessful children, and adding new children. Returns updated node, deleted children, pruned children, and added children. \\
\hline

REVISE\_OR\_NODE(node) & 
Revises a failed OR node by generating new alternative approaches. Returns updated node, added children, and pruned children. \\
\hline

GET\_REMAINING\_EXCLUDING\_NODE(node, parent) & 
Returns all sibling nodes that come after the given node in the parent's children list. Used for deleting remaining children in ordered AND nodes. \\
\hline

RECURSIVELY\_DELETE\_CHILDREN(children) & 
Marks the given children and all their descendants as deleted in the tree structure. Returns list of deleted nodes. \\
\hline

RECURSIVELY\_PRUNE(node\_id) & 
Recursively prunes a node and all its descendants, marking them as pruned in the tree structure. Returns list of pruned nodes. \\
\hline

RECURSIVELY\_MARK\_SUCCESS(node\_id) & 
Recursively marks a node and its relevant descendants as successful. Used to propagate success status through the tree. \\
\hline

PURGE\_STACK(deletedChildren, stack) & 
Removes all deleted or pruned nodes from the execution stack to prevent processing invalid branches. Returns updated stack. \\
\hline

\bottomrule
\end{tabular}
\end{table*}

\section*{Node Status and Type Collections}

\begin{table*}[t]
\centering
\scriptsize
\caption{Node Status and Type Collections}
\begin{tabular}{|p{0.28\linewidth}|p{0.66\linewidth}|}
\hline
\textbf{Collection} & \textbf{Description} \\
\hline

CLOSED\_STATUSES & Node statuses that indicate the node is no longer active or executable: [SUCCESS, PRUNED, DELETED] \\
\hline

FAILED\_STATUSES & Node statuses that imply failure or deleted: [PRUNED, DELETED] \\
\hline

FAILED\_OR\_PRUNED & Node statuses that indicate failure or pruned: [FAIL, PRUNED] \\
\hline

VISITED\_TYPES & Node statuses that indicate a node has been visited or processed in some way: [FAIL, VISITED, PRUNED, SUCCESS] \\
\hline

VALID\_NODE\_TYPES & All valid/expected node types (excluding UNKNOWN): [AND, OR, ACTION] \\
\hline

NON\_LEAF\_NODES & Node types that represent composite (non-leaf) logic structures: [AND, OR] \\
\hline

\end{tabular}
\end{table*}

\begin{table*}[h]
\centering
\caption{\ANDOR Tree Node Attributes}
\label{tab:node_attributes}
\begin{tabular}{|l|l|p{6cm}|}
\hline
\textbf{Attribute} & \textbf{Type} & \textbf{Description} \\
\hline
type & NodeType & The type of the node (e.g., AND, OR, ACTION) \\
\hline
description & str & A human-readable description of the node \\
\hline
id & str & A unique identifier for the node \\
\hline
steps & int & Number of steps or operations this node represents (default: 0) \\
\hline
children & List[Node] & A list of child nodes (default: empty list) \\
\hline
parent & Optional[Node] & A reference to the parent node, if any \\
\hline
status & NodeStatus & The current execution status of the node (default: UNVISITED) \\
\hline
execution\_count & int & Number of times this node has been executed (default: 0) \\
\hline
retry\_count & int & Number of retry attempts for this node (default: 0) \\
\hline
revision\_count & int & Number of revisions or changes to this node (default: 0) \\
\hline
depth & int & Depth of the node within the tree (root is 0, default: 0) \\
\hline
ordered & bool & Whether the execution of child nodes is ordered (default: True) \\
\hline
url & str & Optional URL associated with the node (default: empty string) \\
\hline
action & str & Action or command associated with the node (default: empty string) \\
\hline
reasoning & str & Explanation or rationale behind this node's inclusion or action (default: empty string) \\
\hline
others & dict & Arbitrary additional metadata or context (default: empty dict) \\
\hline
notes & List[str] & Freeform notes or comments related to the node (default: empty list) \\
\hline
\end{tabular}
\end{table*}

% \begin{table}[h]
% \centering
% \caption{Node Limits}
% \label{tab:node_limits}
% \begin{tabular}{|l|c|p{7cm}|}
% \hline
% \textbf{Limit} & \textbf{Value} & \textbf{Description} \\
% \hline
% MAX\_CHILDREN & 5 & Maximum number of children a node can have \\
% \hline
% MAX\_DEPTH & 3 & Maximum depth a node tree can grow \\
% \hline
% MAX\_EXECUTION\_COUNT & 10 & Maximum number of times a node is allowed to execute \\
% \hline
% MAX\_RETRY\_COUNT & 1 & Maximum number of retries allowed for a failed node \\
% \hline
% MAX\_REVISION\_COUNT & 3 & Maximum number of revisions or updates a node can undergo \\
% \hline
% ROOT\_MAX\_REVISION\_COUNT & 50 & Maximum number of revisions or updates a root node can undergo \\
% \hline
% MAX\_FALLBACK\_COUNT & 2 & Maximum number of fallback attempts a node can make \\
% \hline
% MAX\_EXPANSION\_COUNT & 3 & Maximum number of times a node can expand or generate children \\
% \hline
% \end{tabular}
% \end{table}

\section{Planning Tree} 
\label{app:example}
\begin{figure*}
    \centering
    \includegraphics[width=0.9\linewidth]{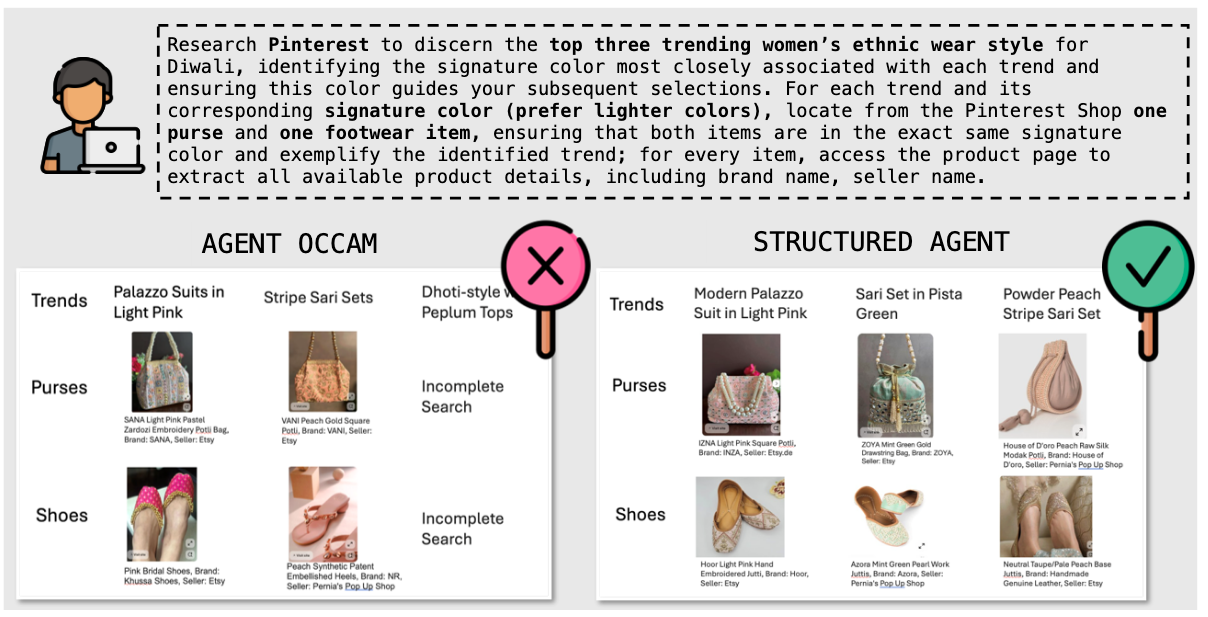}
    \caption{\ao{} vs \SA{} on a Pinterest task: \ao{} terminates its search prematurely when it fails to explore the third trend. On the other hand, \SA{} is able to successfully retrieve items for all three trending women's ethnic wear styles.}
    \label{fig:agentoccam}
\end{figure*}
\vspace{-5mm}
\begingroup

\begin{table*}[!htp]
\tiny
\centering
\begin{tabular}{p{0.9\linewidth}}
\toprule
- [0] (AND) Research Pinterest to identify the top 3 women's ethnic wear trends for Diwali. For each trend: 1) Specify the signature color currently trending. 2) Find one matching footwear style and one purse, both in the same color (same color as the trend) from the Pinterest Shop. 3) Provide the name of a potential brand or seller for each footwear and purse, and include a direct source or pin link if possible. Focus on vibrant, festive, and popular looks. Do not end your search early. Limit your review to only the first 2 scrolls of results. Do not click on external links. [success]

    - [0.1] (Atomic) Search for Diwali ethnic wear trends on Pinterest (type [401] [Diwali ethnic wear trends 2024] [1]) [success]

    - [0.2] (Atomic) Document 3 trending ethnic wear styles with their signature elements (note [Trending Diwali Ethnic Wear 2024: 1. Pink - Anupa Sari with embellished blouse from INDIASPOPUP.COM 2. Pista Green - Crush fabric sari with sequins embroidery border and yellow jacquard blouse from Etsy 3. Powder Peach - Stripe sari set with thread, pearl and sequin embroidery, featuring plunging neckline blouse from KYNAH]) [success]

    - [0.3] (AND) Navigate to Pinterest Shop and search for footwear in Pink, Pista Green, and Powder Peach shades from verified sellers specifically for Diwali looks [success]

        - [0.3.1] (Atomic) Navigate to Pinterest Shop section (click [208]) [success]

        - [0.3.2] (Atomic) type [401] [Pink ethnic juttis Diwali footwear] [1] (type [401] [Pink ethnic juttis Diwali footwear] [1]) [success]

        - [0.3.3] (Atomic) Return to previous page after completing pink footwear search (go\_back) [success]

        - [0.3.4] (Atomic) type [401] [Pista Green ethnic juttis Diwali] [1] (type [401] [Pista Green ethnic juttis Diwali] [1]) [success]

        - [0.3.5] (Atomic) Return to previous page to continue accessory search (go\_back) [success]

        - [0.3.6] (Atomic) type [401] [Powder Peach ethnic juttis Diwali] [1] (type [401] [Powder Peach ethnic juttis Diwali] [1]) [success]

    - [0.4] (OR) Search Pinterest Shop for matching ethnic clutches/purses in three colors by trying alternative search terms: 1) Search "Pink potli Diwali bag" or "Pink ethnic clutch" from verified sellers, 2) Search "Pista Green potli Diwali bag" or "Mint ethnic clutch" from verified sellers, 3) Search "Peach potli Diwali bag" or "Blush ethnic clutch" from verified sellers. Limit to first 2 scrolls of results. [success]

        - [0.4.1] (AND) Search for powder peach ethnic clutches in Shop section [success]

            - [0.4.1.1] (Atomic) Search for powder peach ethnic clutches in Shop section (type [401] [powder peach ethnic clutch Diwali] [1]) [success]

            - [0.4.1.2] (Atomic) Execute search query for powder peach ethnic clutches (type [401] [powder peach ethnic potli clutch bag Diwali] [1]) [success]

            - [0.4.1.4] (Atomic) note [Found matching powder peach potli bags from verified sellers: 1. 5 Elements Peach Brocade Pattern Potli Bag at Pernia's Pop Up Shop 2. AMYRA Peach Poly Silk Pearl Embroidered Potli Bag 3. SWABHIMANN ACCESSORIES Peach Pure Silk Potli Bag] (note [Found matching powder peach potli bags from verified sellers: 1. 5 Elements Peach Brocade Pattern Potli Bag at Pernia's Pop Up Shop 2. AMYRA Peach Poly Silk Pearl Embroidered Potli Bag 3. SWABHIMANN ACCESSORIES Peach Pure Silk Potli Bag]) [success]

            - [0.4.1.5] (Atomic) Modify search query to find matching powder peach potli bags (type [401] [powder peach potli bag Diwali] [1]) [success]

            - [0.4.1.6] (Atomic) click [208] to ensure Shop filter is active before searching for powder peach accessories (click [208] to ensure Shop filter is active before searching for powder peach accessories) [success]

            - [0.4.1.8] (Atomic) Modify search to find powder peach potli bags for Diwali (type [401] [powder peach potli bag Diwali] [1]) [success]

        - [0.5] (Atomic) Document powder peach accessory pairs with seller details (note [Powder Peach - Footwear: Coral Haze Juttis, Jutti Express formal wear, 5 Elements ethnic shoes; Purse: 5 Elements Brocade Potli with tassels, AMYRA Poly silk Pearl Embroidered Potli, Ri Ritu Kumar Net and Silk Zari Potli]) [success]

    - [0.6] (AND) Document pink and pista green accessory pairs with seller details and pin links, ensuring color coordination and authenticity of sellers [success]

        - [0.6.1] (AND) Search and document pink ethnic footwear with seller details [success]

            - [0.6.1.1] (Atomic) Search for pink ethnic footwear for Diwali (type [401] [pink ethnic juttis Diwali] [1]) [success]

            - [0.6.1.2] (Atomic) note [Found matching pink footwear options: 1. B Anu Designs - Baby Pink Punjabi Jutti with Mirror and Bead handwork 2. Miar Designs - Pink Leather Embroidered Juttis at Pernia's Pop Up Shop 3. YASSIO - Pink Silk Floral Embroidered Juttis] (note [Found matching pink footwear options: 1. B Anu Designs - Baby Pink Punjabi Jutti with Mirror and Bead handwork 2. Miar Designs - Pink Leather Embroidered Juttis at Pernia's Pop Up Shop 3. YASSIO - Pink Silk Floral Embroidered Juttis]) [success]

            - [0.6.1.3] (Atomic) Return to previous page to continue accessory search (go\_back) [success]

        - [0.6.2] (Atomic) Document matching pink potli/clutch from search results (note [Found matching pink brocade potli bag from 5 Elements with tassels and beads embellishments, suitable for festive wear]) [success]

        - [0.6.3] (Atomic) Return to previous page after powder peach potli search (go\_back) [success]

        - [0.6.4] (Atomic) Search for pista green ethnic footwear using specific search terms (type [401] [Pista green embroidered juttis Diwali] [1]) [success]

        - [0.6.5] (Atomic) Search for matching pista green clutch/potli (type [401] [Pista green brocade potli Diwali] [1]) [success]

        - [0.6.6] (Atomic) Document complete accessory details with seller information for pista green trend accessories: Footwear - House of Vian's Crepe Embroidered Juttis and 5 Elements' Leatherette Embellished Juttis; Purse - 5 Elements Green Brocade Embroidered Potli with tassels and dupion lining, and Tarini Nirula Green Brocade Stone Embellished Potli with antique gold frame (Document complete accessory details with seller information for pista green trend accessories: Footwear - House of Vian's Crepe Embroidered Juttis and 5 Elements' Leatherette Embellished Juttis; Purse - 5 Elements Green Brocade Embroidered Potli with tassels and dupion lining, and Tarini Nirula Green Brocade Stone Embellished Potli with antique gold frame) [success]

        - [0.6.7] (Atomic) Search for powder peach ethnic footwear (type [401] [powder peach ethnic juttis Diwali] [1]) [success]

        - [0.6.10] (AND) Search for pink ethnic footwear pins from verified sellers and save direct links [success]

            - [0.6.10.1] (Atomic) View details of pink leather embroidered juttis for Pink category (click [17999]) [success]

            - [0.6.10.2] (Atomic) note [Pink leather juttis by Miar Designs - Features: Leather base with coin, thread, pearl, sequins and beads embroidery, Available at Pernia's Pop Up Shop, Price: 40,00, Formal occasion wear] (note [Pink leather juttis by Miar Designs - Features: Leather base with coin, thread, pearl, sequins and beads embroidery, Available at Pernia's Pop Up Shop, Price: 40,00, Formal occasion wear]) [success]

            - [0.6.10.3] (Atomic) Return to previous page to continue purse search (go\_back) [success]

        - [0.6.11] (Atomic) Search for matching pink potli/clutch pins (type [401] [pink ethnic potli clutch Diwali] [1]) [success]

        - [0.6.14] (Atomic) Verify and document color coordination between footwear-purse pairs (note [Color coordination verified: Pink category - Pink Beaded Mirror Work Potli and Pink Brocade Potli perfectly match the pink leather juttis from Miar Designs in shade and festive style]) [success]

    - [0.7] (Atomic) Document direct pin links for pink accessories from verified sellers (note [Found verified seller pin links: 1. Pink Brocade Tassels Embellished Potli by 5 Elements - Direct Pin ID 20730 2. Pink Brocade Ghungroo Embroidered Potli by 5 Elements - Direct Pin ID 20752 3. Pink Beaded Mirror Work Potli by Etsy verified seller - Direct Pin ID 21242]) [success]

    - [0.8] (AND) Review and document direct pin links for all three color categories ensuring each has verified footwear and purse pairs with seller details [success]

        - [0.8.1] (Atomic) Document completed findings for Pink category (note [Pink Category Complete - Footwear: Miar Designs, Purses: Pink Beaded Mirror Work Potli (Etsy), Pink Brocade Potli (5 Elements)]) [success]

        - [0.8.2] (Atomic) Navigate to Pinterest Shop section for ethnic potli clutches (click [20411]) [success]

        - [0.8.7] (Atomic) Search for powder peach potli options (type [401] [powder peach ethnic potli clutch Diwali] [1]) [success]

        - [0.8.8] (Atomic) Search for matching Pista Green potli options with verified sellers and clear product details (type [401] [pista green ethnic potli clutch Diwali] [1]) (Search for matching Pista Green potli options with verified sellers and clear product details (type [401] [pista green ethnic potli clutch Diwali] [1])) [success]

        - [0.8.9] (Atomic) Document verified seller information for Pista Green potlis (note [Verified Pista Green Potlis: 1. PLODE - Green Velvet Swarovski Triangle Potli with crystal embellishments and rhinestone strap 2. Velleso - Green Mirror Work Potli with crystal drops and silver threadwork (24x23cm) 3. The Pink Potli - Green Raw Silk Potli with bead work and tassel embellishments]) [success]

        - [0.8.10] (Atomic) Search for matching Powder Peach ethnic footwear options (type [401] [powder peach ethnic footwear Diwali] [1]) [success]

        - [0.8.11] (Atomic) Search for matching Pista Green ethnic footwear options with verified sellers and clear product details (type [401] [pista green ethnic footwear Diwali] [1]) [success]

        - [0.8.12] (Atomic) Document pin links and details for Pista Green category footwear and purse pairs (note [Found direct pin links for Pista Green category: 1) House of Vian Green Crepe Embroidered Juttis with crystals and gota patti from Pernia's Pop Up Shop 2025, 2) 5 Elements Green Leatherette Embellished Juttis with beadwork from Pernia's Pop Up Shop 2025]) [success]\\
\bottomrule
\end{tabular}
\caption{Planning tree for the task \textit{``Research Pinterest to identify the top 3 women's ethnic wear trends for Diwali. For each trend: 1) Specify the signature color currently trending. 2) Find one matching footwear style and one purse, both in the same color (same color as the trend) from the Pinterest Shop. 3) Provide the name of a potential brand or seller for each footwear and purse, and include a direct source or pin link if possible. Focus on vibrant, festive, and popular looks. Do not end your search early. Limit your review to only the first 2 scrolls of results. Do not click on external links.''}}
\label{tab:example1}
\end{table*}
\FloatBarrier
\endgroup

% \begin{figure}[h!]
%     \centering
%     \includegraphics[width=0.99\linewidth]{images/tree.png}
%     \caption{Planning tree for the task \textit{``Research Pinterest to identify the top 3 women's ethnic wear trends for Diwali. For each trend: 1) Specify the signature color currently trending. 2) Find one matching footwear style and one purse, both in the same color (same color as the trend) from the Pinterest Shop. 3) Provide the name of a potential brand or seller for each footwear and purse, and include a direct source or pin link if possible. Focus on vibrant, festive, and popular looks. Do not end your search early. Limit your review to only the first 2 scrolls of results. Do not click on external links.''}.}
%     \label{fig:tree}
% \end{figure}

\begin{table*}[h!]%
{\footnotesize 
\centering
\caption{%
  Trajectory and final response comparison on task \textbf{Diwali-05}:
  \textit{``Research Pinterest to identify the top 3 women's ethnic wear trends for Diwali. For each trend: 1) Specify the signature color currently trending. 2) Find one matching footwear style and one purse, both in the same color (same color as the trend) from the Pinterest Shop. 3) Provide the name of a potential brand or seller for each footwear and purse, and include a direct source or pin link if possible. Focus on vibrant, festive, and popular looks. Do not end your search early. Limit your review to only the first 2 scrolls of results. Do not click on external links.''}
  Both agents use \texttt{claude-3-5-sonnet-20241022}.
}
\label{tab:trajectory}
\renewcommand{\arraystretch}{1.4}
\begin{tabularx}{\textwidth}{@{} l L L @{}}
\toprule
\textbf{Step} &
  \textbf{\textcolor{agentA}{\SA{}}} &
  \textbf{\textcolor{agentB}{\ao{}}} \\
\midrule

1  & \texttt{search} ``Diwali ethnic wear trends 2024''
   & \texttt{search} ``Diwali ethnic wear trends 2024'' \\

2  & \texttt{click} search bar
   & \texttt{click} first result \\

3  & \texttt{search} ``light pink ethnic purse palazzo style''
   & \texttt{click} search bar \\

4  & \texttt{click} product result
   & \textit{[note]} Top 3 trends: (1)~Palazzo suits – Light Pink, (2)~Stripe sari sets – Powder Peach, (3)~Dhoti-style with peplum tops \\

5  & \texttt{go\_back}
   & \texttt{search} ``light pink purse ethnic'' \\

6  & \texttt{search} ``light pink traditional juttis wedding''
   & \texttt{click} product result \\

7  & \texttt{click} product result
   & \texttt{search} ``light pink ethnic sandals heels'' \\

8  & \texttt{go\_back}
   & \texttt{search} ``light pink ethnic sandals heels'' (retry) \\

9  & \texttt{search} ``pista green traditional potli clutch''
   & \textit{[note]} Purse: SANA Light Pink Zardozi Potli, \$45.00 (Etsy) \\

10 & \texttt{search} ``pista green ethnic potli bag traditional'' (retry)
   & \texttt{search} ``light pink ethnic sandals heels'' (retry) \\

11 & \texttt{go\_back}
   & \texttt{search} ``light pink ethnic sandals heels'' (retry) \\

12 & \texttt{search} ``mint green ethnic jutti traditional''
   & \texttt{go\_back} \\

13 & \texttt{go\_back}
   & \texttt{click} search bar \\

14 & \texttt{click} product result
   & \texttt{click} search bar \\

15 & \texttt{search} ``mint green ethnic bags wedding''
   & \texttt{search} ``light pink ethnic heels sandals'' \\

16 & \texttt{search} ``mint green ethnic footwear wedding''
   & \texttt{click} product result \\

17 & \texttt{search} ``peach traditional clutch ethnic''
   & \textit{[note]} Footwear: Pink Bridal Juttis, \$63.99 (Etsy) \\

18 & \texttt{search} ``peach traditional jutti wedding''
   & \texttt{search} ``powder peach ethnic purse'' \\

19 & —
   & \texttt{click} product result \\

20 & —
   & \textit{[note]} Purse: VANI Peach Gold Potli, \$45.00 (Etsy) \\

21 & —
   & \texttt{search} ``powder peach ethnic sandals heels'' \\

22 & —
   & \texttt{click} product result \\

23 & —
   & \textit{[note]} Footwear: Peach Embellished Heels, \$49.00 (Pernia's) \\

24 & —
   & \texttt{search} ``ethnic purse for dhoti style'' \\

25 & —
   & \texttt{stop} (explicit failure on trend~3) \\

% \midrule
% \textbf{Total steps} & 18 & 25 \\

\midrule
\multicolumn{3}{@{}l}{\textbf{Final Response}} \\[2pt]

\textbf{Trend 1}
  & \textbf{Light Pink Palazzo Suit.}
    \textit{Bag:} IZNA Light Pink Square Potli, \$45.00 (Etsy.de); zardozi embroidery, pearl handle.
    \textit{Footwear:} Hoor Light Pink Hand Embroidered Jutti, \$39.51 (Etsy).
  & \textbf{Light Pink Palazzo Suit.}
    \textit{Bag:} SANA Zardozi Potli, \$45.00 (Etsy).
    \textit{Footwear:} Pink Bridal Juttis, \$63.99 (Etsy). \\

\textbf{Trend 2}
  & \textbf{Pista Green Crush Fabric Sari Set.}
    \textit{Bag:} ZOYA Mint Green Gold Drawstring Bag, \$57.50 (Etsy); mirror embroidery.
    \textit{Footwear:} Azora Mint Green Pearl Work Juttis (Pernia's Pop-Up Shop).
  & \textbf{Powder Peach Stripe Sari Set.}
    \textit{Bag:} VANI Peach Gold Potli, \$45.00 (Etsy).
    \textit{Footwear:} Peach Embellished Heels, \$49.00 (Pernia's Pop-Up Shop). \\

\textbf{Trend 3}
  & \textbf{Powder Peach Stripe Sari Set (KYNAH).}
    \textit{Bag:} House of D'oro Peach Raw Silk Modak Potli, \$59.00 (Pernia's).
    \textit{Footwear:} Pale Peach Handcrafted Punjabi Jutti, \$75.00 (Etsy).
  & \textbf{Dhoti-style with Peplum Tops.}
    Accessories not found; failure explicitly acknowledged via \texttt{stop}. \\

\midrule
\bottomrule
\end{tabularx}
}%
\end{table*}

\end{document}